\documentclass{article}

    \PassOptionsToPackage{numbers, compress}{natbib}

    \usepackage[preprint]{neurips_2025}

\usepackage[table]{xcolor}
\definecolor{tabfirst}{rgb}{1, 0.7, 0.7} 
\definecolor{tabsecond}{rgb}{1, 0.85, 0.7} 
\definecolor{tabthird}{rgb}{1, 1, 0.7} 

\usepackage[utf8]{inputenc} 
\usepackage[T1]{fontenc}    
\usepackage{hyperref}       
\usepackage{url}            
\usepackage{booktabs}       
\usepackage{amsfonts}       
\usepackage{nicefrac}       
\usepackage{microtype}      

\usepackage{amsmath,amssymb}
\usepackage{multirow}
\usepackage{caption}
\usepackage{adjustbox}
\usepackage{algorithm}
\usepackage{algpseudocode}
\usepackage{subcaption}
\usepackage{wrapfig} 
\usepackage{enumitem}

\usepackage{tikz}
\usepackage[font=small]{caption}
\usepackage[super]{nth}
\usepackage{tabularx}
\usepackage{makecell}

\usepackage{ulem}

\usepackage{graphicx, wrapfig}
\usepackage{placeins}

\title{UAV4D: Dynamic Neural Rendering of Human-Centric UAV Imagery using Gaussian Splatting}

\author{%
  Jaehoon Choi\textsuperscript{\rm 1} 
  \And 
  Dongki Jung\textsuperscript{\rm 1} 
  \And
  Christopher Maxey\textsuperscript{\rm 1,2}
  \And 
  Sungmin Eum\textsuperscript{\rm 2}
  \and 
  \textbf{Yonghan Lee}\textsuperscript{\rm 1}  
  \quad
  \textbf{Dinesh Manocha\textsuperscript{\rm 1}}
  \quad 
  \textbf{Heesung Kwon\textsuperscript{\rm 2}} \\ 
  University of Maryland, College Park\textsuperscript{\rm 1}\quad
  DEVCOM Army Research Laboratory\textsuperscript{\rm 2}\\
}

\begin{document}

\maketitle

\begin{abstract}
  Despite significant advancements in dynamic neural rendering, existing methods fail to address the unique challenges posed by UAV-captured scenarios, particularly those involving monocular camera setups, top-down perspective, and multiple small, moving humans, which are not adequately represented in existing datasets. 
  In this work, we introduce UAV4D, a framework for enabling photorealistic rendering for dynamic real-world scenes captured by UAVs. Specifically, we address the challenge of reconstructing dynamic scenes with multiple moving pedestrians from monocular video data without the need for additional sensors. We use a combination of a 3D foundation model and a human mesh reconstruction model to reconstruct both the scene background and humans. We propose a novel approach to resolve the scene scale ambiguity and place both humans and the scene in world coordinates by identifying human-scene contact points. Additionally, we exploit the SMPL model and background mesh to initialize Gaussian splats, enabling holistic scene rendering.  We evaluated our method on three complex UAV-captured datasets: VisDrone, Manipal-UAV, and Okutama-Action, each with distinct characteristics and $10-50$ humans. Our results demonstrate  the benefits of our approach over existing methods in novel view synthesis, achieving a 1.5 dB PSNR improvement and superior visual sharpness. 
\end{abstract}

\section{Introduction}
\label{sec:intro}

\begin{figure}[t]
    \centering
    \includegraphics[width=0.85\linewidth]{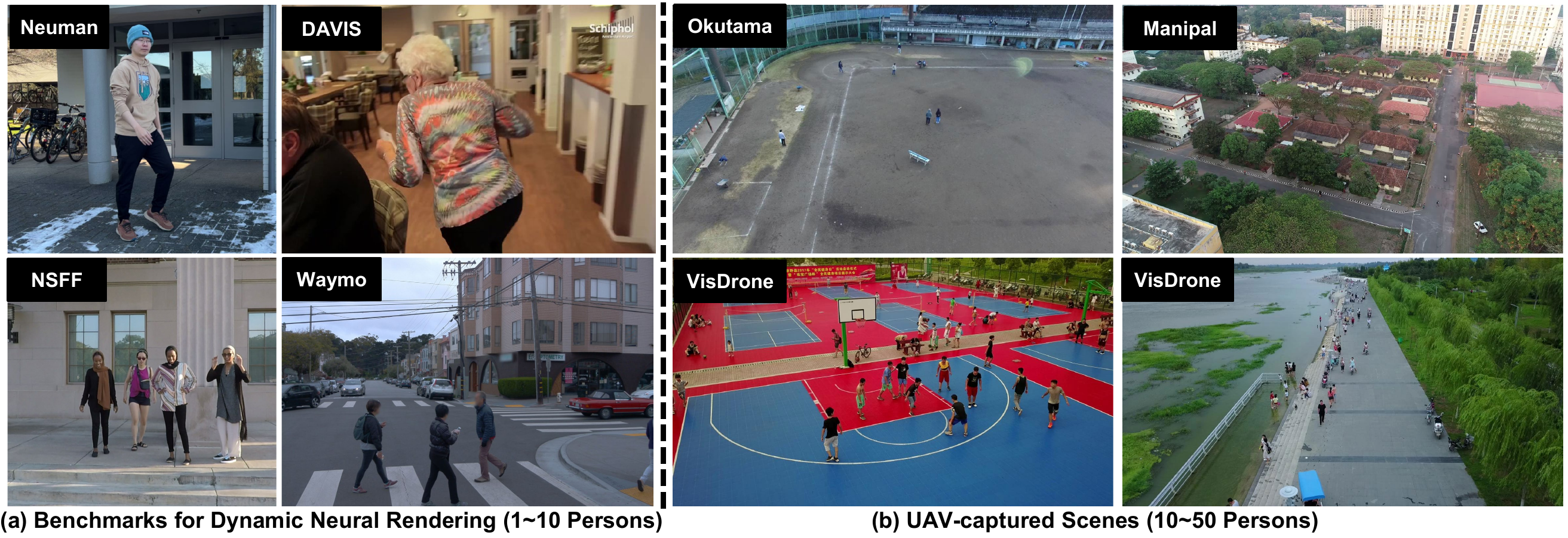}
    \vspace{-2mm}
     \caption{\textbf{Comparison between the Existing Benchmarks for Dynamic Neural Rendering and UAV captured Datasets.} UAVs typically cover wide areas from a top-down perspective, capturing dynamic scenes \cite{barekatain2017okutama, akshatha2023manipal, cao2021visdrone} with a distinct setup: they contain numerous human instances (ranging from 10 to 50 per frame), each occupying only a tiny fraction of the image area. 
     }
     \vspace{-7mm}
     \label{fig:datasets} 
\end{figure}

Unmanned Aerial Vehicles (UAVs) have become essential tools for capturing large-scale environments such as streets and urban areas, supporting applications like world mapping, human activity analysis, and security monitoring.
Recent advances in neural rendering \cite{mildenhall2021nerf,kerbl20233d} have enabled photorealistic scene reconstructions from UAV data, driving progress in large-scale mapping.
However, most of the existing works \cite{turki2022mega,lin2024vastgaussian,liu2024citygaussian,maxey2024uav,maxey2024tk} primarily focus on static scenes, overlooking the more challenging dynamic scenes, even though real-world environments are inherently dynamic, involving human activities and complex background changes. Although dynamic neural rendering methods are rapidly evolving, most existing approaches \cite{zhu2025dynamic} focus on phone-captured video scenes \cite{yang2024deformable,wu20244d,stearns2024dynamic,som2024}, multi-view synchronized camera setups \cite{luiten2024dynamic,xu20244k4d}, or multi-sensor systems scenarios \cite{chen2025omnire} such as multi-camera rigs and LiDAR sensors, as seen in Fig. \ref{fig:datasets}.
Despite their relevance, dynamic neural rendering methods tailored for UAV scenarios remain underexplored. To the best of our knowledge, only a few works \cite{maxey2024uav,maxey2024tk} have attempted this, and they have some limitations.
Some approaches use implicit neural representations \cite{fridovich2023k} for volumetric rendering to generate novel-view images but face challenges in terms of maintaining high-fidelity outputs.

Applying dynamic neural rendering to UAV scenes presents two fundamental challenges.
First, due to hardware constraints, UAVs typically lack access to a diverse range of sensors and cannot utilize multi-view synchronized camera systems or multi-sensor configurations. 
Consequently, scene reconstruction must rely solely on monocular dynamic video, which presents substantial challenges \cite{gao2022monocular} for both accurate geometry reconstruction and photorealistic rendering—particularly in dynamic environments. 
Furthermore, most dynamic objects in UAV-captured scenes are relatively small and multiple moving humans often appear simultaneously. 
These conditions pose significant challenges, making it difficult to apply the state-of-the-art approaches \cite{zhu2024motiongs,som2024,zheng2025gstar,stearns2024dynamic} that rely on depth estimation \cite{yang2024depth}, optical flow \cite{xu2022gmflow} and point tracking models \cite{yang2023track,karaev2024cotracker,doersch2023tapir}. 
Even recent 4D reconstruction methods~\cite{zhang2024monst3r,wang2025continuous,feng2025st4rtrack} powered by 3D foundation models~\cite{wang2024dust3r} face difficulties when reconstructing small dynamic humans.
This challenge arises primarily because existing large-scale datasets for dynamic scenes rarely include scenarios with small and moving human instances.
As a result, many of these previously mentioned methods are ineffective in UAV-based scenarios.
Inspired by recent progress in human-scene reconstruction \cite{jiang2022neuman,kocabas2024hugs}, we focus specifically on reconstructing amd rendering moving humans, which are a key component of dynamic scenes captured by UAVs.

\noindent\textbf{Main Results:} 
We present UAV4D, a novel neural rendering framework that reconstructs dynamic human instances and static backgrounds from UAV-captured real-world data, enabling temporally consistent and photorealistic novel-view synthesis. Our methods begin with reconstructing the joint geometry of small dynamic humans and static backgrounds from monocular UAV videos by integrating SMPL-based human meshes with dense background geometry. We leverage the 3D foundation model \cite{wang2025vggt} to estimate camera poses and perform dense background reconstruction, yielding consistent and robust background meshes across diverse scenarios. We observe that HMR2 \cite{goel2023humans} and SAM2 \cite{ravi2024sam} exhibit strong performance in UAV contexts, and we adopt them to recover 4D human motion trajectories as temporally aligned SMPL body meshes \cite{loper2023smpl} from monocular videos. However, the human and background meshes suffer from inconsistent scales, and the absolute global positioning of the human meshes remains unknown. To resolve these discrepancies, we propose a scale optimization technique that aligns the background geometry from the 3D foundation model to the metric human mesh, estimating an optimal global scale parameter. Using the optimized scale parameter, we unify the background and human meshes within a consistent world coordinate system over time. Finally, we localize the human mesh in the world space by unprojecting the 2D ground contact point into a 3D location on the reconstructed background mesh.

We then initialize separate 3D Gaussian representations for humans and background using their respective geometric reconstructions. 
We maintain separate Gaussian splats for humans and background, which are jointly optimized and composited to render photorealistic full frames. 
Our approach explicitly decomposes dynamic human instances and static backgrounds by independently optimizing their Gaussian splats. 
By initializing with a strong Gaussian prior from the human SMPL mesh, we ensure that the optimization process properly learns the human Gaussian splats, rather than neglecting them due to their small pixel regions.
The key contributions of this work include:
\begin{itemize}[leftmargin=2em]

    \item We introduce UAV4D, the first neural rendering framework that reconstructs dynamic human motions and static backgrounds from UAV-captured monocular videos, enabling temporally consistent and photorealistic novel-view synthesis of full scenes.
    \item We propose a novel method for resolving the scale ambiguity issue in 3D foundation models and for aligning multiple human meshes with the background mesh. We also present a human placement method by identifying the ground contact point. By initializing both the human and background meshes in world coordinates, we decompose the human and background Gaussian splats to render the complete image.
    \item To the best of our knowledge, UAV4D is the first to enable dynamic Gaussian splatting for UAV-captured scenes with large number of small moving humans (e.g., up to 50). We evaluated its performance on three datasets with unique characteristics, our method achieves state-of-the-art rendering quality, outperforming prior methods by up to 1.5 dB in PSNR.
\end{itemize}
\vspace{-2mm}



\section{Related Work}
\noindent\textbf{3D Reconstruction for Static and Dynamic Scenes.}
Traditional 3D reconstruction typically consists of two components: Structure-from-Motion (SfM) and Multi-View Stereo (MVS) systems. SfM \cite{wu2013towards,schonberger2016structure} extracts and matches features across multi-view images to estimate camera poses and sparse 3D points via triangulation \cite{hartley2003multiple}. MVS \cite{furukawa2015multi,schonberger2016pixelwise} then estimates dense depth maps for each image by performing an accurate and efficient global search for the optimal 3D plane hypotheses. 
Recently, learning-based SfM \cite{teed2018deepv2d,teed2021droid,wei2020deepsfm,wang2024vggsfm} and MVS \cite{yao2018mvsnet,yu2020fast,kim2021just,zhang2023vis,wang2024dust3r,wang2025vggt} approaches have leveraged large-scale datasets to enable end-to-end differentiable SfM and dense reconstruction. 
Notably, DUSt3R \cite{wang2024dust3r} and MASt3R \cite{leroy2024grounding,duisterhof2024mast3r} have emerged as breakthroughs by estimating globally consistent point maps from just two views without requiring any camera information. 
VGGT \cite{wang2025vggt} also leverages point maps during training and modifies the network architecture by incorporating alternative attention mechanisms for improved global alignment.
However, these methods are primarily designed for static scenes and struggle to handle dynamic components such as humans. 
Early methods for non-rigid reconstruction relied on RGB-D sensors \cite{zollhofer2014real,newcombe2015dynamicfusion,dou2016fusion4d,innmann2016volumedeform} for dense tracking and reconstruction, or employed non-rigid SfM approaches \cite{akhter2010trajectory,garg2013dense,zhu2014complex}.
Alternatively, researchers have utilized learning-based methods that leverage large datasets to train neural networks. 
Some works \cite{zhang2022structure,jung2021dnd} train monocular depth estimation networks using pre-computed optical flow, or jointly optimize depth and camera poses \cite{kopf2021robust}. 
More recently, concurrent studies \cite{leroy2024grounding,duisterhof2024mast3r,feng2025st4rtrack,lu2024align3r} have achieved remarkable ability in dynamic scenes by fine-tuning MASt3R \cite{duisterhof2024mast3r} with large-scale dynamic datasets to estimate both camera poses and point maps.
However, none of these previous methods perform well on UAV-captured datasets, as the humans in our data are relatively small compared to those in their training datasets.
Therefore, we decompose the scene into static and dynamic components, leveraging the prediction results of VGGT for separate tasks within our pipeline.

\noindent\textbf{Dynamic Neural Rendering.} 
In the field of neural rendering, Neural Radiance Fields (NeRF) \cite{mildenhall2021nerf} and 3D Gaussian Splatting (3DGS) \cite{kerbl20233d} are milestone works that enable novel view synthesis of static scenes from multi-view images. 
NeRF-based methods \cite{barron2021mip,barron2022mip,muller2022instant,barron2023zip} use a multilayer perceptron (MLP) to represent the scene implicitly, while 3DGS \cite{kerbl20233d} employs a set of 3D Gaussian primitives that are differentiably splatted onto the 2D image plane. Researchers have increasingly adopted 3DGS for its more efficient rendering performance \cite{celarek2025does}. Earlier methods were limited to static scenes, but researchers quickly extended them to handle dynamic scenes.
One popular direction \cite{yang2024deformable,wu20244d,stearns2024dynamic,lin2024gaussian} is deformation field-based methods, which often use a time-conditioned deformation network to warp 3D Gaussians into each time frame.
Another line of work \cite{yang2023real,duan20244d} directly models 4D Gaussian primitives by integrating 3D Gaussians with the time dimension.
Some approaches \cite{zhu2024motiongs,som2024,zheng2025gstar,stearns2024dynamic} estimate 3D scene motion and point correspondences using pretrained models \cite{xu2022gmflow,doersch2023tapir,yang2024depth,karaev2024cotracker,yang2023track}  to deform 3D Gaussians.
HuGS \cite{kocabas2024hugs} presents an approach similar to ours, using Gaussian splats to represent both humans and the scene. However, as their target scenes consist of phone-captured videos \cite{jiang2022neuman}, they focus exclusively on a single, large person within the scene.
Most of these methods are not designed for UAV-captured datasets, as they struggle to handle scenes with multiple pedestrians (more than 10 people) and very small human figures. UAV-Sim \cite{maxey2024uav} and TK-planes \cite{maxey2024tk} are similar tasks to ours, as they use multi-resolution feature spaces in NeRF to represent small dynamic objects. However, we take a different approach by reconstructing the human and background meshes to initialize the Gaussian splats.    We then render all the Gaussian splats to generate the full images. 



\section{Method}
\label{sec:digitaltwin}
Our method takes as input monocular videos of human-centric scenes \cite{akshatha2023manipal,cao2021visdrone,barekatain2017okutama} captured by UAVs. Given this input, our goal is to enable dynamic neural rendering using 3D Gaussian splatting \cite{kerbl20233d}. To achieve this, our approach involves initializing the human-scene from 2D images, reconstructing the 3D geometry of both the human and the scene, and optimizing the Gaussian splats for dynamic neural rendering.

\subsection{Preliminaries}
\label{sec:Preliminaries}
\textbf{3D Gaussian Splatting (3DGS)} \cite{kerbl20233d} represents scenes as an unordered set of 3D Gaussian primitives, rendered through a differentiable rasterization process \cite{zwicker2002ewa}. 
Each Gaussian component $g_i$ is defined as its mean $\mu_{i} \in \mathbb{R}^3$ and 3D covariance $\Sigma_{i} \in \mathbb{R}^{3\text{x}3}$. 
The covariance matrix is decomposed into two learnable components, a scaling matrix $S_{i} \in \mathbb{R}^{3}$ and a rotation matrix $R_{i} \in \mathbb{R}^{3\text{x}3}$, as $\Sigma_{i} = R_{i}S_{i}{S_{i}}^{T}{R_{i}}^{T}$.
Furthermore, each Gaussians store an opacity logit $o_i \in \mathbb{R}$ and color $c_i$ defined by spherical harmonics (SH) coefficients. 
To render an image from a given view, the color of each pixel is computed by 
$\alpha$-blending K ordered Gaussians using the following equation: $C = \sum_{i=1}^{K} c_{i}\alpha_{i}\prod_{j=1}^{i-1}(1-\alpha_{j})$
where density $\alpha_{i}$ is computed by the multiplication of 2D Gaussians with covariance $\Sigma'_{i} \in \mathbb{R}^{2\text{x}2}$  and a learnable point-wise opacity $o_i$. 

\textbf{Skinned Multi-Person Linear (SMPL) Model} \cite{loper2023smpl} is a commonly used parametric human body model. The SMPL model exploits a template human mesh in the canonical rest pose, defined as $\mathcal{M}_{h} \in (\mathcal{V}_{c}, \mathcal{F})$ with vertices $\mathcal{V}_{c} \in \mathbb{R}^{6890 \text{x} 3}$ . 
It takes body shape parameters $\theta, \beta$ as input and outputs a posed 3D mesh with deformed vertices, where $\theta \in \mathbb{R}^{24\text{x}3\text{x}3}$ and $\beta \in \mathbb{R}^{10}$ represent the pose and shape parameters, respectively. The pose parameters $\theta$ consist of the global rotation of the root joint (pelvis), and the 23 local rotations of other articulated joints relative to their parents along the kinematic chain.
SMPL uses $n_{b}$ predefined joints and Linear Blend Skinning (LBS) weights $W$, defined as $W(v_{c}) = \{w_1, ..., w_{n_{b}}\}$.
The vertex $v_{c}$ in the canonical template mesh can be deformed to the articulated space via the LBS, as $v = (\Sigma^{n_{b}}_{\tiny{k=1}} W_{k}(v_{c})B_{k})v_{c}$, where $B = \{B_{1}, ..., B_{n_{b}} \}$ is the target joint transformation. 

\begin{figure}[t]
    \centering
    \includegraphics[width=0.99\linewidth]{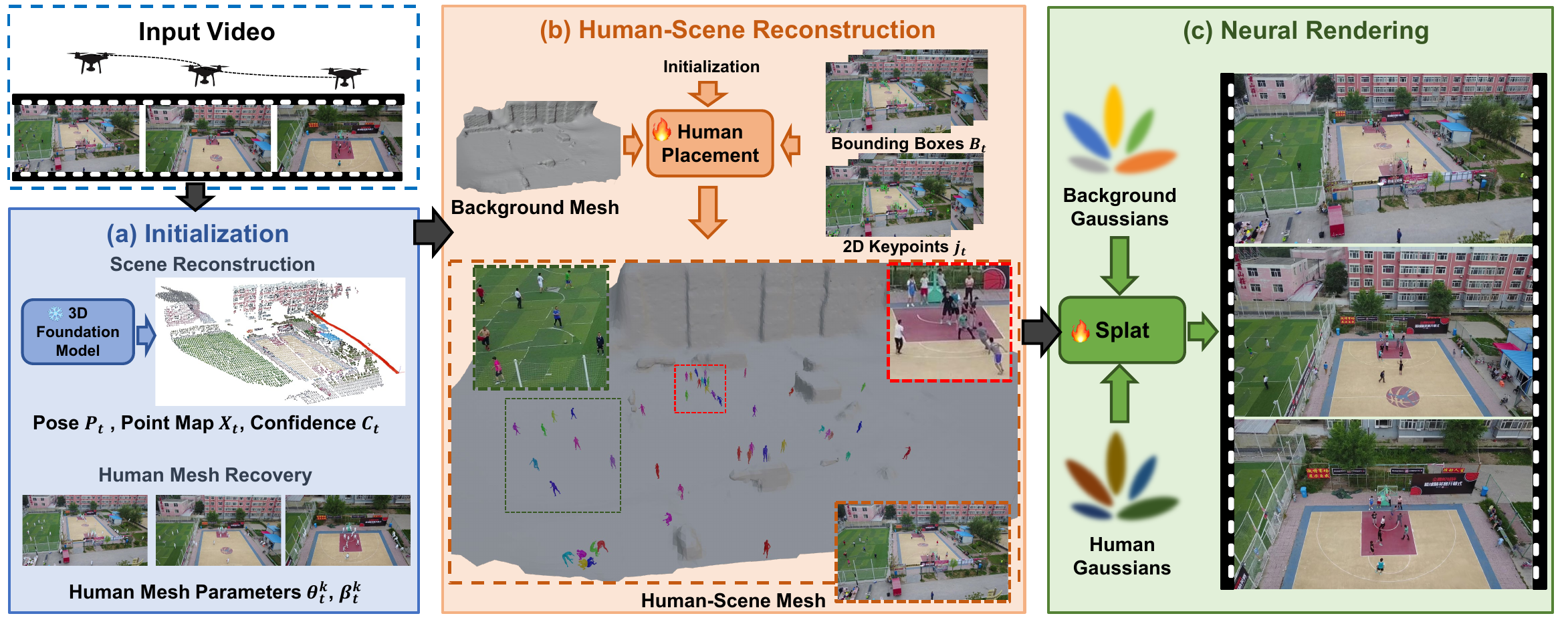}
    \vspace{-2mm}
    \caption{
    \textbf{Overview of Our Approach}: In Section \ref{sec:Initialization}, we begin the initialization process for reconstructing the background scene and the human mesh. Then, in Section \ref{sec:HumanPlacement}, we reconstruct the background mesh, determine the optimal scale for alignment, and position the human mesh by identifying the ground contact points. In Section \ref{sec:rendering}, we jointly optimize the human and background Gaussians for rendering.}
    \label{fig:framework}
    \vspace{-2mm}
\end{figure}



\subsection{Initialization}
\label{sec:Initialization}
At this stage, we first utilize off-the-shelf models \cite{wang2025vggt,goel2023humans,ravi2024sam} to obtain initial estimates, such as camera poses, point maps, human masks, human meshes, and other relevant data, for reconstructing the geometry of both the background scene and the human.

\noindent\textbf{Scene.} Given a monocular video sequence $\left\{ I_t \in \mathbb{R}^{H \times W \times 3} \right\}_{\scriptscriptstyle{t=1}}^{\scriptscriptstyle{N}}$, we first employ a 3D foundation model \cite{wang2025vggt} to estimate camera poses $\left\{P_{t}\right\}_{\scriptscriptstyle{t=1}}^{\scriptscriptstyle{N}}$ including intrinsics $K$ and extrinsics, rotation $R_t$ and translation $T_t$, per-view point map $X_t \in \mathbb{R}^{H \times W \times 3}$ and  corresponding confidence $C_t \in \mathbb{R}^{H \times W \times 1}$. 

\noindent\textbf{Human.}
Due to the relatively small size of the human subjects, it is challenging to apply recent 3D tracking systems \cite{goel2023humans} trained on standard tracking tasks.
Since our goal is precise reconstruction and rendering, we use bounding boxes $\left\{B^k_t\right\}_{\scriptscriptstyle{k=1}}^{\scriptscriptstyle{K_t}}$ as prompts for SAM2 \cite{ravi2024sam} to generate accurate instance masks for $K_t$ humans.
These human segmentation masks $\left\{M^k_t\right\}_{\scriptscriptstyle{k=1}}^{\scriptscriptstyle{K_t}}$ are then fed into HMR2.0 \cite{goel2023humans} to predict camera-frame human meshes from each image $I_t$ for $K_t$ detected people. 
Thus, we can obtain the human mesh parameters for each human, defined as $\left\{ \theta^k_t, \beta^k_t \right\}_{\scriptscriptstyle{k=1}}^{\scriptscriptstyle{K_t}}$ at timestep $t$.

\noindent\textbf{Human Mesh Refinement.}
However, due to the challenging nature of UAV imagery, HMR2.0 \cite{goel2023humans} often produces noisy initializations. 
We introduce a refinement method to remove human mesh parameters with poor quality. 
Since the human meshes reside in the camera coordinate system, we project each individual human mesh into the image space and compute the bounding box $\left\{\bar{B}^k_t\right\}_{k=1}^{K_t}$ for each projected human. 
We then calculate the ratio between $B^k_t$ and $\bar{B}^k_t$, and remove any human meshes for which $\bar{B}^k_t/B^k_t$ exceeds a threshold $\eta_{box}$.
Furthermore, we measure the overlap between dilated human masks $M^k_{t-1}, M^k_{t+1}$ from the previous and next time frames.
If the current bounding box with missing human mesh parameters overlaps in both adjacent frames, we interpolate the missing parameters $\theta^k_t, \beta^k_t$ using the human mesh parameters from the previous frame ($\theta^k_{t-1}, \beta^k_{t-1}$) and the next frame ($\theta^k_{t+1}, \beta^k_{t+1}$) applying linear interpolation to both the quaternion pose parameters and the shape vectors.



\subsection{Joint Human-Scene Reconstruction}
\label{sec:HumanPlacement}
Although our initialization provides a reasonable estimate of human poses, all predictions remain in the camera coordinate space. 
As a result, when the camera moves, the estimated human motion becomes implausible. 
To align the human and scene in the world coordinate, we reconstruct the background region and identify human-ground contact points to position the human mesh in the scene accurately. 
We observe that UAVs typically capture wide-area footage with the full human body visible, allowing us to leverage 2D bounding boxes to estimate the human-ground contact points.
However, aligning the human mesh with the background remains challenging due to two key issues: (1) in complex environments, the reconstructed background geometry is often noisy and structurally intricate, making it difficult to reliably detect human-scene contact points; and (2) the inherent scale ambiguity of the 3D foundation model \cite{wang2025vggt}.

\noindent\textbf{Background Geometry Reconstruction.} To address the first issue, we first utilize all point maps $\left\{ X_t \right\}_{t=1}^{N}$ and corresponding confidence map $\left\{ C_t \right\}_{t=1}^{N}$ for background reconstruction. 
The point map represents a dense, pixel-aligned 3D location map for each corresponding image in the world coordinate frame.
We stack all these point maps to represent the 3D geometry of the entire scene, although they include noisy points. 
Then, we compute the $\eta_{conf}$ percentile of the pixel-aligned confidence values and use it to filter the point maps, removing noisy dynamic regions and intricate geometries with low confidence. 
The filtered point maps are then stacked, and Poisson surface reconstruction \cite{kazhdan2006poisson} is applied to generate the background mesh $\mathcal{M}_{bg}$. 
Thanks to this filtering, the background mesh is generated as a clean, flat ground surface, free from noisy dynamic human regions or complex structures that would otherwise make it difficult to identify human-scene contact points.
\noindent\textbf{Scale Optimization.} Furthermore, to address the scale ambiguity issue of the 3D foundation model \cite{wang2025vggt}, we propose an alignment method between the background and the human mesh. 
Since the human mesh generated by SMPL \cite{loper2023smpl} is defined in metric scale, we can align the background mesh relative to the metric human mesh. 
We introduce a scale parameter, $\sigma$, which adjusts the point maps $X_t$, background mesh $\mathcal{M}_{bg}$, and camera poses accordingly. 

Our idea is to optimize the scale parameter from a \textit{bone length} perspective. 
Inspired by \cite{muller2024reconstructing}, we leverage the correlation between observed 2D keypoints and 3D joints for this alignment.
We first use ViTPose \cite{xu2022vitpose} to detect 2D keypoints $j^{k}$ for each person $k$ in the image. Then, we lift the 2D keypoints $j_t^{k}$ to 3D joints using scaled point maps $X_t$, defined as $\hat{J}^{k}_{t} = \sigma X_{t}(i,j)$ where $(i, j) \in j^{k}_{t}$. 
Given the human mesh parameters $\theta^k_t, \beta^k_t$ from HMR2.0 \cite{goel2023humans}, we can obtain the 3D location $J_{i}$ of interest joint $i$ from the mesh.   
Then we compute the 3D bone length $d^k_{(p,c)}$ for each joint pair $(p,c)$ associated with the main body joints $\mathcal{J}_{body}$  of person $k$ in the image at time $t$, as defined by: 
\begin{equation} \label{bonelength}
\hat{d}^{k}_{(p,c), t} = J_{p}(\theta^{k}_{t},\beta^{k}_{t}) - J_{c}(\theta^{k}_{t},\beta^{k}_{t}) \quad \text{and}\quad d^{k}_{(p,c), t}(\sigma) = \sigma X_{t}(p) - \sigma X_{t}(c).
\end{equation}

The loss function $L(\sigma)$ is computed as the L2 norm loss between the predicted 3D joint-based bone lengths $\hat{d}^{k}_{(p,c), t}$ and the corresponding ground-truth values $d^{k}_{(p,c), t}(\sigma)$. Specifically, the loss is calculated across all time steps $t$, individuals $k$, and joints $(p, c)$ in the main body, where minimizing this loss ensures that the predicted bone lengths align closely with the ground truth. The scale parameter $\sigma$ is optimized using the L-BFGS algorithm \cite{nocedal1980updatingLBFGS} applied to this loss.
The loss function for optimizing the scale parameter $\sigma$ is as follows:
\begin{equation} \label{bonelength_eq}
L(\sigma) = \sum_{t}\sum_{k \in \mathcal{K}}\sum_{(p,c) \in \mathcal{J}_{\scriptscriptstyle{body}}} ||\hat{d}^{k}_{(p,c), t} - d^{k}_{(p,c), t}(\sigma)||_{2}.
\end{equation}
The update rule is $ \sigma_{n+1} = \sigma_n - \alpha \mathcal{H}_{n} \nabla L(\sigma_n)$
where $\alpha$ is the learning rate and, $\mathcal{H}_{n}$ is the approximation of the inverse Hessian. The optimization proceeds for $T_{opt}$ steps, and the objective is minimized through gradient-based optimization. 

\noindent\textbf{Human Placement.} 
After determining the optimal scale $\sigma$, we apply this scale to the background mesh $M_{\text{bg}}$ and camera poses $P_t$ to align them with the human mesh.
In conjunction with the camera pose, it is necessary to initialize the root translation in the world coordinate system. 
Our approach takes advantage of the fact that UAV-captured images provide top-down views, which facilitate the identification of the ground contact point. 
This contact point serves as a plausible initial estimate for the root translation.
Given the background mesh $M_{\text{bg}}$, we render the depth map $D_t$ from the specified viewpoint. 
Using the 2D bounding box $B^k_t = (x_{\text{min}}, y_{\text{min}}, x_{\text{max}}, y_{\text{max}})$ projected from the human mesh, we detect the 2D contact point as $(x_c, y_c) = \left((x_{\text{min}} + x_{\text{max}})/2,\: y_{\text{max}}\right)$. 
Subsequently, we use the depth value from $D_{t}$ to unproject this 2D contact point for each individual human into the world coordinate system as follows: 
\begin{equation} \label{root_translation}
\psi_{t} = \sigma R^T[K^{-1}D_{x_c, y_c}[x_c, y_c, 1]^T] - \sigma R^TT .
\end{equation}
$\psi_{t}$ is the 3D ground contact point in the world coordinate. 

\begin{wrapfigure}{r}{0.6\linewidth}
    \centering
    \includegraphics[width=\linewidth]{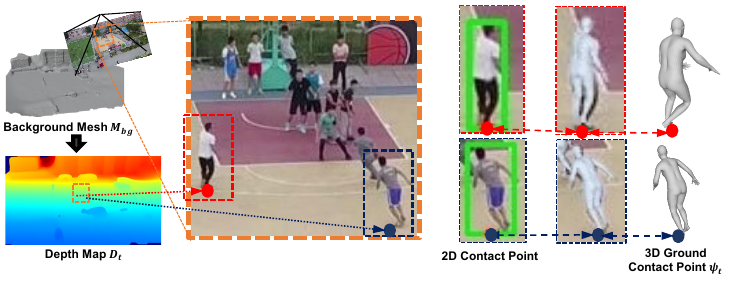}
    \vspace{-6mm}
    \caption{
    \textbf{Human Placement.} We identify the 2D contact point from the bounding box and use the depth from the mesh to unproject the 3D ground contact point.
    }
    \vspace{-3mm}
    \label{fig:human_contact}
\end{wrapfigure}
With this information, we can accurately position all individual human meshes within the background mesh, with all meshes defined in the world coordinate system, as seen in Fig. \ref{fig:human_contact}.

\subsection{3D Gaussian Scene Modeling}
\label{sec:rendering}
We model the scene using distinct Gaussian representations for dynamic humans and the static background, ensuring both temporal coherence and rendering clarity.
We employ standard 3D Gaussians to model the background Gaussians $g_{bg}$ and the human Gaussians $g_{h}$. Following Sec. \ref{sec:Preliminaries}, both Gaussians $g_{h}$ and $g_{bg}$ contain learnable attributes $(\mu, r, s, c, o)$, which represent the 3D mean, 3D rotation, scaling factor, opacity factor, and color value, respectively.
These are then merged into a unified human-scene primitive $g_{all} = g_{bg} + g_{h}$, which is subsequently fed into the Gaussian rasterizer \cite{kerbl20233d} for rendering.  

\noindent\textbf{Background Gaussians.} The background is represented by static Gaussians $g_{bg}$ , which are initialized using stacked multi-view point maps $X_{t}$. For UAV scenes captured at high altitudes, we use all point maps without filtering. In contrast, for UAV scenes captured at low altitudes, we employ a confidence map to filter the point maps and use human masks $M^k_{t}$ to remove point maps corresponding to human regions, stacking only the static background regions.

\noindent\textbf{Human Gaussians.} For human Gaussians, inspired by GART \cite{lei2024gart}, we utilize the canonical template mesh $\mathcal{M}_{h}$ in the rest pose to represent the 3D Gaussian splats. The skinning transformation for each Gaussian, $A(t) = (R_{A}(t), T_{A}(t))$, is derived from the LBS weight $LBS(g_{h}, \theta_{t})$, which transform the canonical Gaussian positions $\mu$ and rotations $r$ to the world frame.
\begin{equation} \label{LBS_operation}
g_{h}(t) = (R_{A}(t)\mu + T_{A}(t) + \psi_{t},\: R_{A}(t) \cdot r,\: s, c, o). 
\end{equation}  
The changes in $\theta_{t}$ over time result in updates to the transformations of the key joints and linearly interpolates Gaussians to obtain the deformed position $\mu(t)$ and rotations $r(t)$ at different time steps. 

\noindent\textbf{Optimization.} We optimize all Gaussian attributes $g_{bg}$, $g_{h}$, the human poses of all SMPL parameters $\theta_t, \psi_{t}$ for each frame $t$ and the corresponding skimming weights.  

The overall loss function for optimization is defined as:
\begin{equation} \label{eq:total_loss}
    L_{total} = (1-\lambda_{pho})L_{1} + \lambda_{pho}L_{ssim} + \lambda_{o}L_{o} + L_{smpl},
\end{equation}
where $L_{1}$ and $L_{ssim}$ represent the photometric loss between the rendered full image and the ground truth image.
$L_{smpl}$ consists of various loss terms related to human Gaussian splats and human regions. Details are provided in the supplementary material.

\section{Experimental Results}
\label{sec:experimental_resuts}

To measure the effectiveness of UAV4D, we evaluate our method on novel view synthesis tasks, using every 8th frame as the held-out test set. 
We report PSNR, SSIM \cite{wang2004image}, and LPIPS \cite{zhang2018unreasonable} for full images and human-related regions, to assess dynamic reconstruction capabilities.
The human regions are obtained using a pretrained segmentation model which get a prompt from groundtruth bonding box.  

\subsection{Datasets and Baselines}

\noindent\textbf{VisDrone:} The VisDrone dataset \cite{cao2021visdrone} is a standard benchmark for object detection and tracking in the UAV research community. It was captured at altitudes of 40 to 50 meters in China under varying weather and illumination conditions. 
This dataset includes many densely packed small objects in UAV images. 
We selected four different scenes that primarily consist of pedestrians and complex background environments.

\noindent\textbf{Okutama-Action:} The Okutama-Action dataset \cite{barekatain2017okutama} contains video sequences of 12 action categories, captured using cameras mounted on two flying UAVs at altitudes ranging from 10 to 45 meters. We chose four scenes that exhibit significant camera variation and distinct viewing angles.

\noindent\textbf{Manipal-UAV:} The Manipal-UAV dataset \cite{akshatha2023manipal} is designed for person detection in UAV images. It was acquired in India using two UAVs flying at altitudes between 10 and 50 meters. Unlike the previous datasets, it explicitly provides altitude information. We selected four scenes captured at 40 and 50 meters, representing the highest-altitude cases.



\noindent\textbf{Baselines:} To evaluate rendering quality, we compare our method with four comparative methods: TK-Planes \cite{maxey2024tk}, 3DGS \cite{kerbl20233d}, Deformable-3DGS \cite{yang2024deformable}, and 4DGS \cite{wu20244d}. 
To the best of our knowledge, TK-Planes \cite{maxey2024tk} is the only work that has specifically designed dynamic NeRFs for UAV scenes, while 4DGS \cite{wu20244d} and Deformable-3DGS \cite{yang2024deformable} represent the state-of-the-art methods in dynamic Gaussian-splatting. 

\subsection{Quantitative Comparison}
\begin{table}[t]
    \centering
    \caption{
    \textbf{Quantitative Comparison} of our method with recent comparative works on the three datasets. 
    Our method achieves the best performance across every category except for PSNR on the Manipal-UAV dataset. 
    Please see the Appendix for detailed comparisons on the four scenes in each dataset.
    Red, orange, and yellow indicate the first, second, and third best performing algorithms for each metric, respectively. 
    }
    \begin{adjustbox}{width=0.99\linewidth,center}
    \begin{tabular}{l|ccc|ccc|ccc}
    \toprule
    \multirow{2}{*}{Method}&\multicolumn{3}{c|}{Okutama-Action}&\multicolumn{3}{c|}{Manipal-UAV}&\multicolumn{3}{c}{VisDrone}\\
    & PSNR \(\uparrow\)  & SSIM \(\uparrow\)  & LPIPS \(\downarrow\) & PSNR \(\uparrow\)  & SSIM \(\uparrow\)  & LPIPS \(\downarrow\) & PSNR \(\uparrow\)  & SSIM \(\uparrow\)  & LPIPS \(\downarrow\) \\
    \midrule
TK-Planes \cite{maxey2024tk}           &  \cellcolor{tabthird}28.05 &  \cellcolor{tabthird}0.796 & 0.417 & 27.98 & 0.718 & 0.396 &  \cellcolor{tabthird}23.98 &  \cellcolor{tabthird}0.644 & 0.450 \\
3DGS \cite{kerbl20233d}                & \cellcolor{tabsecond}29.90 & \cellcolor{tabsecond}0.867 & \cellcolor{tabsecond}0.228 & 29.46 &  \cellcolor{tabthird}0.856 &  \cellcolor{tabthird}0.150 & \cellcolor{tabsecond}24.59 & \cellcolor{tabsecond}0.782 & \cellcolor{tabsecond}0.220 \\
4DGS \cite{wu20244d}                    & 25.90 & 0.780 &  \cellcolor{tabthird}0.410 &  \cellcolor{tabthird}30.87 & 0.843 & 0.253 & 22.58 & 0.628 &  \cellcolor{tabthird}0.440 \\
DeformableGS \cite{yang2024deformable} & 22.64 & 0.748 & 0.443 &  \cellcolor{tabfirst}31.22 & \cellcolor{tabsecond}0.894 & \cellcolor{tabsecond}0.114 & 15.46 & 0.522 & 0.614 \\
UAV4D (Ours) &  \cellcolor{tabfirst}30.36 &  \cellcolor{tabfirst}0.875 &  \cellcolor{tabfirst}0.184 & \cellcolor{tabsecond}30.94 &  \cellcolor{tabfirst}0.897 &  \cellcolor{tabfirst}0.084 &  \cellcolor{tabfirst}26.03 &  \cellcolor{tabfirst}0.800 &  \cellcolor{tabfirst}0.147 \\
    \bottomrule
    \end{tabular}
    \end{adjustbox}
    \label{tab:whole-region}
    \vspace{-5mm}
\end{table}

\begin{table}[t]
    \centering
    \caption{
    \textbf{Quantitative Comparison} of our method with recent comparative works is conducted on the three datasets, focusing on \textbf{human-only} regions cropped using precise human masks. Due to the small size of the human regions, it is not possible to compute the LPIPS metric. Consequently, we report only the PSNR and SSIM metrics.
    Please see the Appendix for detailed comparisons on the four scenes in each dataset.
    Red, orange, and yellow indicate the first, second, and third best performing algorithms for each metric, respectively. 
    }
    \begin{adjustbox}{width=0.7\linewidth,center}
    \begin{tabular}{l|cc|cc|cc}
    \toprule
    \multirow{2}{*}{Method}&\multicolumn{2}{c|}{Okutama-Action}&\multicolumn{2}{c|}{Manipal-UAV}&\multicolumn{2}{c}{VisDrone}\\
    & PSNR \(\uparrow\)  & SSIM \(\uparrow\)  & PSNR \(\uparrow\)  & SSIM \(\uparrow\)  & PSNR \(\uparrow\)  & SSIM \(\uparrow\)    \\
    \midrule
TK-Planes \cite{maxey2024tk}           & \cellcolor{tabsecond}19.14 &  \cellcolor{tabthird}0.631 & 18.38 & 0.752 & \cellcolor{tabsecond}17.65 &  \cellcolor{tabthird}0.544 \\
3DGS \cite{kerbl20233d}                &  \cellcolor{tabthird}19.07 & \cellcolor{tabsecond}0.656 & 17.99 & 0.748 &  \cellcolor{tabthird}16.51 & \cellcolor{tabsecond}0.548 \\
4DGS \cite{wu20244d}                    & 18.18 & 0.602 & \cellcolor{tabsecond}19.23 & \cellcolor{tabsecond}0.770 & 16.25 & 0.510 \\
DeformableGS \cite{yang2024deformable} & 16.33 & 0.571 &  \cellcolor{tabthird}18.65 &  \cellcolor{tabthird}0.762 & 12.59 & 0.354 \\
UAV4D (Ours)                            &  \cellcolor{tabfirst}19.49 &  \cellcolor{tabfirst}0.664 &  \cellcolor{tabfirst}19.38 &  \cellcolor{tabfirst}0.784 &  \cellcolor{tabfirst}17.98 &  \cellcolor{tabfirst}0.604 \\
    \bottomrule
    \end{tabular}
    \end{adjustbox}
    \label{tab:human-region}
    \vspace{-4mm}
\end{table}

In Table \ref{tab:whole-region}, we present the rendering quality of novel-view synthesis across three datasets. 
We compute the average rendering metrics across 4 scenes for VisDrone \cite{cao2021visdrone}, 4 scenes for Manipal-UAV \cite{akshatha2023manipal}, and 4 scenes for Okutama-Action \cite{barekatain2017okutama}.
Due to the small size of the human subjects in UAV datasets, the rendering metrics for the full image do not reflect the dynamic rendering quality of the human regions directly. 
Moreover, since the Manipal-UAV dataset was captured at an exceptionally high altitude, the rendering metrics of the full image are less informative regarding the human regions and instead provide a better indication of the background region's rendering quality.
In contrast, the VisDrone dataset contains multiple humans, with relatively larger pixel regions compared to other datasets, allowing the full image rendering quality to be more directly correlated with the holistic scene rendering quality. 
Furthermore, certain scenes in VisDrone feature a significant number of moving pedestrians (approximately 30–50 individuals), presenting a particularly challenging scenario for dynamic reconstruction.
Our method demonstrates superior rendering quality compared to other recent approaches, underscoring its robustness. 
In the Okutama-Action dataset, our method also exhibits better rendering quality than alternative methods. 
The Okutama-Action dataset is characterized by substantial camera variation and distinct viewing angles, which can destabilize deformation-field-based methods. However, our method shows strong resilience in handling these dynamic scenes, which involve substantial pose variation and changes in viewing angle.


\begin{figure}[t]
\captionsetup[subfigure]{labelformat=empty}
\centering
\makebox[3pt]{\raisebox{60pt}{\rotatebox[origin=c]{90}{\hspace{2.4em} \tiny uav0084 \hspace{6em} \tiny uav0079}}}
\begin{subfigure}{0.192\linewidth} 
  \centering
\includegraphics[width=1\linewidth]{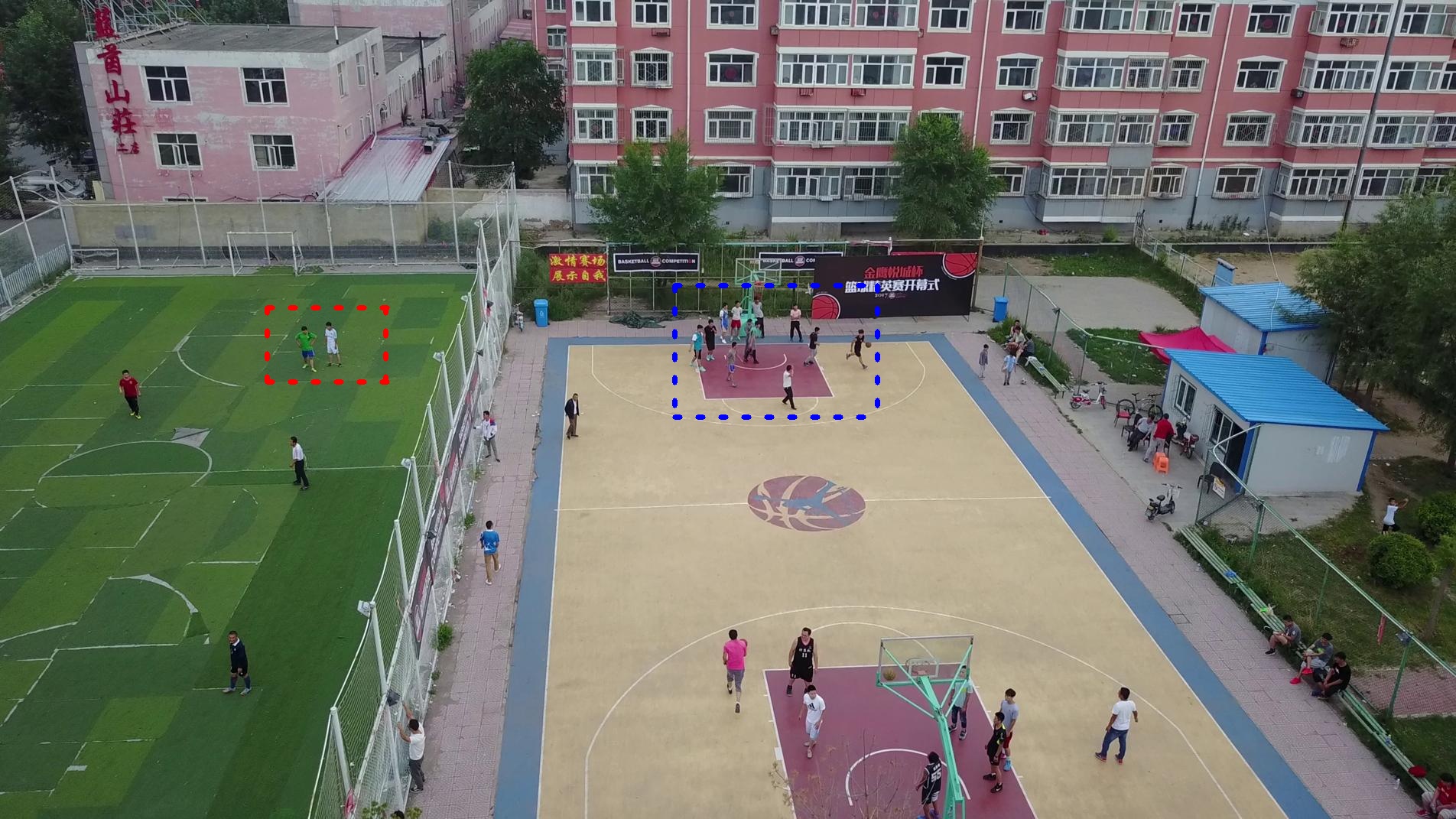} \\[0.5ex]
\includegraphics[width=0.485\linewidth]{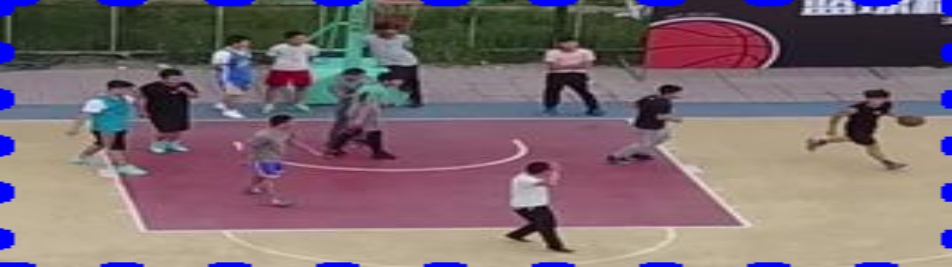} 
\hfill
\includegraphics[width=0.485\linewidth]{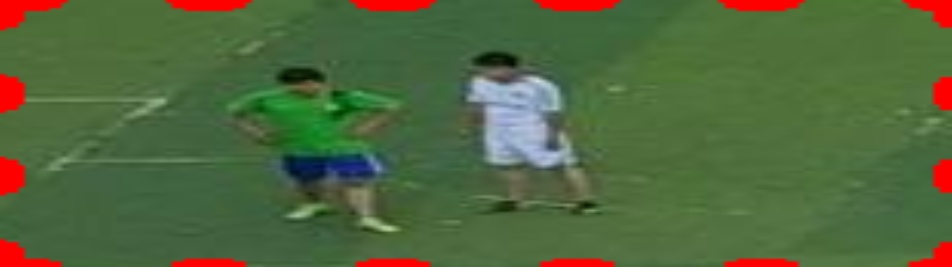} \\[0.5ex]
\includegraphics[width=1\linewidth]{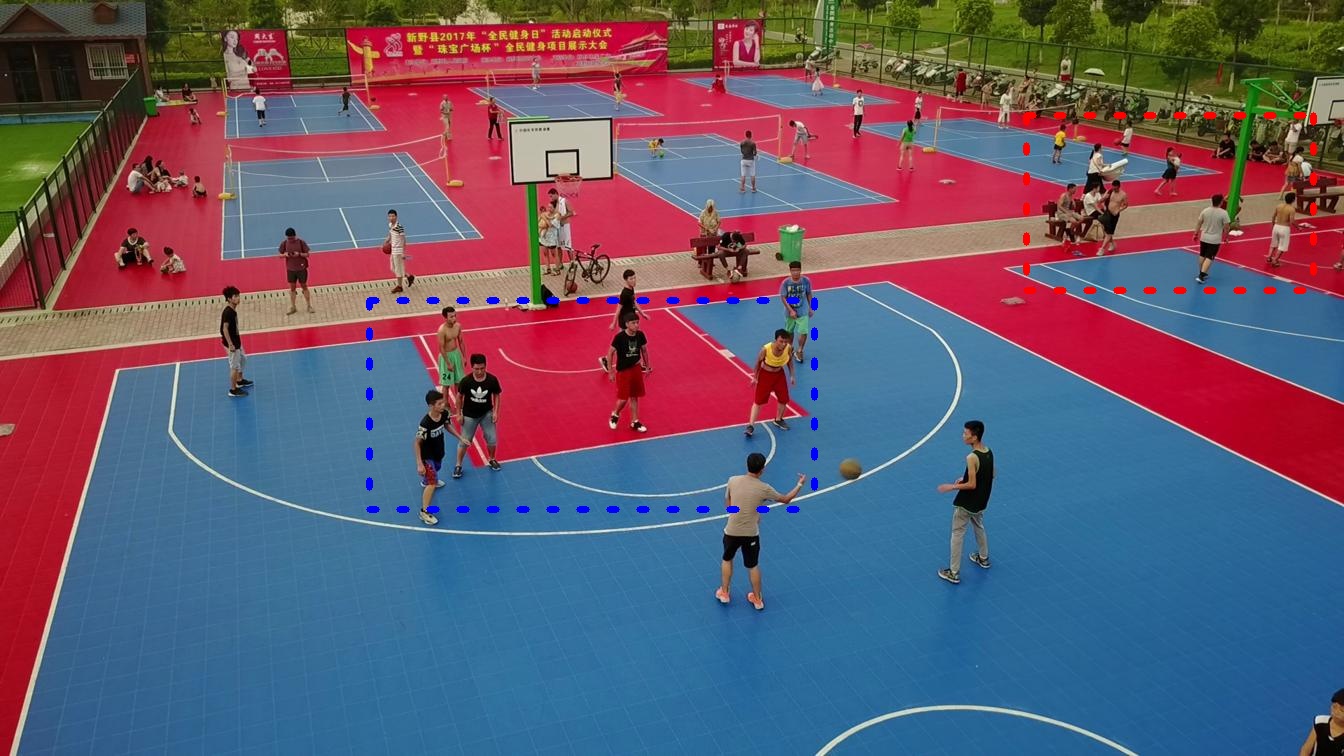} \\[0.5ex]
\includegraphics[width=0.485\linewidth]{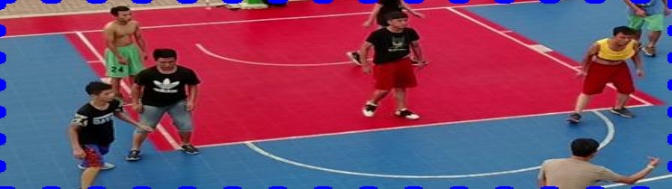} 
\hfill
\includegraphics[width=0.485\linewidth]{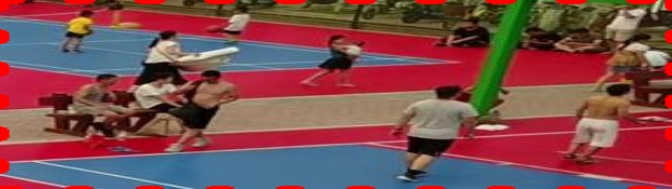} \\
  \vspace{-2mm}
  \caption{Scene}
\end{subfigure}
\begin{subfigure}{0.192\linewidth} 
  \centering
\includegraphics[width=1\linewidth]{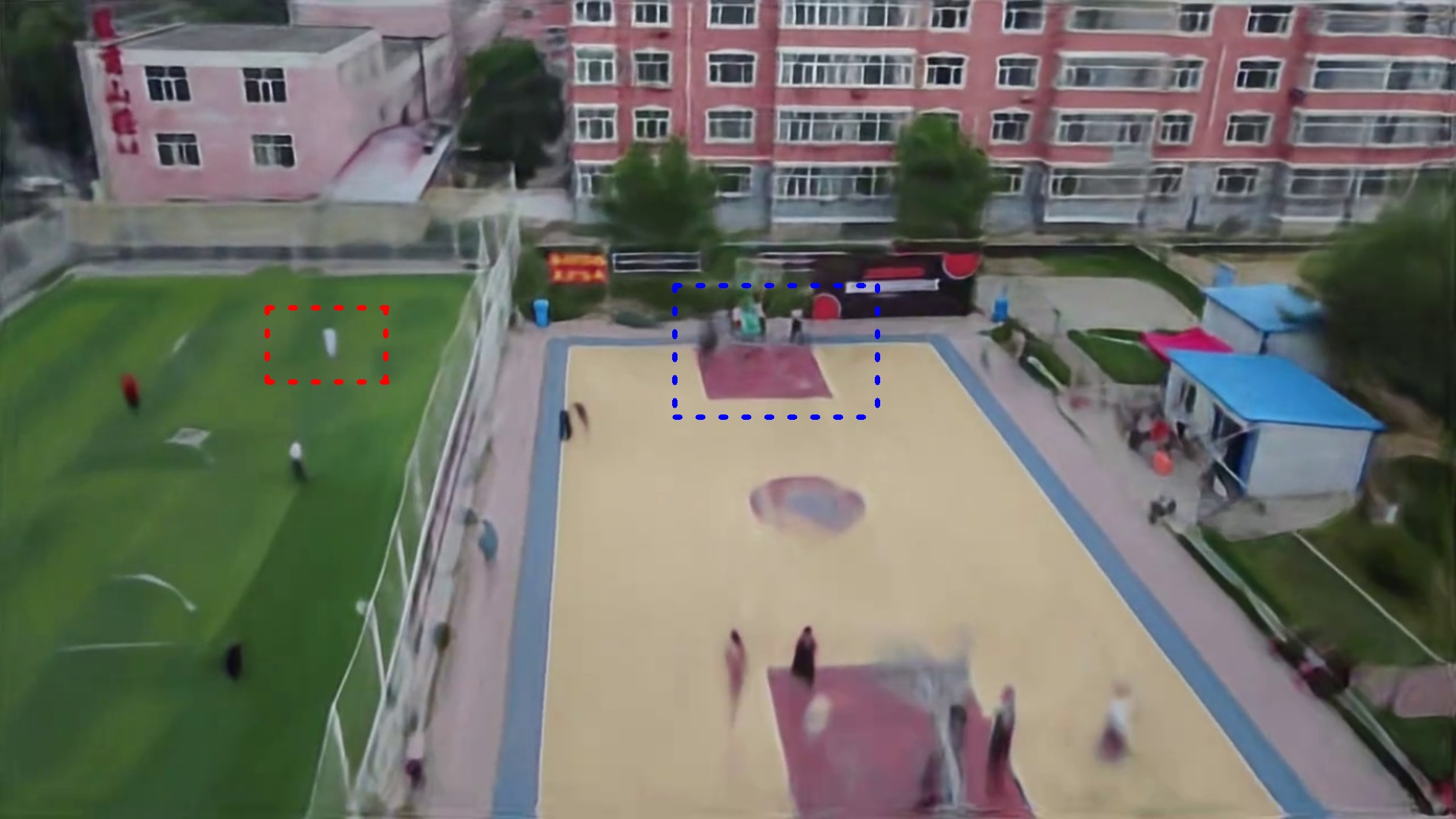} \\[0.5ex]
\includegraphics[width=0.485\linewidth]{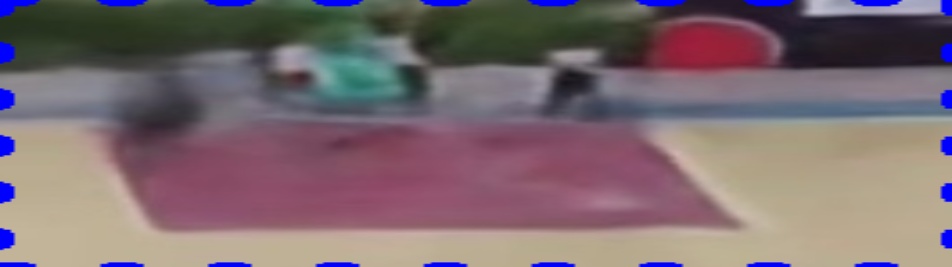} 
\hfill
\includegraphics[width=0.485\linewidth]{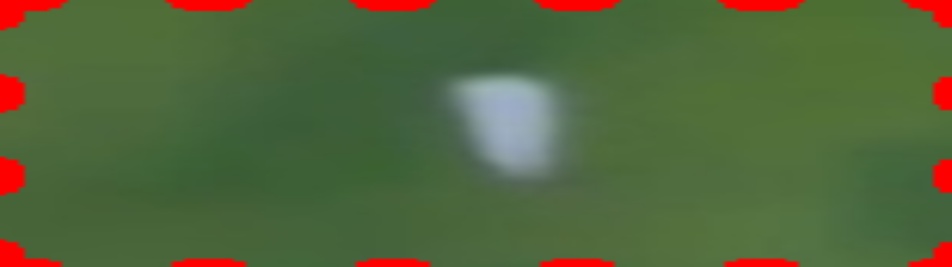} \\[0.5ex]
\includegraphics[width=1\linewidth]{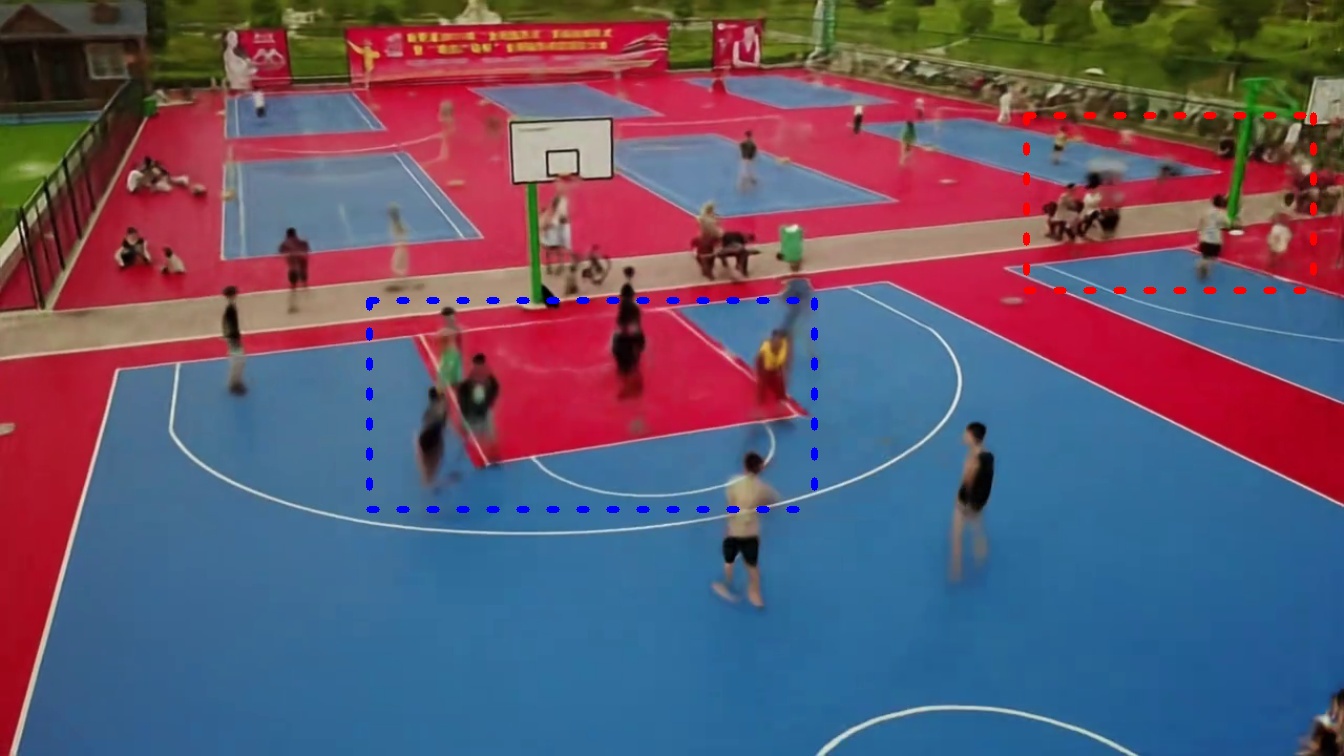} \\[0.5ex]
\includegraphics[width=0.485\linewidth]{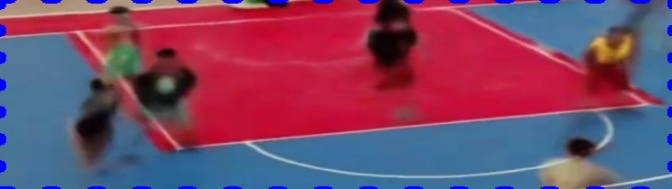} 
\hfill
\includegraphics[width=0.485\linewidth]{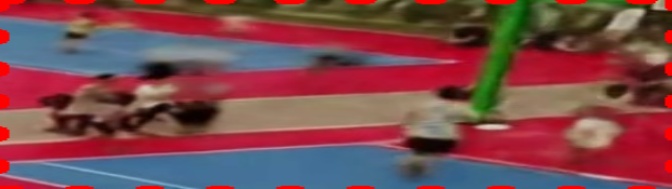} \\
  \vspace{-2mm}
  \caption{TK-Planes \cite{maxey2024tk}}
\end{subfigure}
\begin{subfigure}{0.192\linewidth} 
  \centering
\includegraphics[width=1\linewidth]{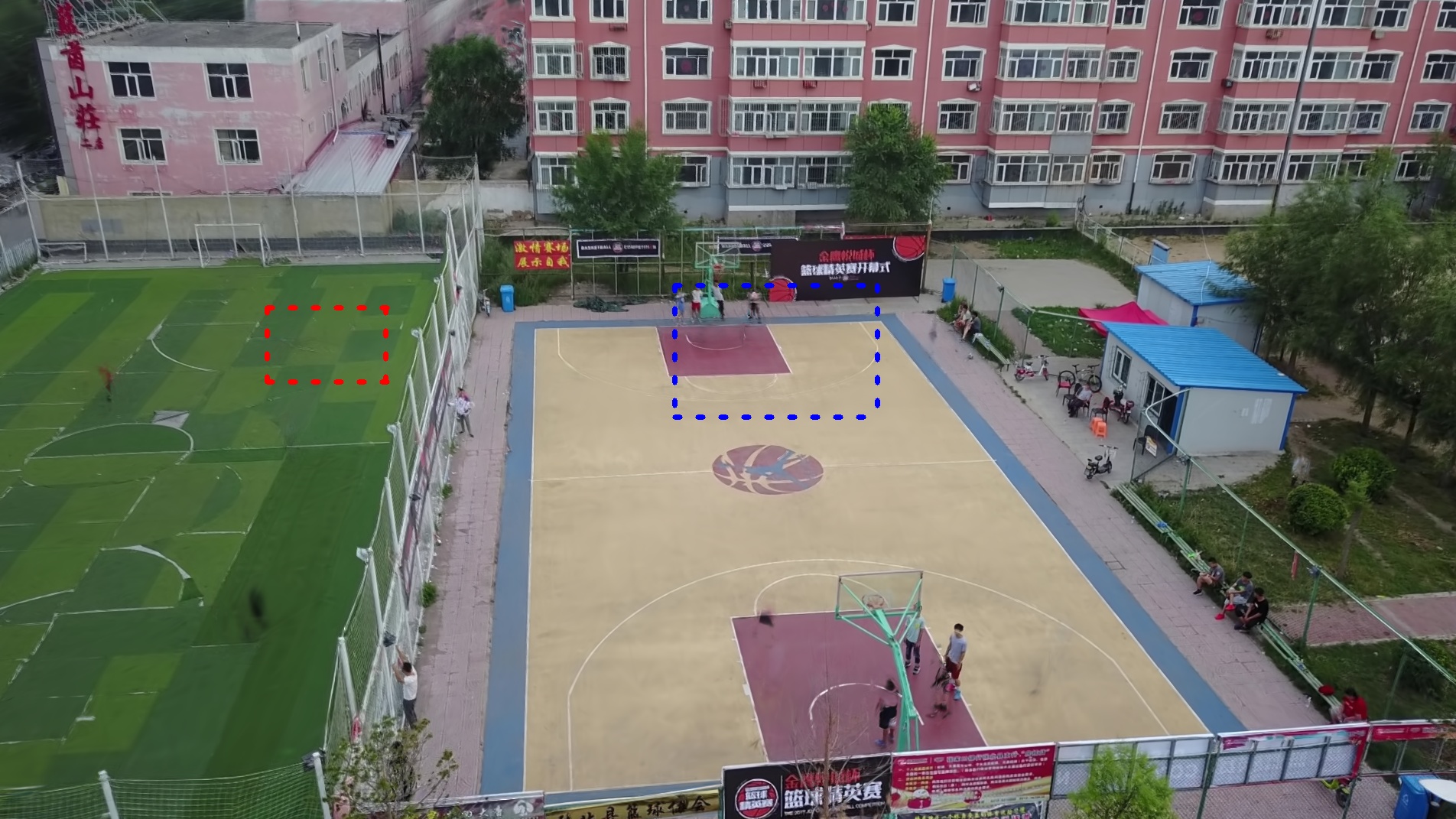} \\[0.5ex]
\includegraphics[width=0.485\linewidth]{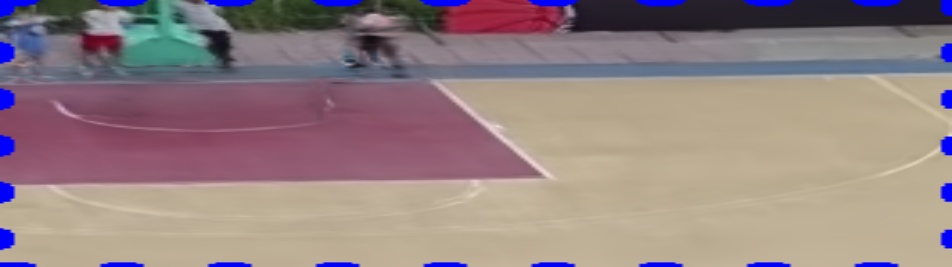} 
\hfill
\includegraphics[width=0.485\linewidth]{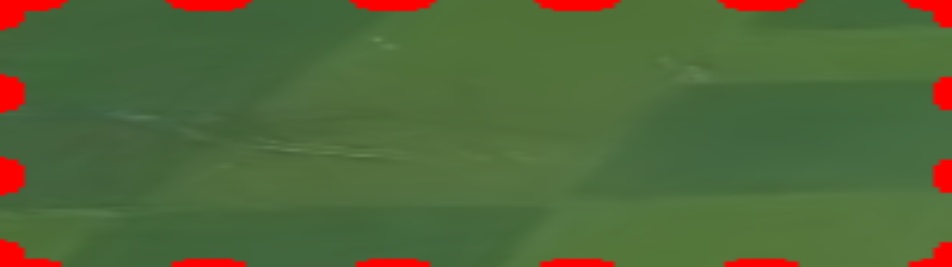} \\[0.5ex]
\includegraphics[width=1\linewidth]{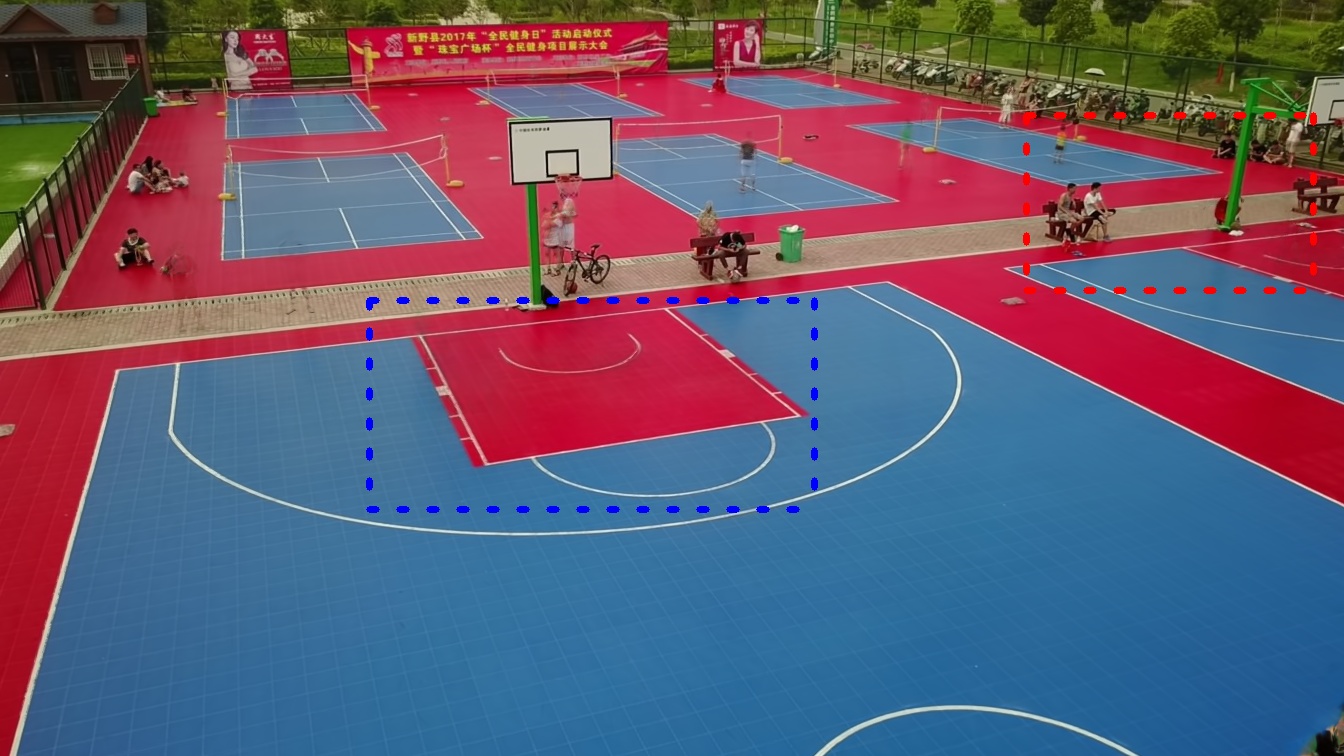} \\[0.5ex]
\includegraphics[width=0.485\linewidth]{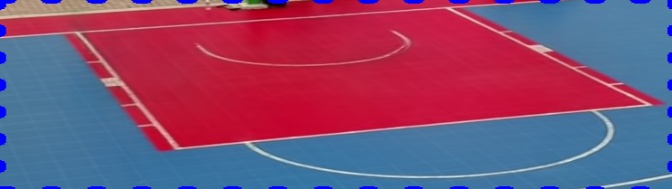} 
\hfill
\includegraphics[width=0.485\linewidth]{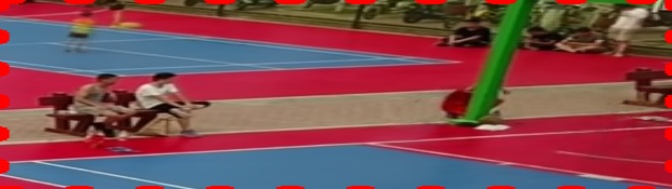} \\
  \vspace{-2mm}
  \caption{3DGS \cite{kerbl20233d}}
\end{subfigure}
\begin{subfigure}{0.192\linewidth} 
  \centering
\includegraphics[width=1\linewidth]{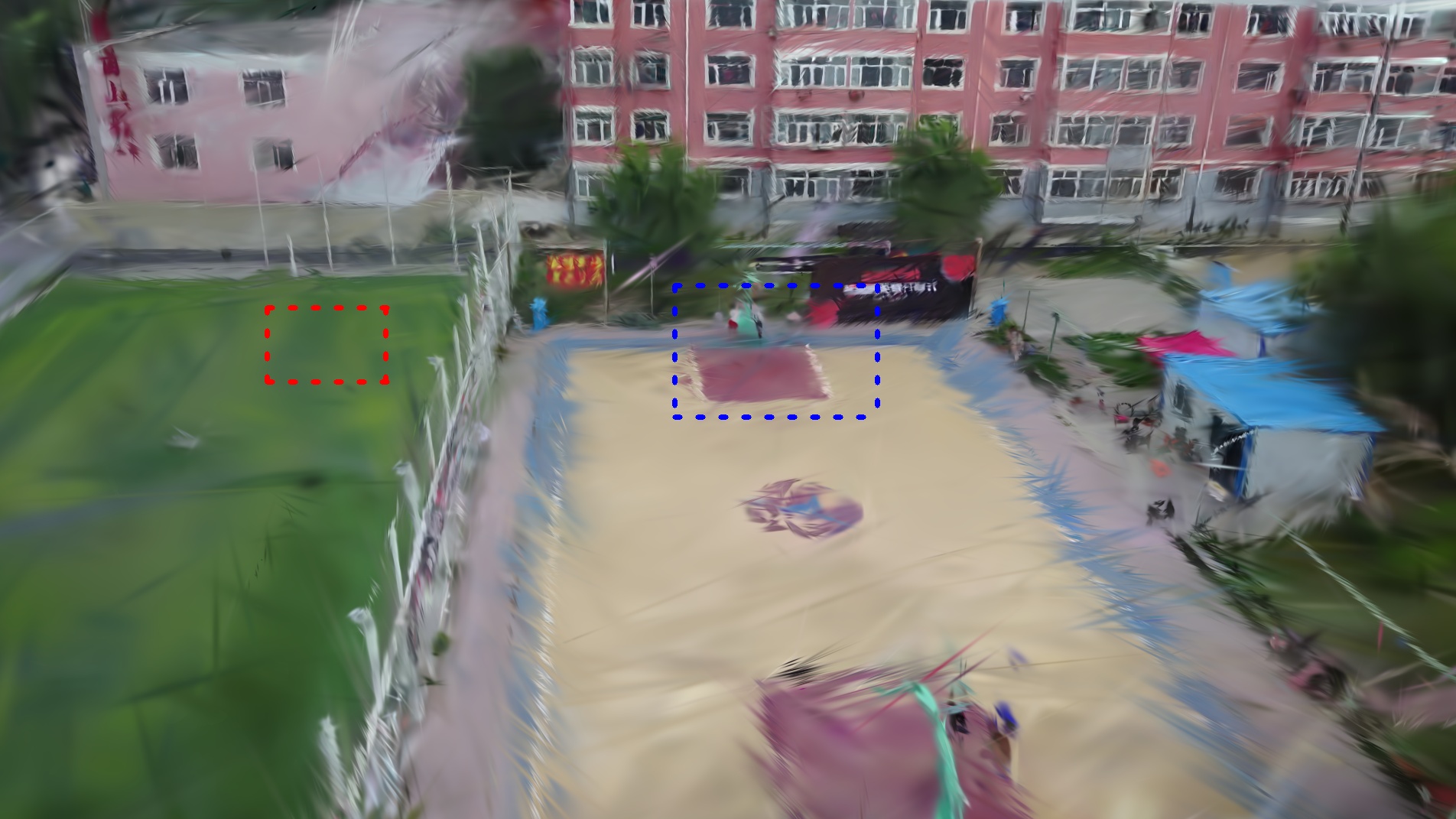} \\[0.5ex]
\includegraphics[width=0.485\linewidth]{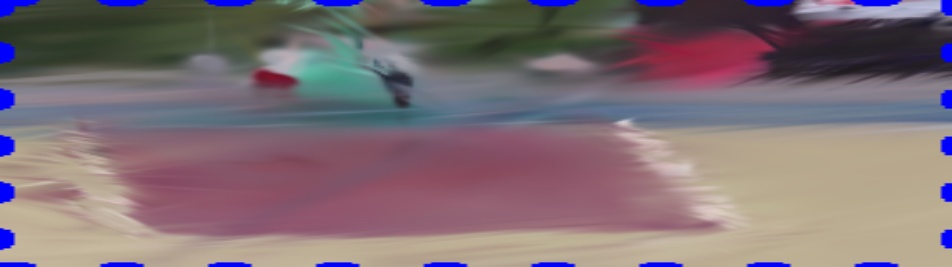} 
\hfill
\includegraphics[width=0.485\linewidth]{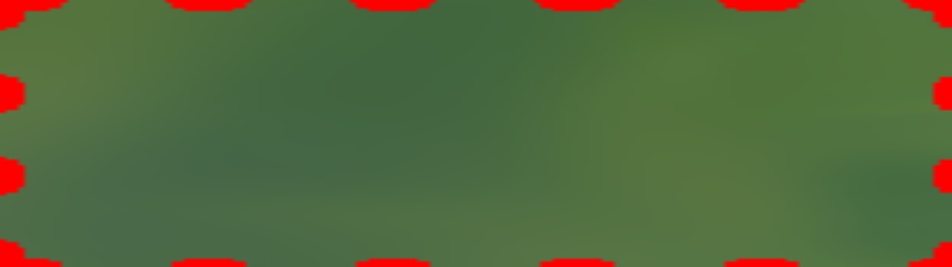} \\[0.5ex]
\includegraphics[width=1\linewidth]{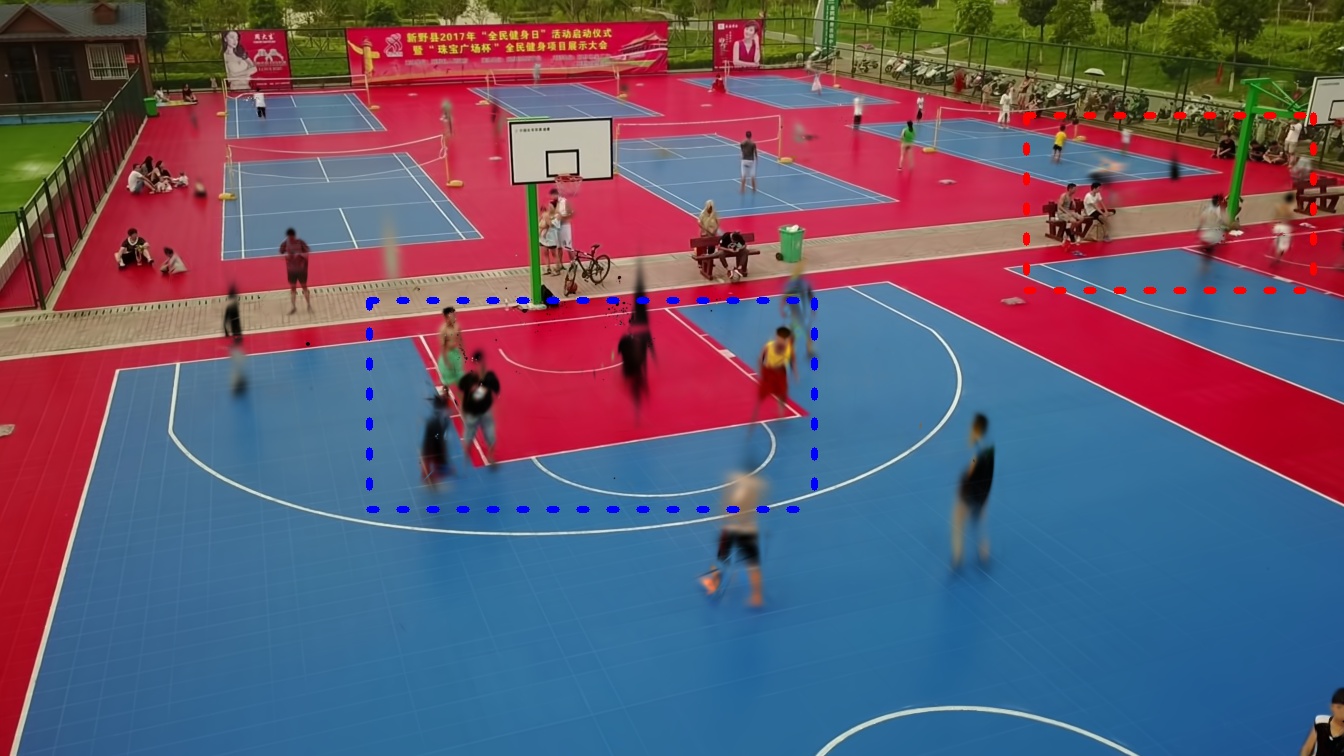} \\[0.5ex]
\includegraphics[width=0.485\linewidth]{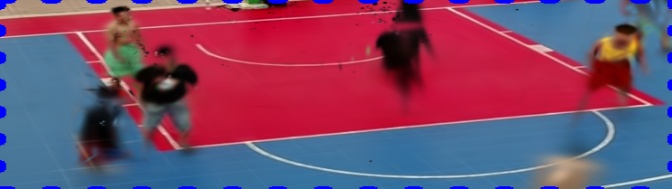} 
\hfill
\includegraphics[width=0.485\linewidth]{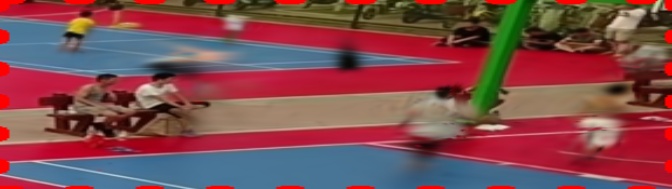} \\
  \vspace{-2mm}
  \caption{4DGS \cite{wu20244d}}
\end{subfigure}
\begin{subfigure}{0.192\linewidth} 
  \centering
\includegraphics[width=1\linewidth]{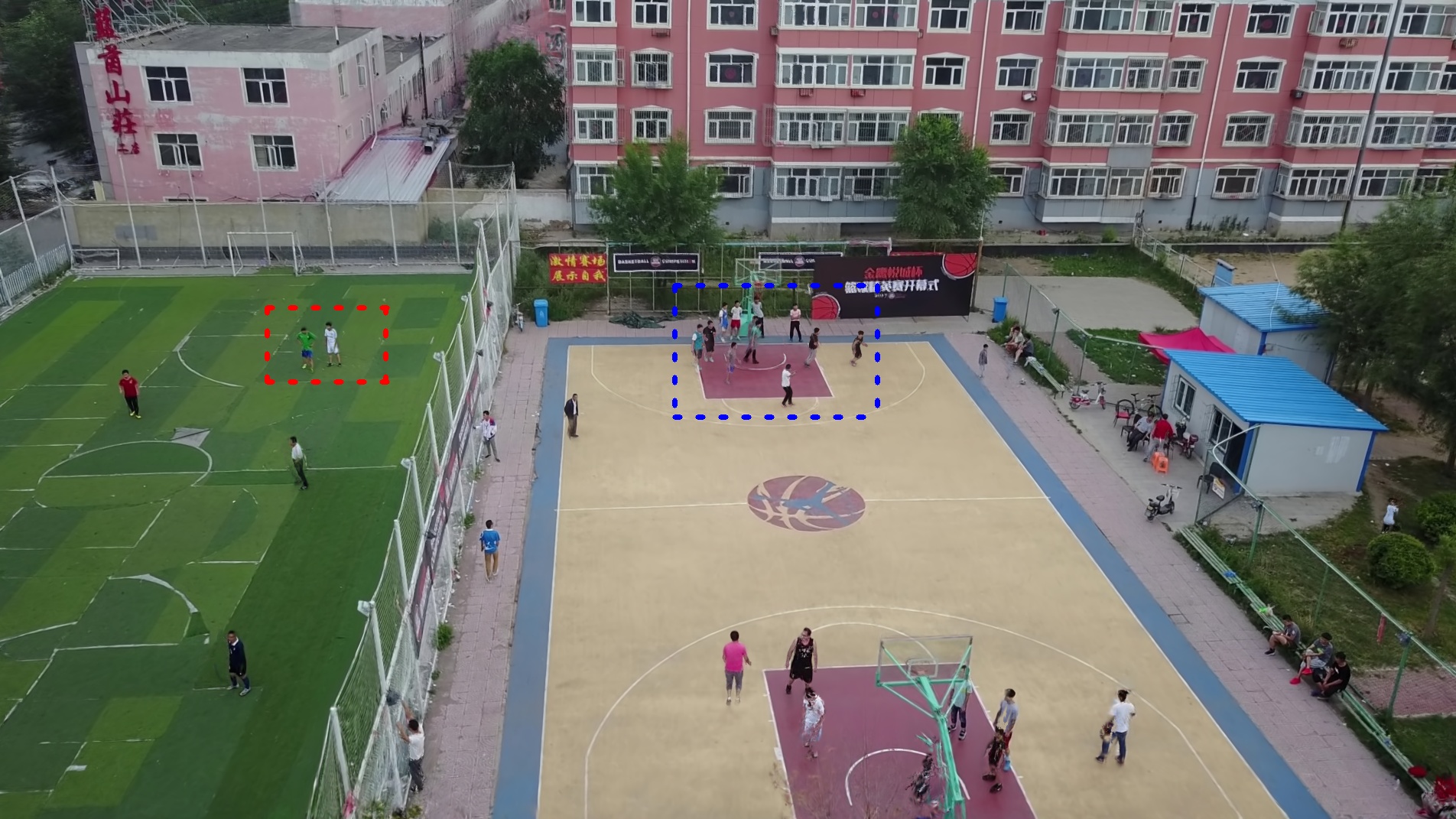} \\[0.5ex]
\includegraphics[width=0.485\linewidth]{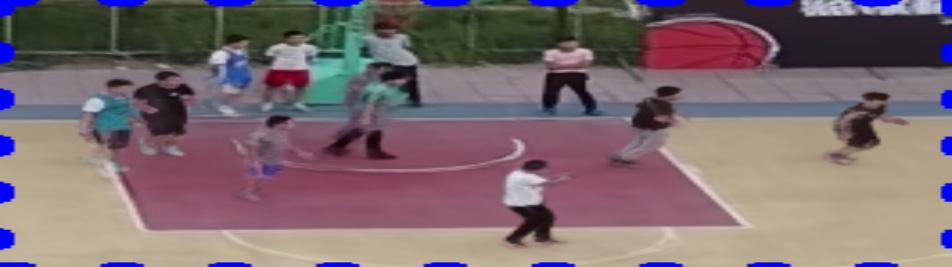} 
\hfill
\includegraphics[width=0.485\linewidth]{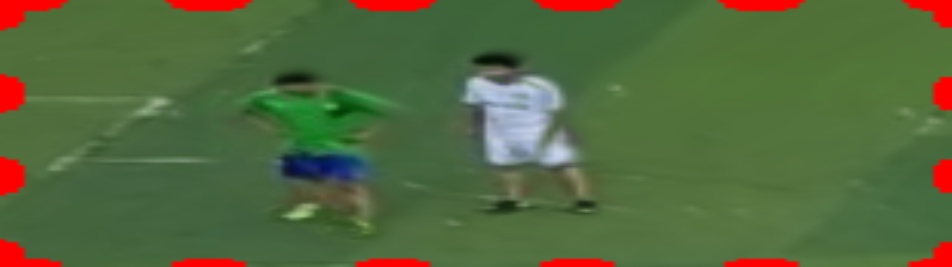} \\[0.5ex]
\includegraphics[width=1\linewidth]{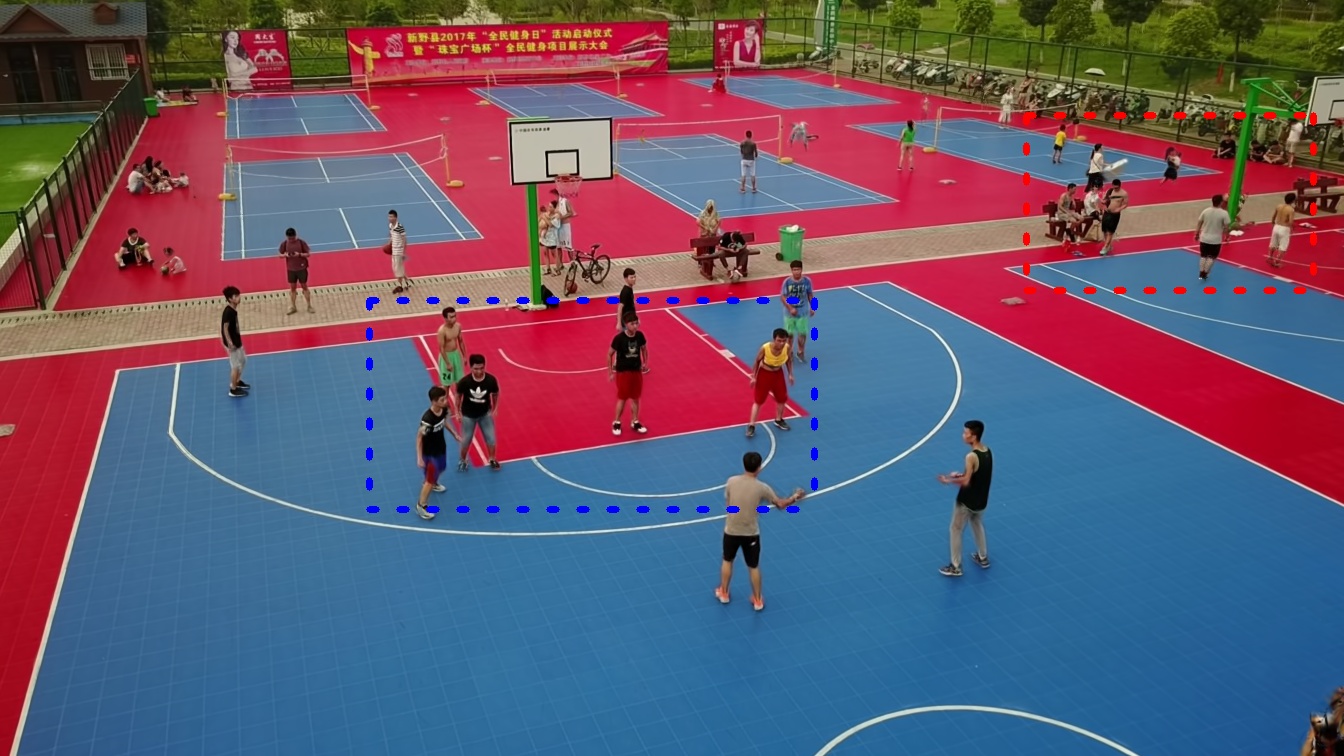} \\[0.5ex]
\includegraphics[width=0.485\linewidth]{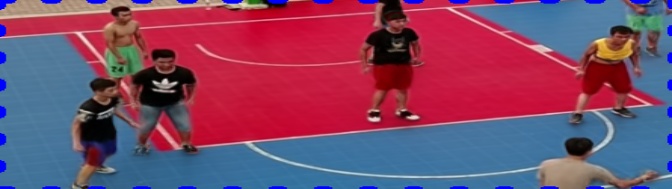} 
\hfill
\includegraphics[width=0.485\linewidth]{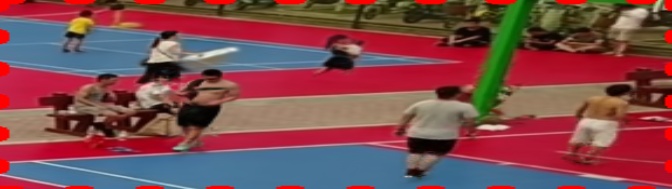} \\
  \vspace{-2mm}
  \caption{UAV4D (Ours)}
\end{subfigure}
\vspace{-6mm}
\caption{
\textbf{Quantitative Comparison of VisDrone Dataset \cite{cao2021visdrone}.} 
We visualize the zoomed-in blue and red regions, which emphasize the dynamic humans. Our method demonstrates superior capability in reconstructing small, moving humans compared to other existing approaches.
} 
\vspace{-2mm}
\label{fig:visdrone}
\end{figure}

In the VisDrone and Okutama-Action datasets, we observe that the training of 4DGS \cite{wu20244d} and Deformable-GS \cite{yang2024deformable} is unstable, with the training of deformation-based fields often failing. 
We attribute this instability to the presence of multiple small moving objects within the scenes, which indicates that a single deformation field is insufficient to capture the required deformations accurately. 
As a result, the training process for these models becomes highly unstable. 
In contrast, TK-planes \cite{maxey2024tk} demonstrates greater robustness across various dynamic scenes, yielding superior performance.
For the Manipal-UAV dataset, our method outperforms others in most cases, except for the PSNR values for Deformable-GS \cite{yang2024deformable}. We observe that Deformable-GS \cite{yang2024deformable} delivers superior rendering performance in regions with dense foliage and moving vegetation (e.g., swaying leaves). 
Our method currently struggles to represent such deformations, particularly in swaying trees, which cover large areas in UAV-captured datasets. 

\begin{wraptable}{r}{0.4\textwidth}
    \centering
    \caption{
    \textbf{Ablation Study} on VisDrone Dataset. 
    Please see the Appendix for details.
    }
    \begin{adjustbox}{width=1\linewidth,center}
    \begin{tabular}{l|ccc}
    \toprule
   Method & PSNR \(\uparrow\)  & SSIM \(\uparrow\)  & LPIPS \(\downarrow\) \\
    \midrule
Ours & 26.03 & 0.800 & 0.147  \\
\hline
Ours wo SMPL & 24.33 & 0.763 & 0.212 \\
Ours wo Scale & 23.02 & 0.651 & 0.25  \\
Ours wo Refine & 25.79 & 0.772 & 0.165 \\
    \bottomrule
    \end{tabular}
    \end{adjustbox}
    \label{tab:ablation}
\end{wraptable}

Table \ref{tab:human-region} reports the rendering quality across three datasets, focusing on human-only regions cropped using precise human masks.
Due to the small pixel size of the human regions, we are unable to use the LPIPS \cite{zhang2018unreasonable} metric, which relies on pretrained neural networks to measure metrics. 
Across all datasets, our method demonstrates superior rendering quality for human regions compared to other approaches. 
This indicates that our strategy effectively reconstructs and renders dynamic humans.

\noindent\textbf{Ablation Study}
Table \ref{tab:ablation} presents the results of an ablation study, averaged across four scenes in the Visdeon dataset.
``Ours wo SMPL'' indicates the version where we did not use human Gaussian splats, thus reconstructing only the background splats, which limits the model to static humans and removes moving people.
"Ours wo Scale" refers to the version where the human mesh is placed without scale optimization. 
In this case, we set the initial scale $\sigma$ to 40 and place the human mesh without any scale adjustment, leading to misalignment between the human regions and the human mesh.
"Ours wo Refine" follows the full pipeline but omits the human mesh refinement process described in Sec. \ref{sec:Initialization}. This version shows a slight improvement in artifacts caused by inaccurate in the human mesh.
\begin{figure}[t]
\captionsetup[subfigure]{labelformat=empty}
\centering
\makebox[3pt]{\raisebox{60pt}{\rotatebox[origin=c]{90}{\hspace{2.4em} \tiny{1\_2\_2} \hspace{6.5em} \tiny{1\_2\_8}}}}
\begin{subfigure}{0.192\linewidth} 
  \centering
\includegraphics[width=1\linewidth]{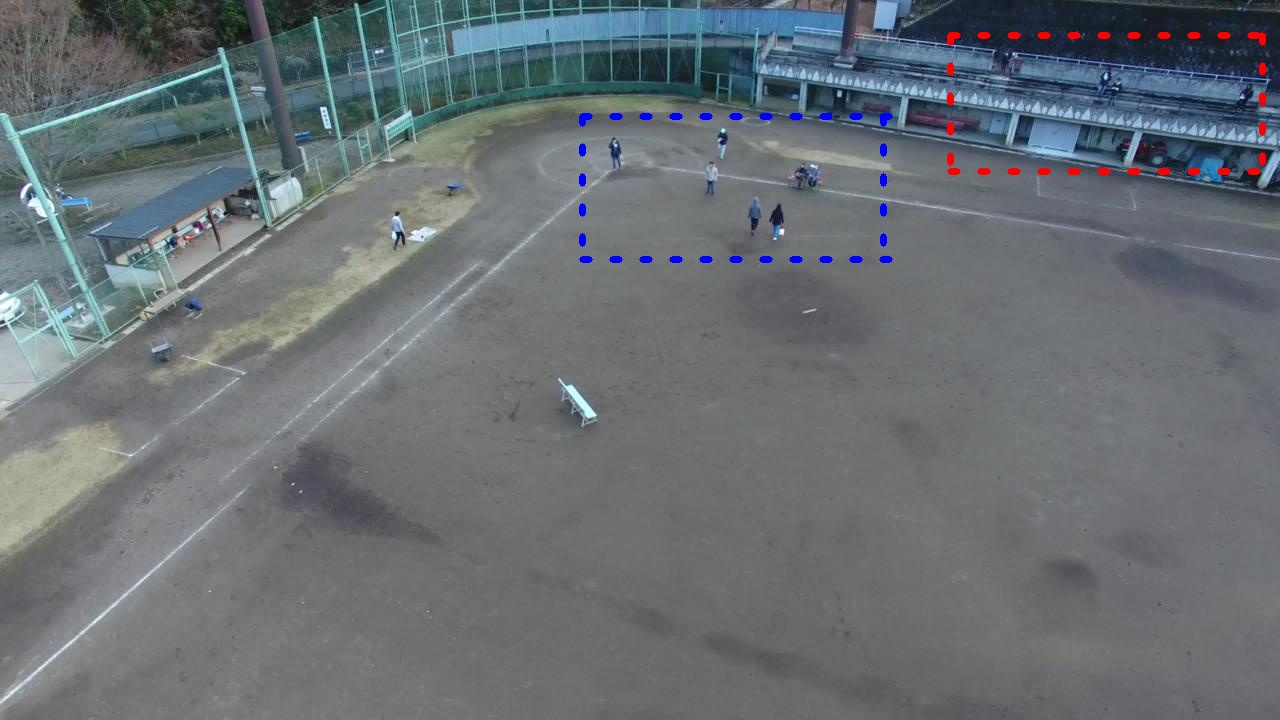} \\[0.5ex]
\includegraphics[width=0.485\linewidth]{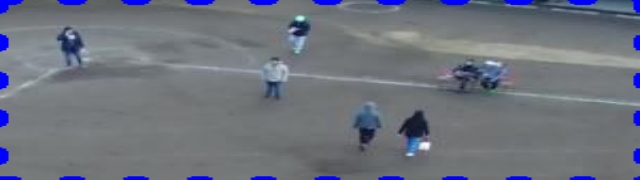} 
\hfill
\includegraphics[width=0.485\linewidth]{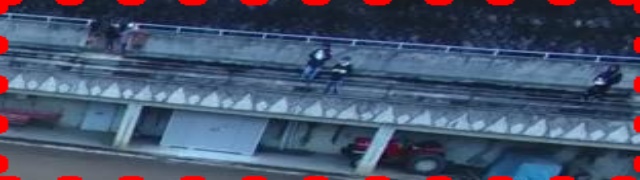} \\[0.5ex]
\includegraphics[width=1\linewidth]{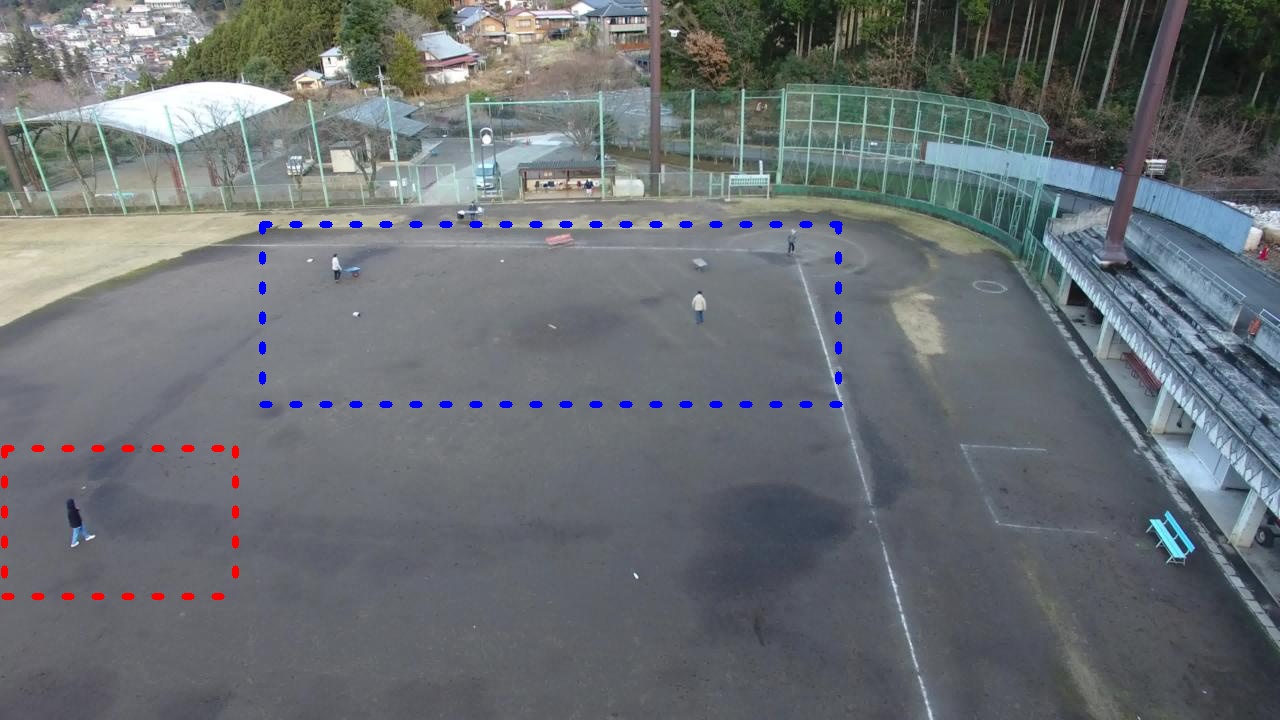} \\[0.5ex]
\includegraphics[width=0.485\linewidth]{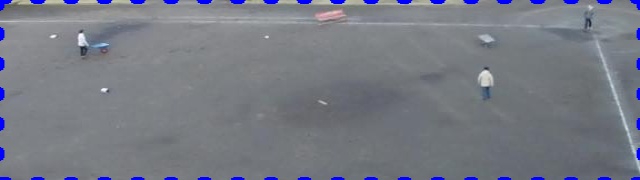} 
\hfill
\includegraphics[width=0.485\linewidth]{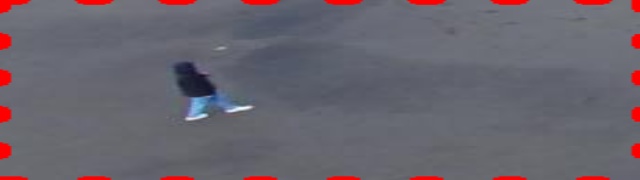} \\
  \vspace{-2mm}
  \caption{Scene}
\end{subfigure}
\begin{subfigure}{0.192\linewidth} 
  \centering
\includegraphics[width=1\linewidth]{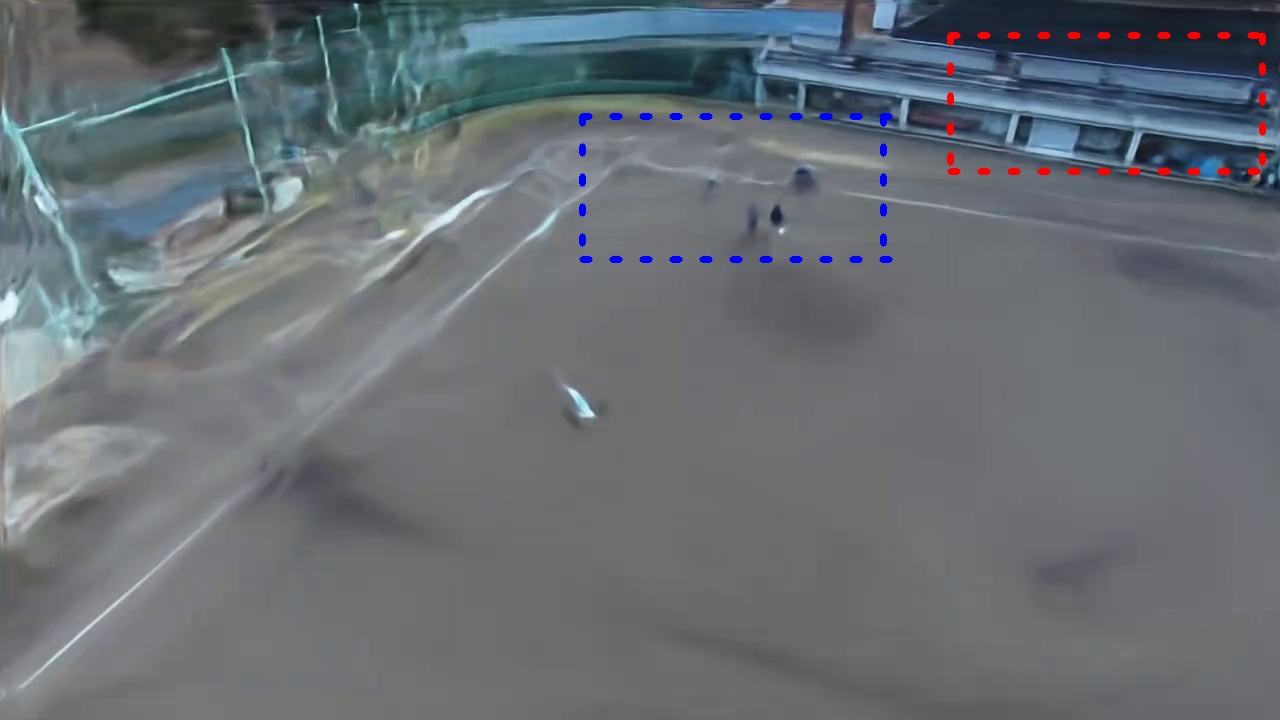} \\[0.5ex]
\includegraphics[width=0.485\linewidth]{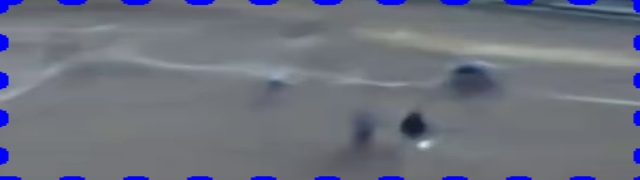} 
\hfill
\includegraphics[width=0.485\linewidth]{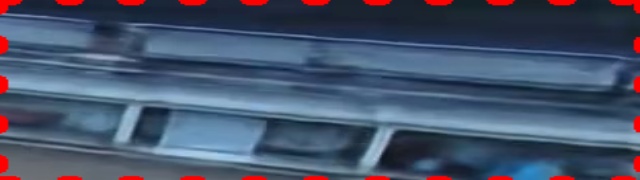} \\[0.5ex]
\includegraphics[width=1\linewidth]{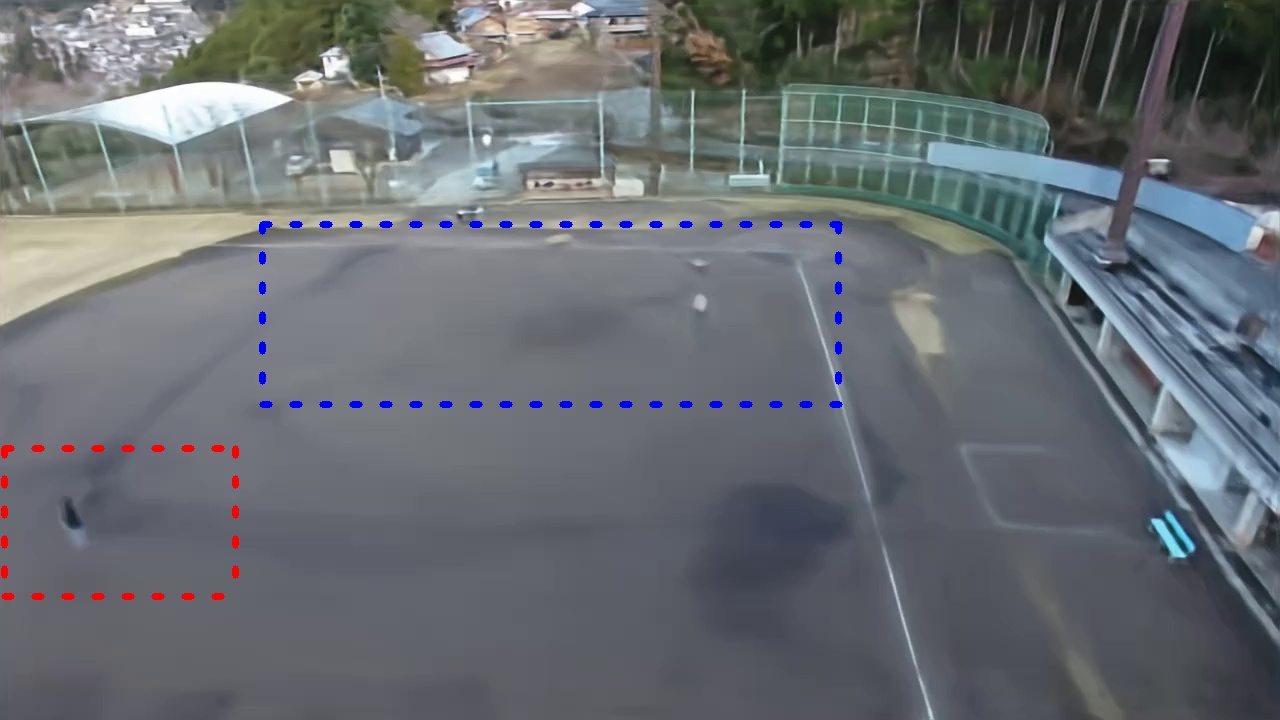} \\[0.5ex]
\includegraphics[width=0.485\linewidth]{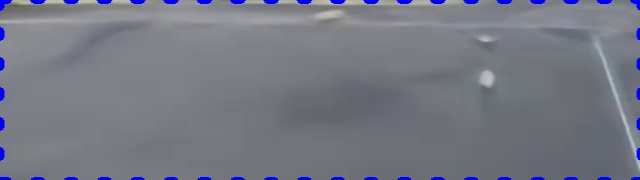} 
\hfill
\includegraphics[width=0.485\linewidth]{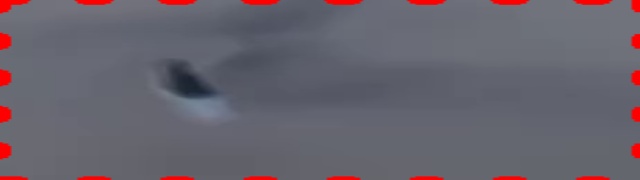} \\
  \vspace{-2mm}
  \caption{TK-Planes \cite{maxey2024tk}}
\end{subfigure}
\begin{subfigure}{0.192\linewidth} 
  \centering
\includegraphics[width=1\linewidth]{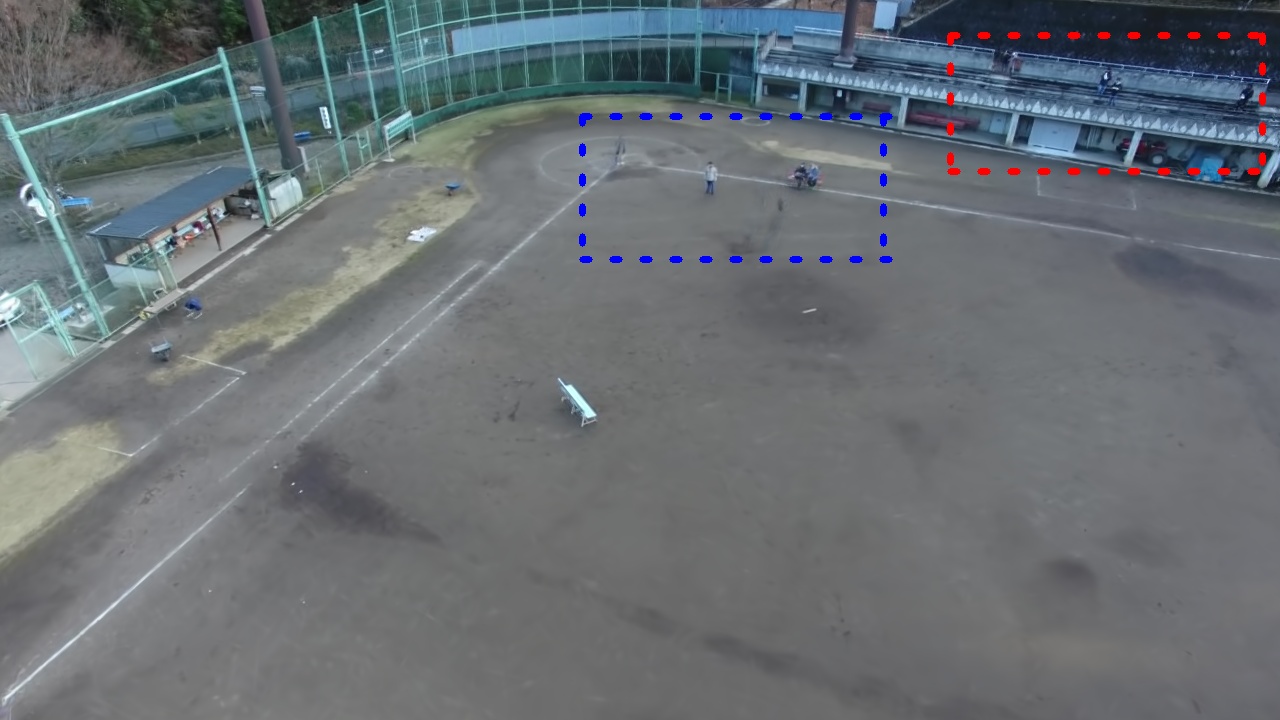} \\[0.5ex]
\includegraphics[width=0.485\linewidth]{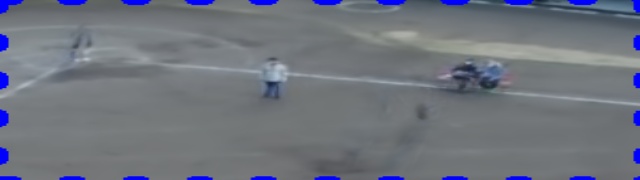} 
\hfill
\includegraphics[width=0.485\linewidth]{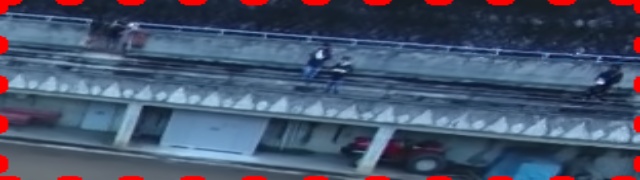} \\[0.5ex]
\includegraphics[width=1\linewidth]{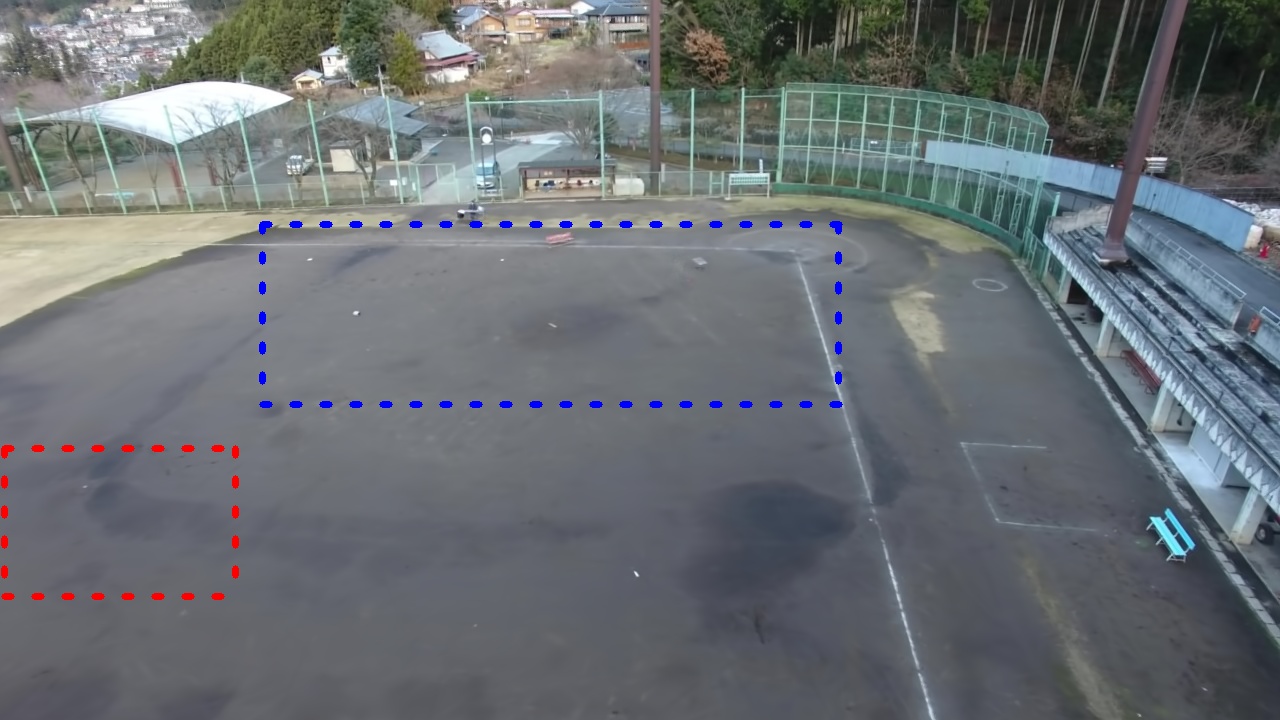} \\[0.5ex]
\includegraphics[width=0.485\linewidth]{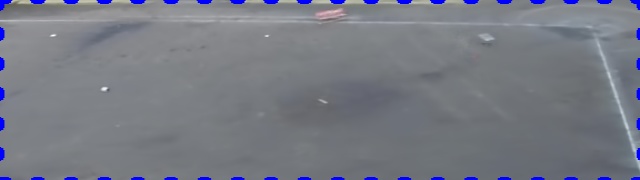} 
\hfill
\includegraphics[width=0.485\linewidth]{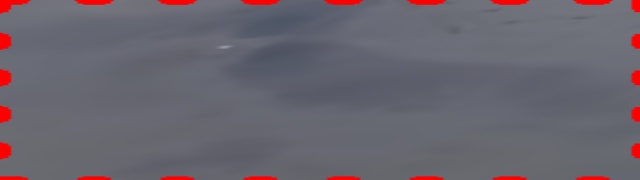} \\
  \vspace{-2mm}
  \caption{3DGS \cite{kerbl20233d}}
\end{subfigure}
\begin{subfigure}{0.192\linewidth} 
  \centering
\includegraphics[width=1\linewidth]{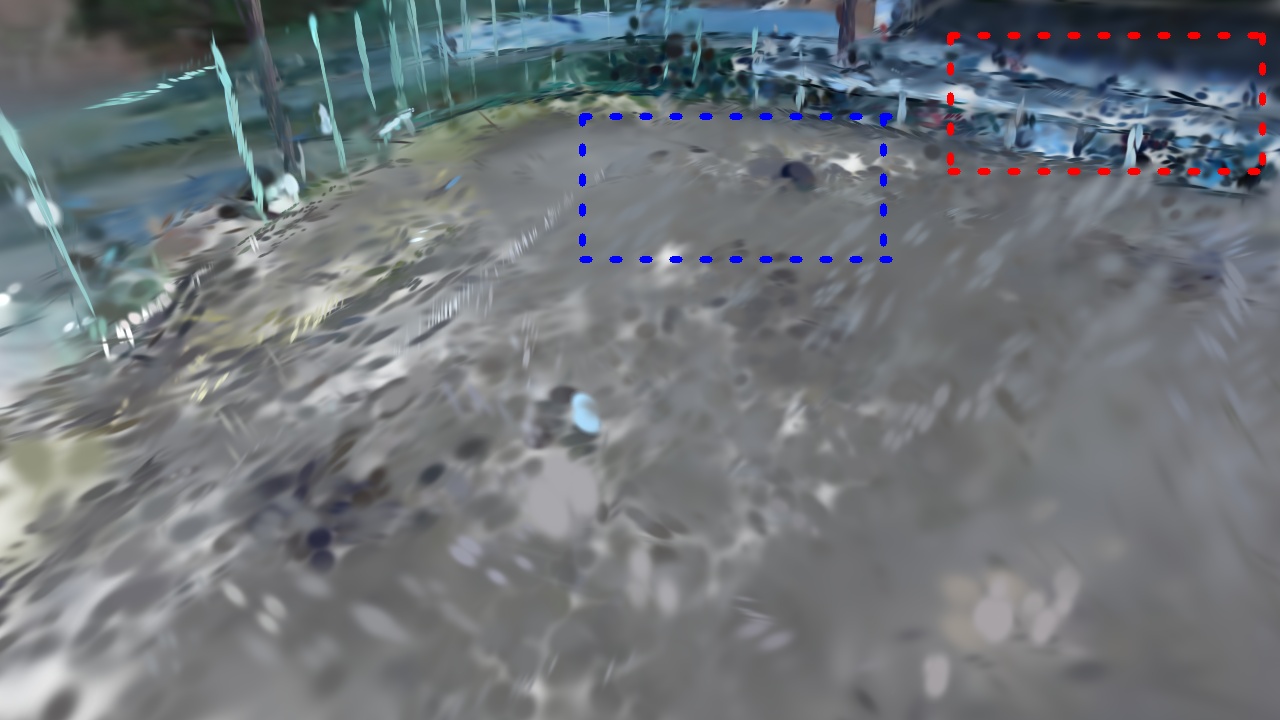} \\[0.5ex]
\includegraphics[width=0.485\linewidth]{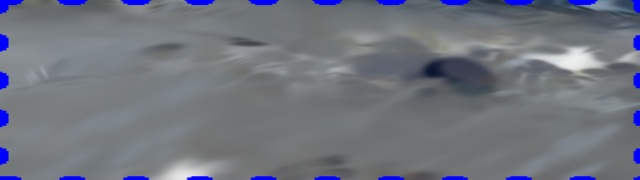} 
\hfill
\includegraphics[width=0.485\linewidth]{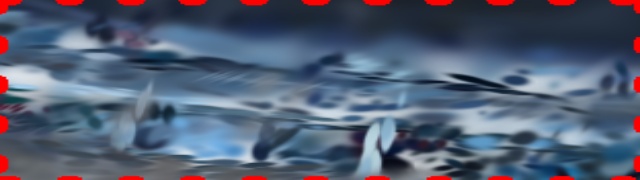} \\[0.5ex]
\includegraphics[width=1\linewidth]{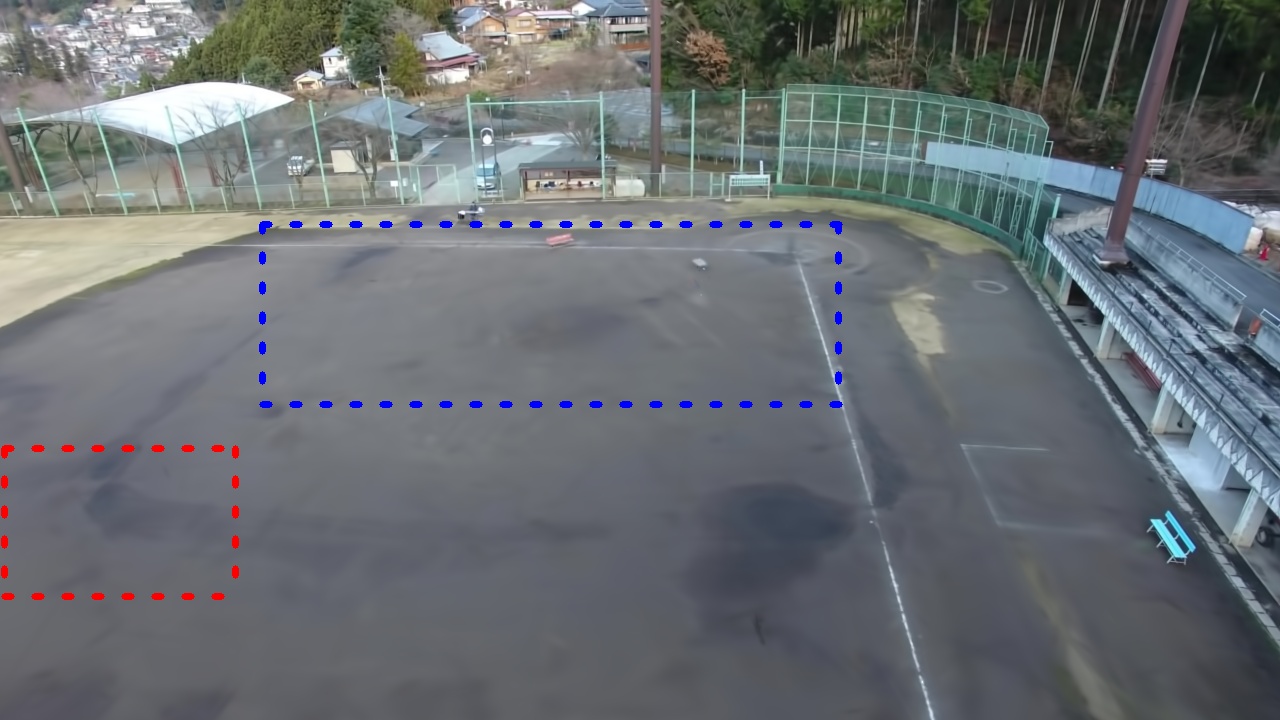} \\[0.5ex]
\includegraphics[width=0.485\linewidth]{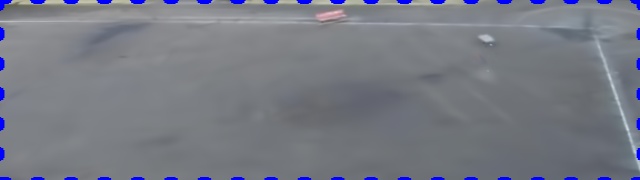} 
\hfill
\includegraphics[width=0.485\linewidth]{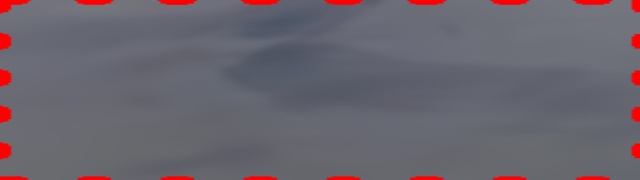} \\
  \vspace{-2mm}
  \caption{4DGS \cite{wu20244d}}
\end{subfigure}
\begin{subfigure}{0.192\linewidth} 
  \centering
\includegraphics[width=1\linewidth]{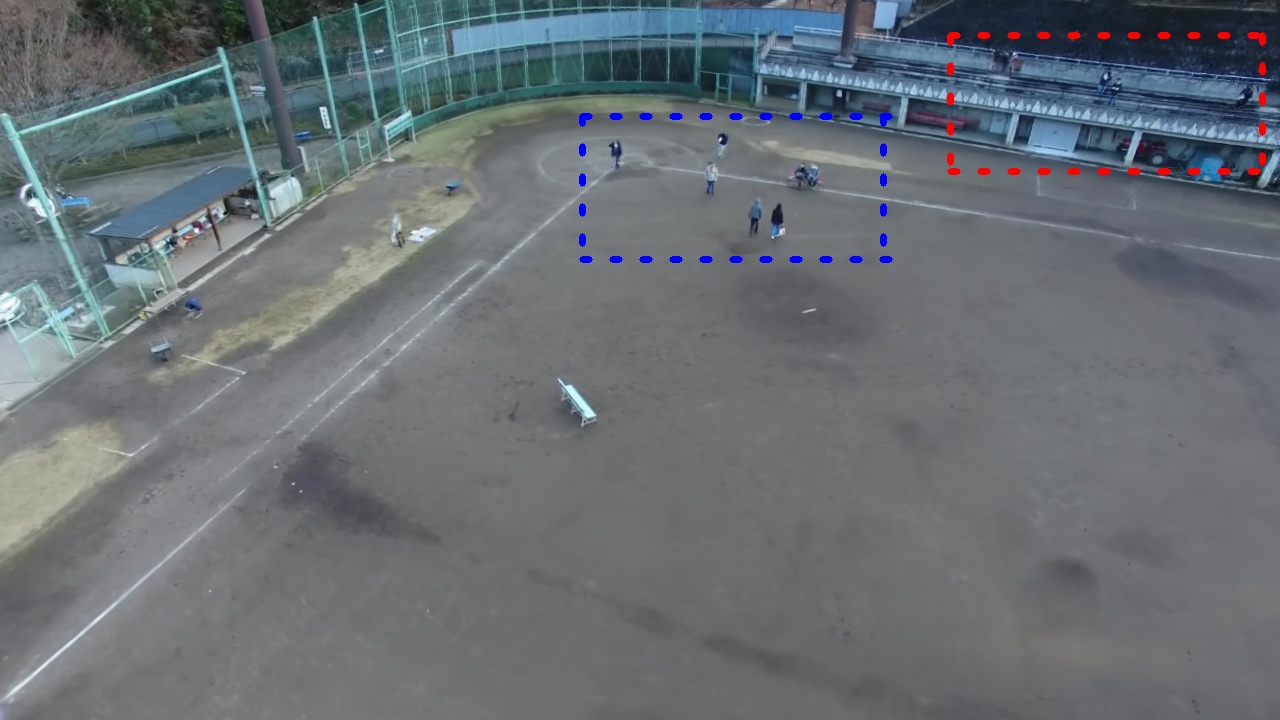} \\[0.5ex]
\includegraphics[width=0.485\linewidth]{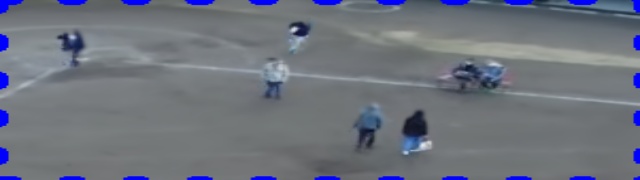} 
\hfill
\includegraphics[width=0.485\linewidth]{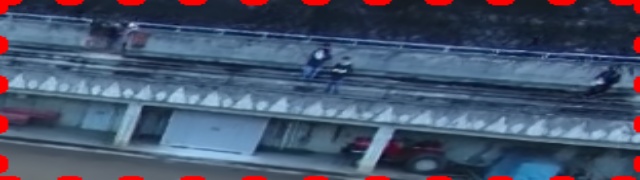} \\[0.5ex]
\includegraphics[width=1\linewidth]{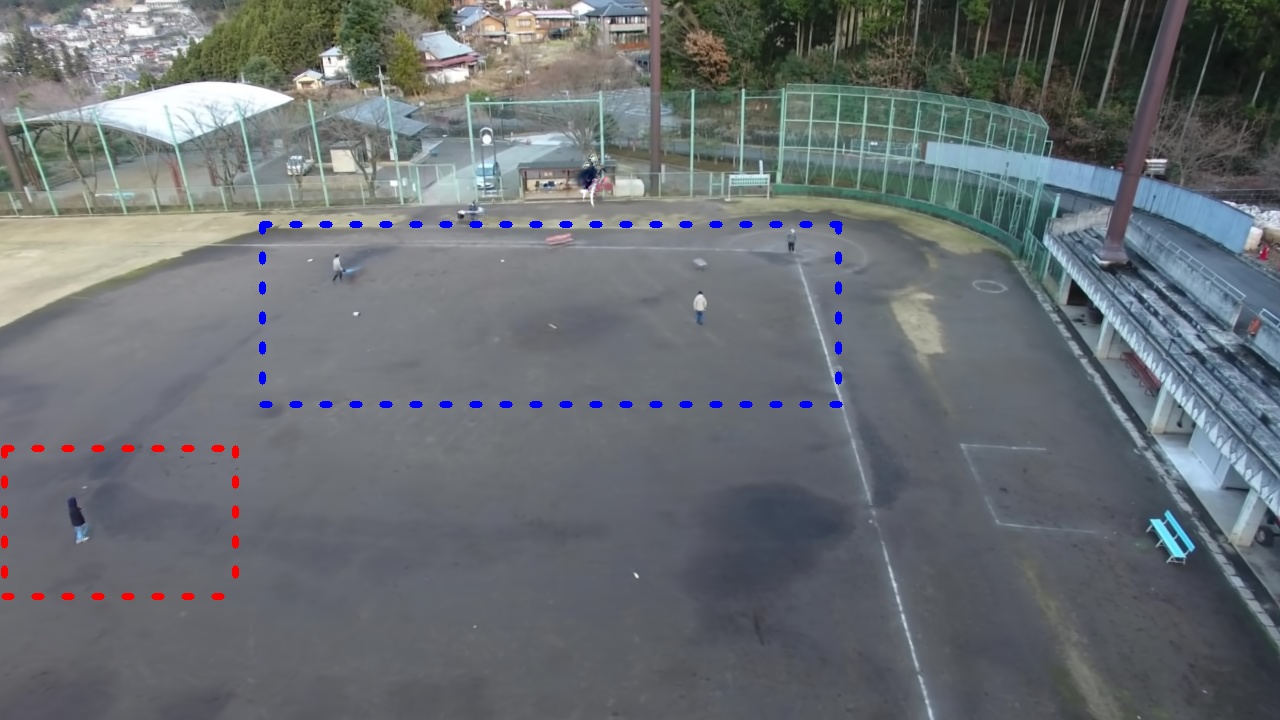} \\[0.5ex]
\includegraphics[width=0.485\linewidth]{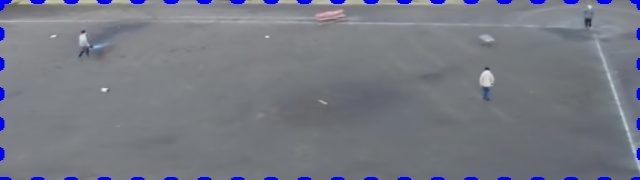} 
\hfill
\includegraphics[width=0.485\linewidth]{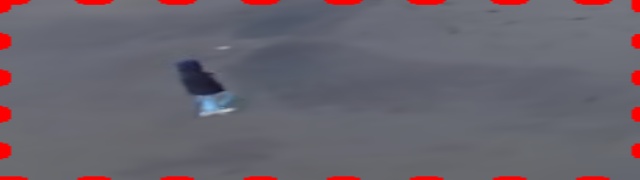} \\
  \vspace{-2mm}
  \caption{UAV4D (Ours)}
\end{subfigure}
\vspace{-6mm}
\caption{
\textbf{Quantitative Comparison of Okutama-Action Dataset \cite{barekatain2017okutama}.} 
We visualize the zoomed-in blue and red regions, which emphasize the dynamic humans. Our method demonstrates superior capability in reconstructing small, moving humans compared to other existing approaches.
} 
\vspace{-4mm}
\label{fig:okutama}
\end{figure}

\begin{figure}[t]
\captionsetup[subfigure]{labelformat=empty}
\centering
\makebox[3pt]{\raisebox{60pt}{\rotatebox[origin=c]{90}{\hspace{2.7em} \tiny{50\_RD\_P1} \hspace{4.5em} \tiny{40\_GND\_P1}}}}
\begin{subfigure}{0.192\linewidth} 
  \centering
\includegraphics[width=1\linewidth]{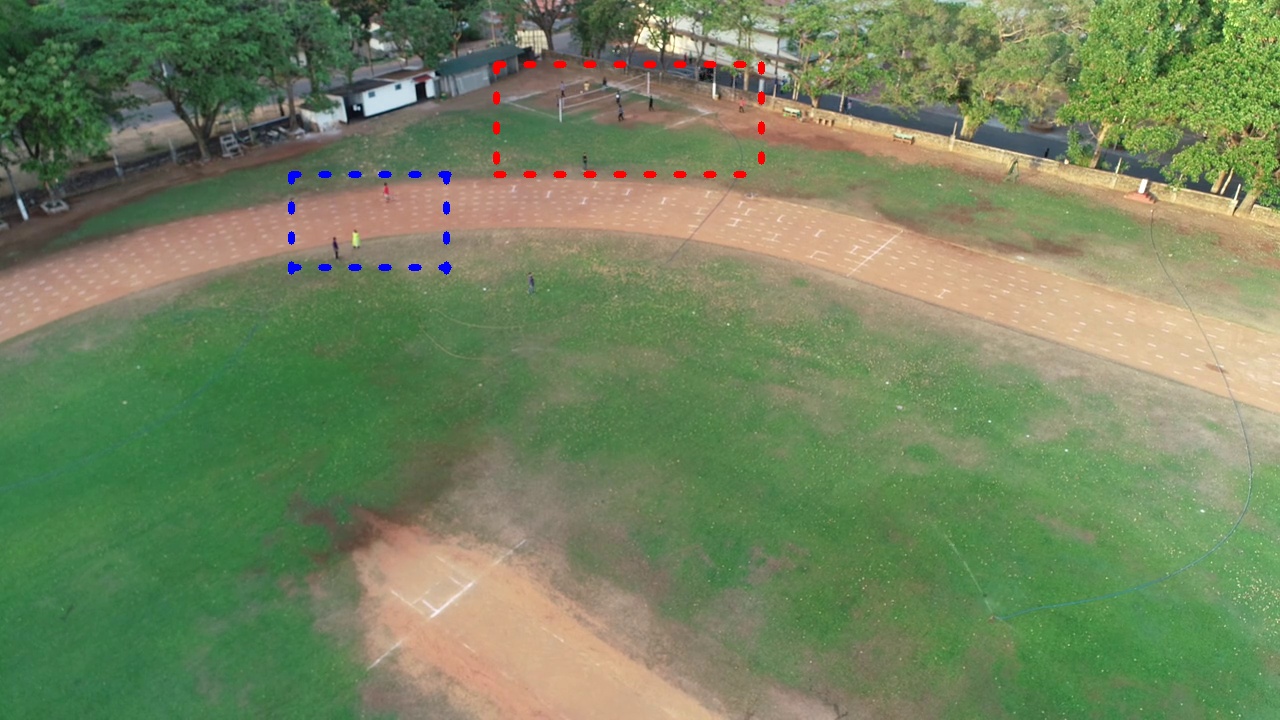} \\[0.5ex]
\includegraphics[width=0.485\linewidth]{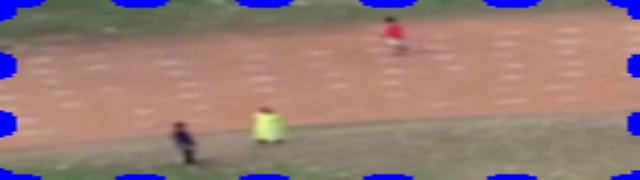} 
\hfill
\includegraphics[width=0.485\linewidth]{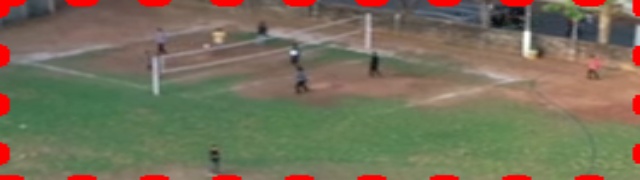} \\[0.5ex]
\includegraphics[width=1\linewidth]{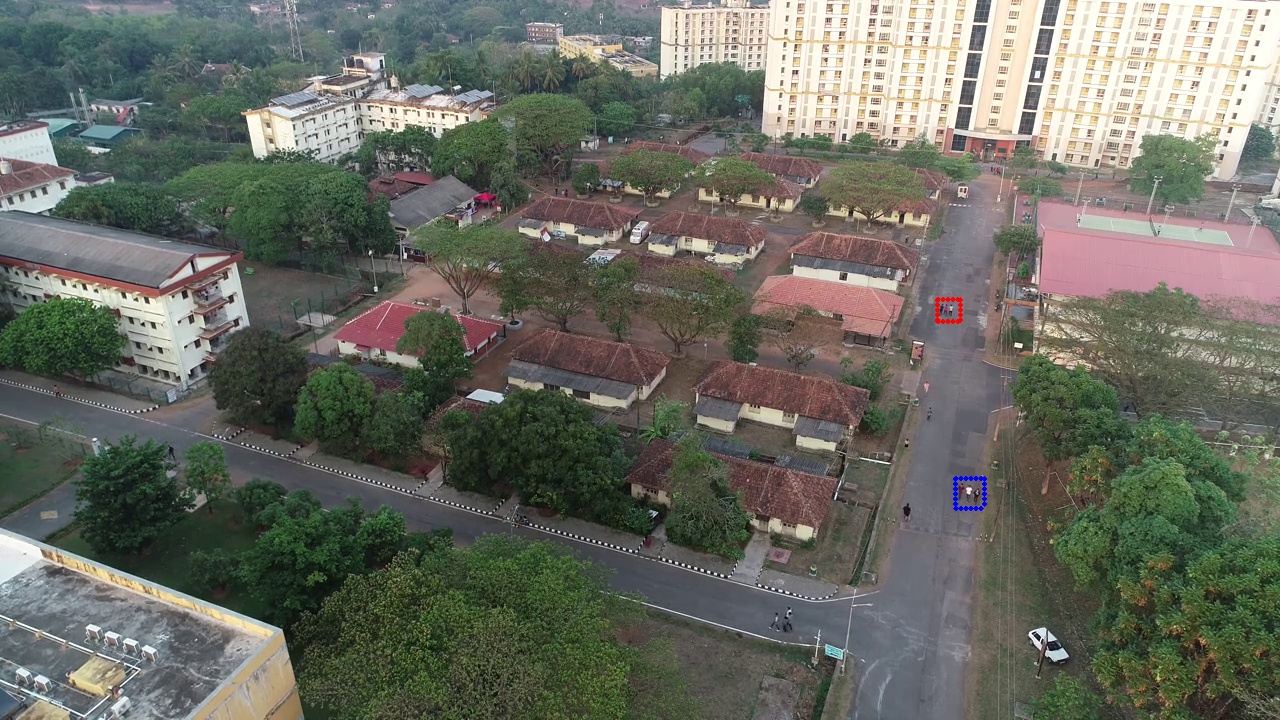} \\[0.5ex]
\includegraphics[width=0.485\linewidth]{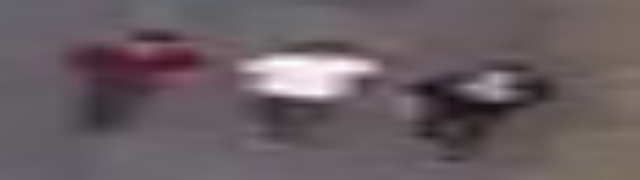} 
\hfill
\includegraphics[width=0.485\linewidth]{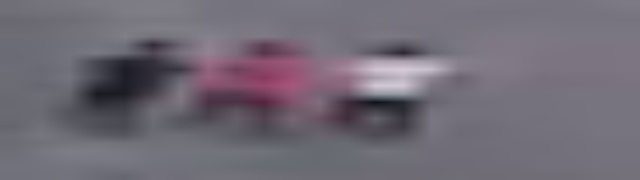} \\
  \vspace{-2mm}
  \caption{Scene}
\end{subfigure}
\begin{subfigure}{0.192\linewidth} 
  \centering
\includegraphics[width=1\linewidth]{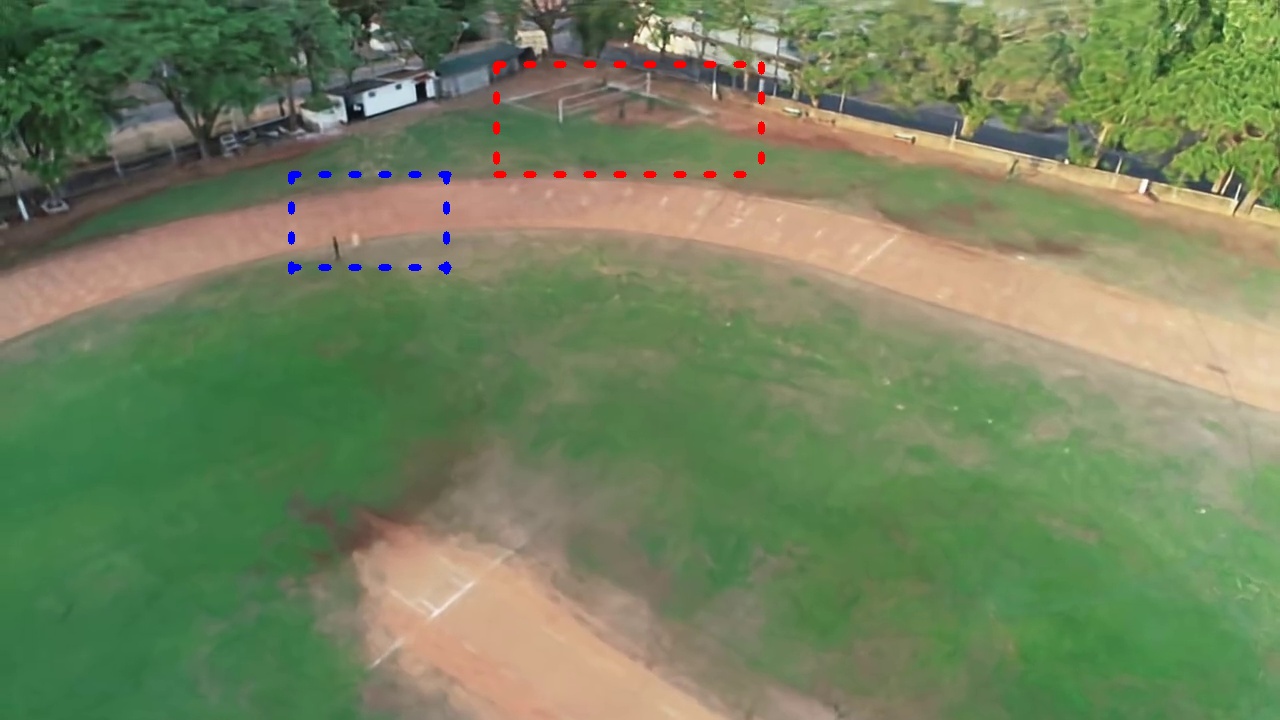} \\[0.5ex]
\includegraphics[width=0.485\linewidth]{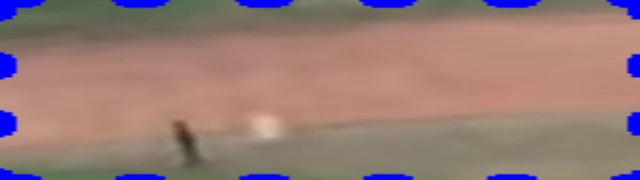} 
\hfill
\includegraphics[width=0.485\linewidth]{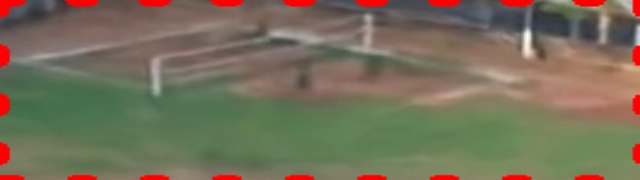} \\[0.5ex]
\includegraphics[width=1\linewidth]{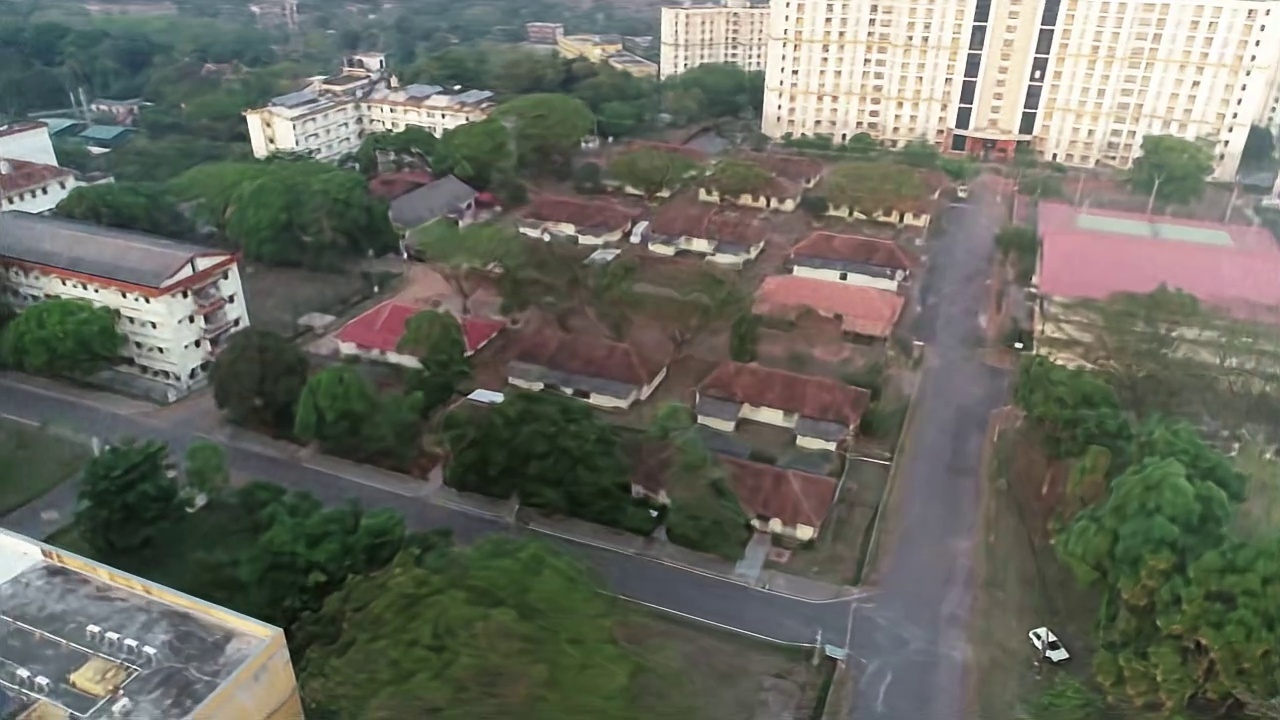} \\[0.5ex]
\includegraphics[width=0.485\linewidth]{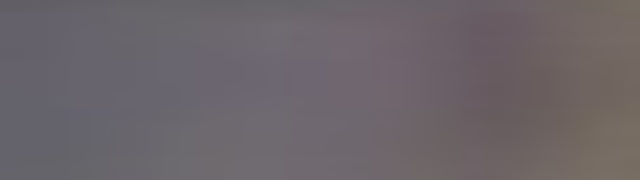} 
\hfill
\includegraphics[width=0.485\linewidth]{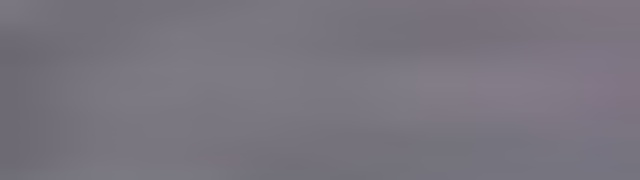} \\
  \vspace{-2mm}
  \caption{TK-Planes \cite{maxey2024tk}}
\end{subfigure}
\begin{subfigure}{0.192\linewidth} 
  \centering
\includegraphics[width=1\linewidth]{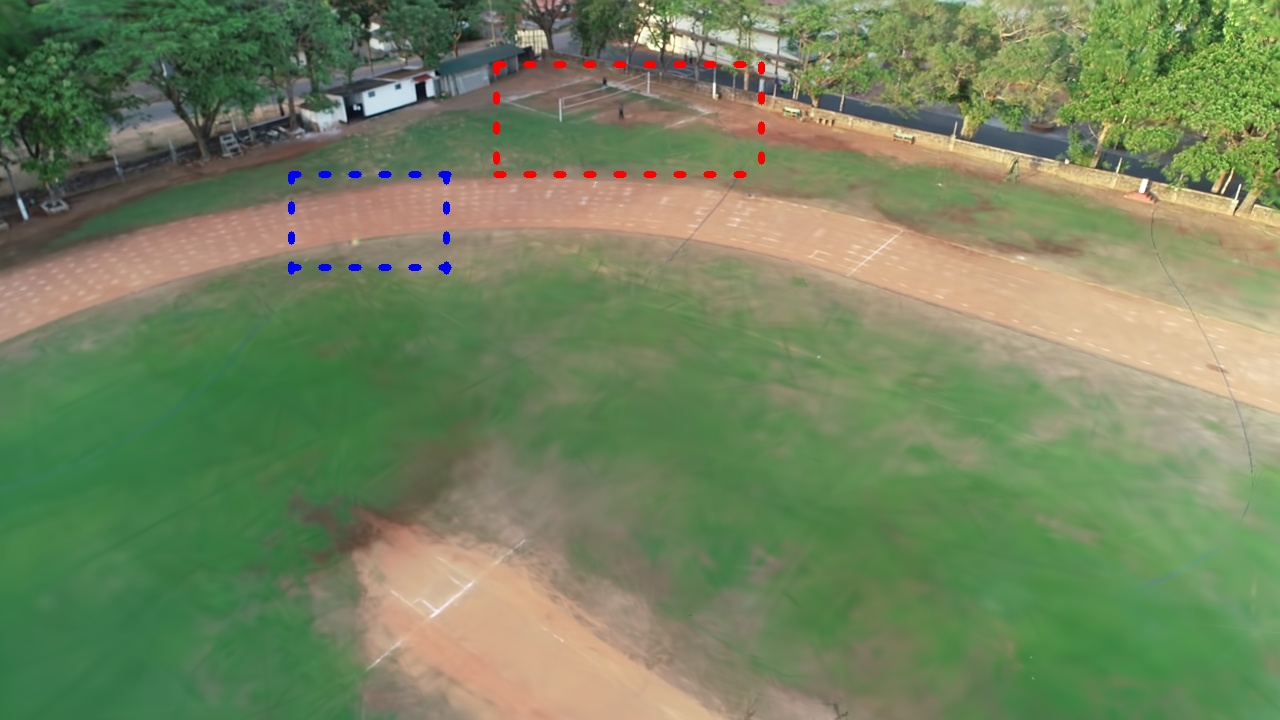} \\[0.5ex]
\includegraphics[width=0.485\linewidth]{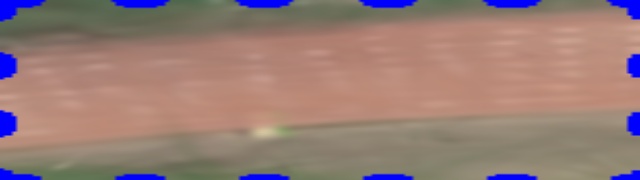} 
\hfill
\includegraphics[width=0.485\linewidth]{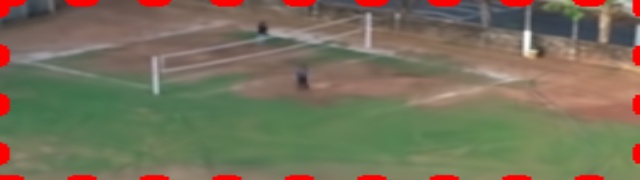} \\[0.5ex]
\includegraphics[width=1\linewidth]{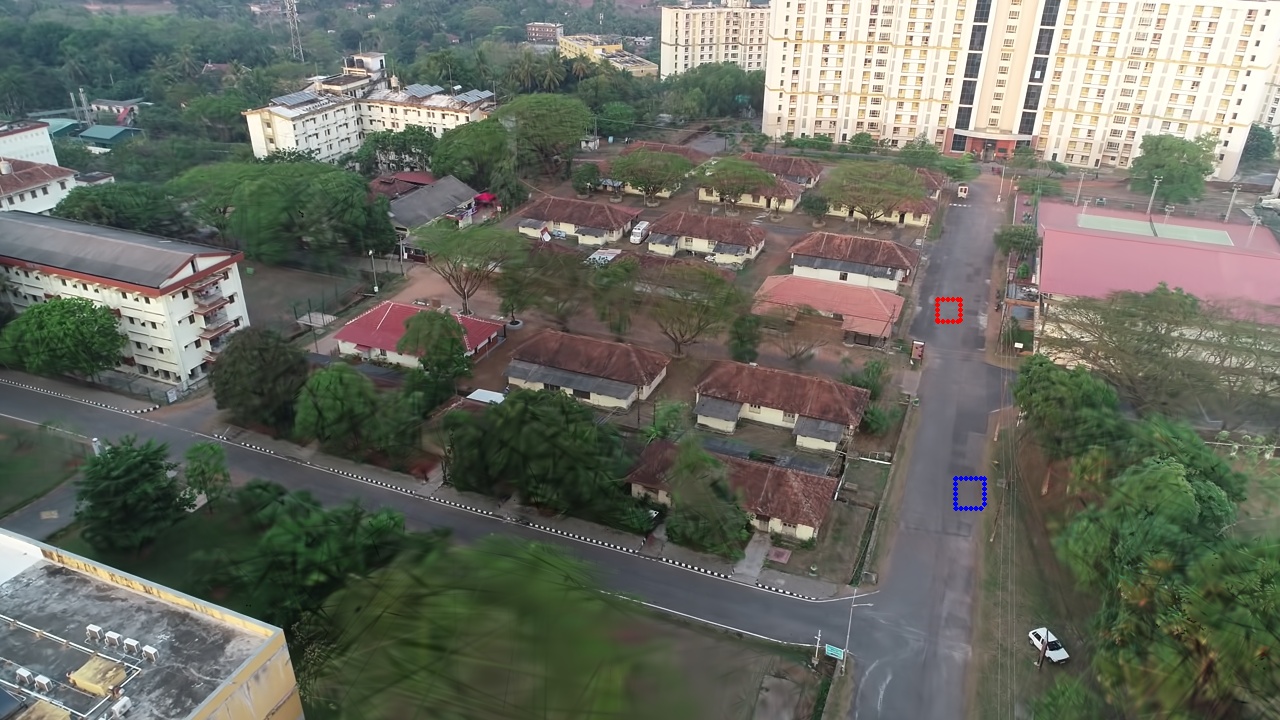} \\[0.5ex]
\includegraphics[width=0.485\linewidth]{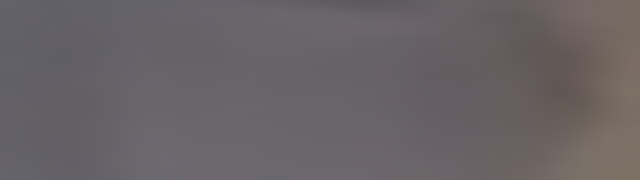} 
\hfill
\includegraphics[width=0.485\linewidth]{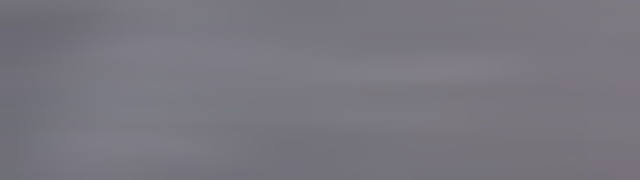} \\
  \vspace{-2mm}
  \caption{4DGS \cite{wu20244d}}
\end{subfigure}
\begin{subfigure}{0.192\linewidth} 
  \centering
\includegraphics[width=1\linewidth]{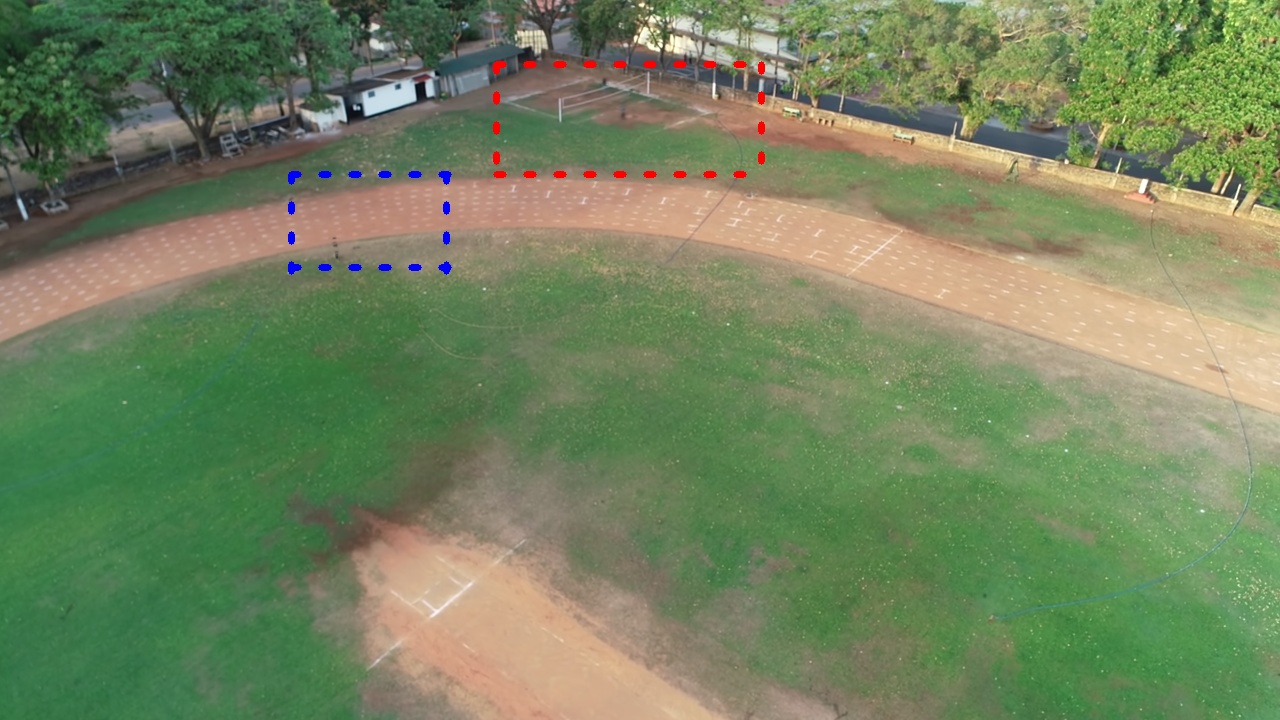} \\[0.5ex]
\includegraphics[width=0.485\linewidth]{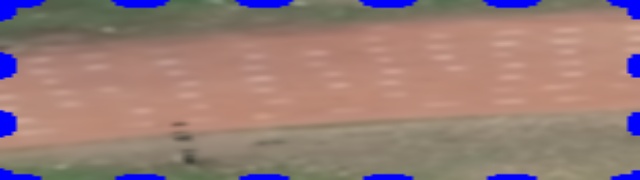} 
\hfill
\includegraphics[width=0.485\linewidth]{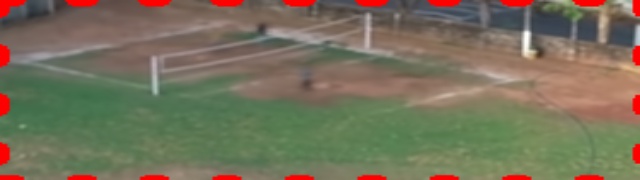} \\[0.5ex]
\includegraphics[width=1\linewidth]{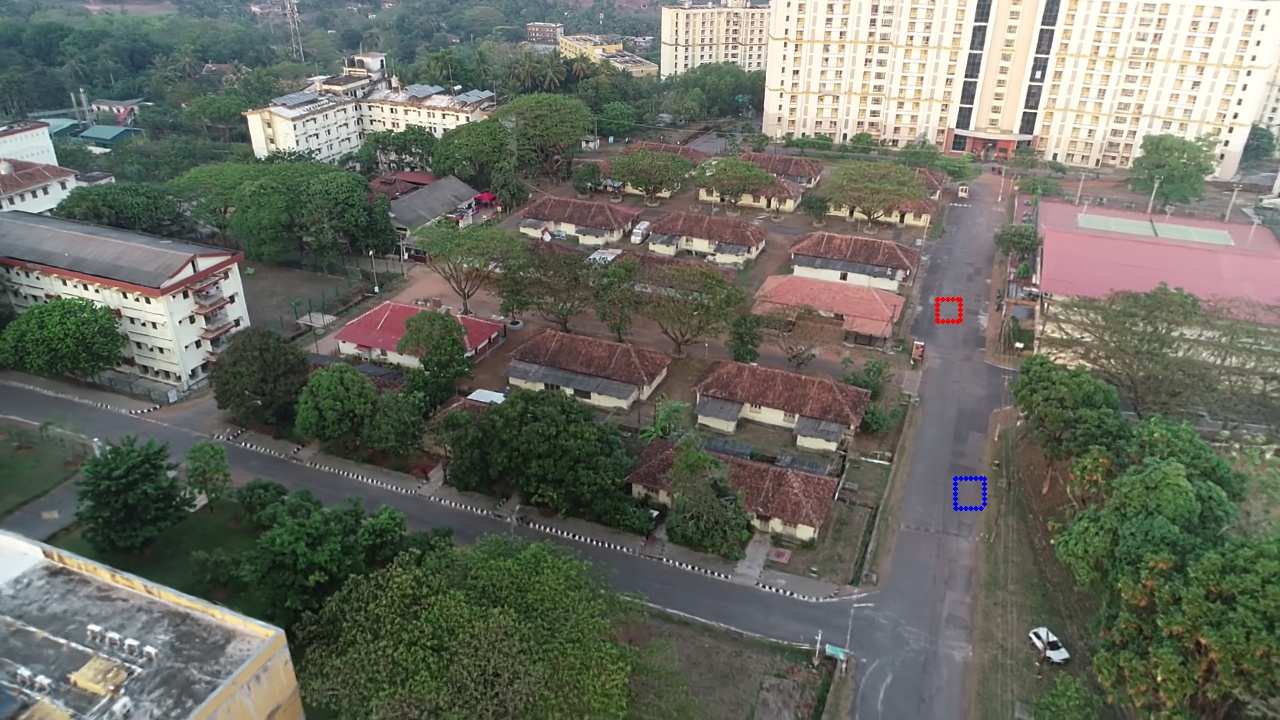} \\[0.5ex]
\includegraphics[width=0.485\linewidth]{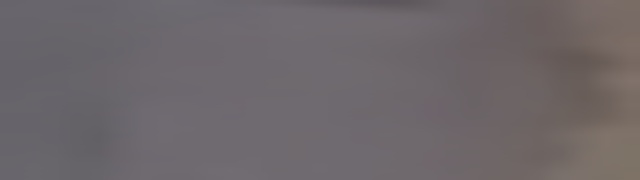} 
\hfill
\includegraphics[width=0.485\linewidth]{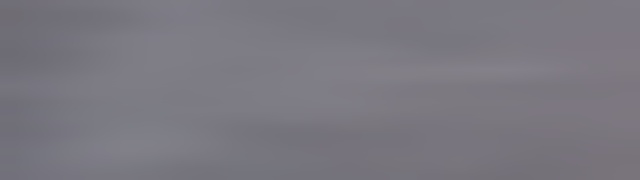} \\
  \vspace{-2mm}
  \caption{Deform-GS \cite{yang2024deformable}}
\end{subfigure}
\begin{subfigure}{0.192\linewidth} 
  \centering
\includegraphics[width=1\linewidth]{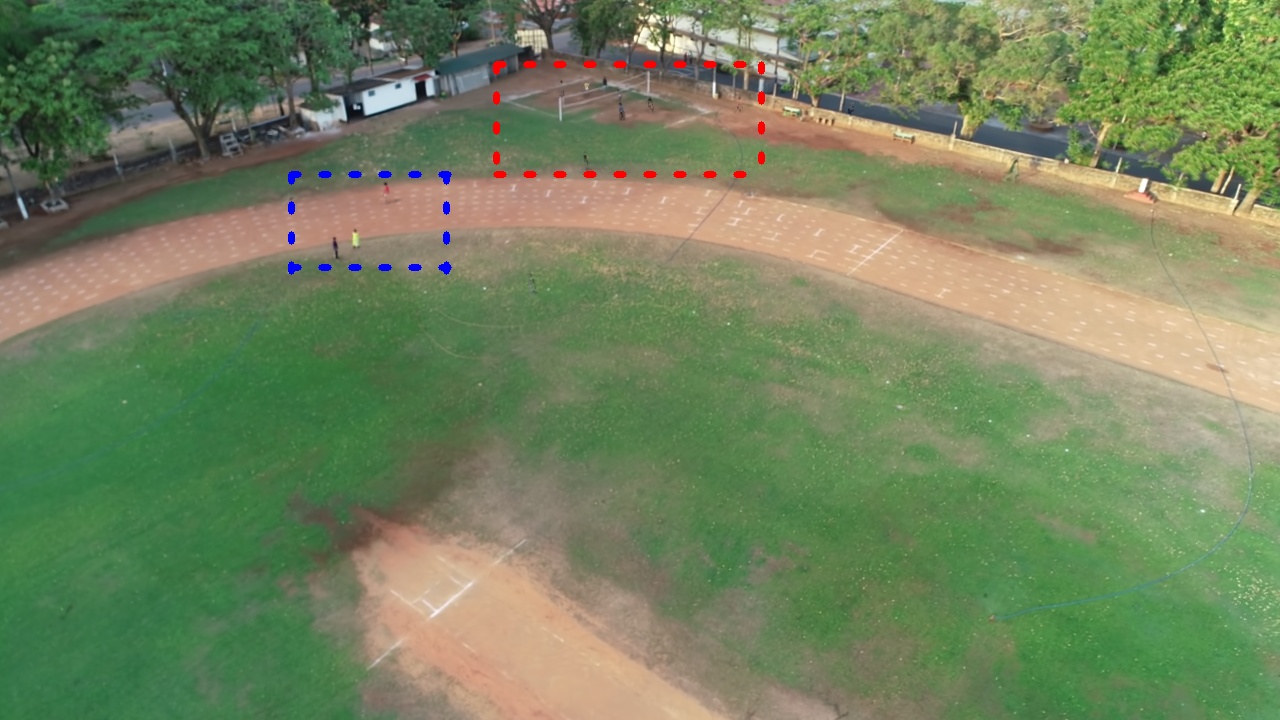} \\[0.5ex]
\includegraphics[width=0.485\linewidth]{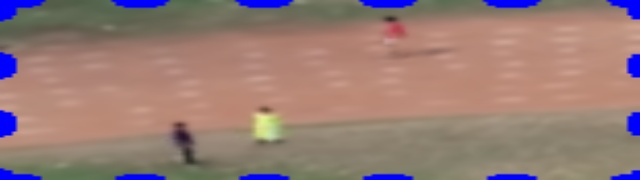} 
\hfill
\includegraphics[width=0.485\linewidth]{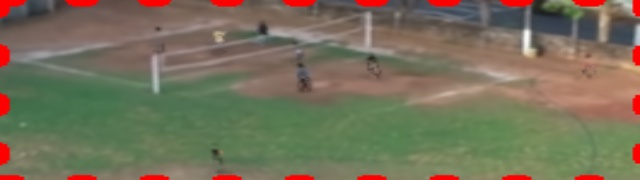} \\[0.5ex]
\includegraphics[width=1\linewidth]{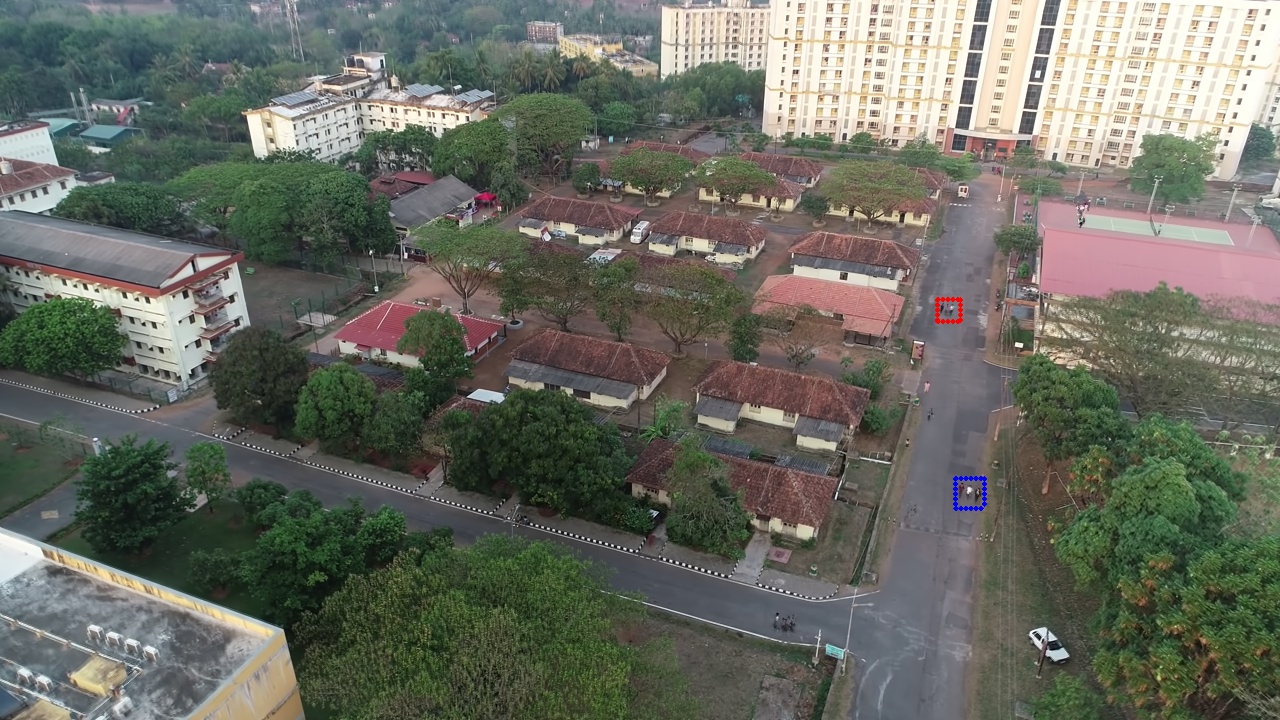} \\[0.5ex]
\includegraphics[width=0.485\linewidth]{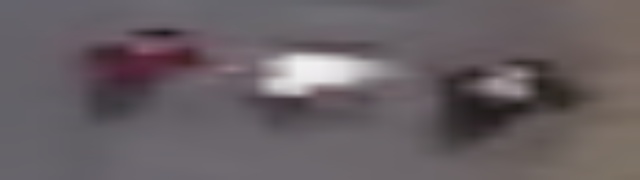} 
\hfill
\includegraphics[width=0.485\linewidth]{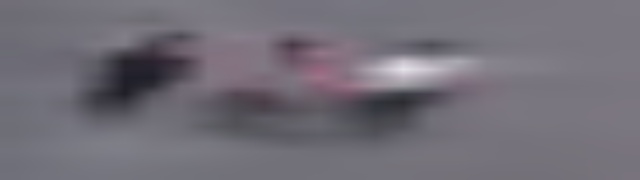} \\
  \vspace{-2mm}
  \caption{UAV4D (Ours)}
\end{subfigure}
\vspace{-6mm}
\caption{
\textbf{Quantitative Comparison of Manipal-UAV Dataset \cite{akshatha2023manipal}.} 
We visualize the zoomed-in blue and red regions, which emphasize the dynamic humans. Our method demonstrates superior capability in reconstructing small, moving humans compared to other existing approaches. The rendering results of 3DGS \cite{kerbl20233d} are not included, as thie method ignores moving humans, similar to other methods.
} 
\vspace{-4mm}
\label{fig:manipal}
\end{figure}

\begin{figure}[h]
    \centering
    \includegraphics[width=0.99\linewidth]{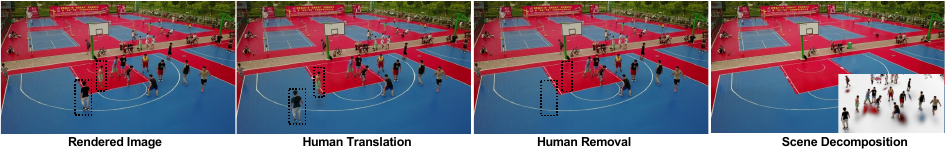}
    \vspace{-2mm}
    \caption{
    \textbf{Scene Editing Samples}: Due to the nature of scene decomposition, we can move or remove the humans within the black-dotted boxes. The fourth image illustrates the decomposition of the background and humans. For additional visual results, please refer to the supplementary materials.}
    \label{fig:editing}
    \vspace{-2mm}
\end{figure}

\subsection{Qualitative Comparison}

Figure \ref{fig:visdrone} presents visual examples from the VisDrone dataset \cite{cao2021visdrone}. Overall, our method achieves the highest rendering quality, producing a photorealistic background and sharp human regions, where the detailed shapes of the humans are more accurately recovered compared to other methods. TK-planes \cite{maxey2024tk} generates blurred renderings, while 3DGS \cite{kerbl20233d} is unable to represent moving individuals, although it successfully reconstructs clear background regions. The training of 4DGS \cite{wu20244d} occasionally fails. A similar trend is observed in the Okutama-Action dataset \cite{barekatain2017okutama}, as shown in Figure \ref{fig:okutama}.
Figure \ref{fig:manipal} displays examples from the Manipal-UAV dataset \cite{akshatha2023manipal}.
Our method can reconstructs extremely small humans, occupying only around 10x10 pixels, outperforming all other methods.
However, Deformable-GS \cite{yang2024deformable} shows better performance in rendering swaying leaves and trees, which are frequently observed in this dataset.

\noindent\textbf{Scene Editing} Since our Gaussian scene is composed of separate human and background Gaussians, our system enables scene editing. As demonstrated in Fig. \ref{fig:editing}, we can easily remove, and translate humans, thanks to the decomposition of the foreground and background. This capability can be extended to downstream applications, such as simulating dynamic scenes captured by UAV that are expensive to capture in the real world.


\section{Conclusion, Limitations, and Future Work}
\label{sec:conclusion}
We have introduced UAV4D, a method for dynamic Gaussian splatting in UAV-captured environments. 
We propose a scale optimization and human placement technique to reconstruct human-scene geometry in the world coordinate system. 
This human-scene geometry is then used to initialize 3D Gaussian splats for dynamic scenes.
It is important to note that the explicit representation of human geometry in our method is crucial for accurately modeling small, moving humans, ensuring they are not neglected during the optimization of Gaussian splats. 
Our joint optimization approach demonstrates improved performance in novel view synthesis, both for background and human regions.
However, our method has several limitations. 
First, it relies on pretrained models (e.g. HMR2.0 \cite{goel2023humans} or SAM2 \cite{ravi2024sam}). 
In particular, HMR2.0 \cite{goel2023humans} still exhibits inaccuracies in human mesh parameters, which leads to rendering artifacts in the scene. 
Additionally, other dynamic components, such as bicycles or swaying trees, are not effectively rendered using static background Gaussian splats, except for the human regions.
In future work, we plan to modify our human mesh refinement algorithm to better align with 2D images and incorporate segmentation information to represent other dynamic components using dynamic Gaussian splats. Our approach can be combined with downstream UAV applications such as person detection \cite{maxey2024uav} and human action recognition \cite{barekatain2017okutama}.

{\small
\bibliographystyle{plainnat}
\bibliography{main}
}

\appendix

\section{Dataset \& Training Details}

\begin{figure}[h]
\captionsetup[subfigure]{labelformat=empty}
\centering
\begin{subfigure}{0.24\linewidth} 
  \centering
\includegraphics[width=1\linewidth]{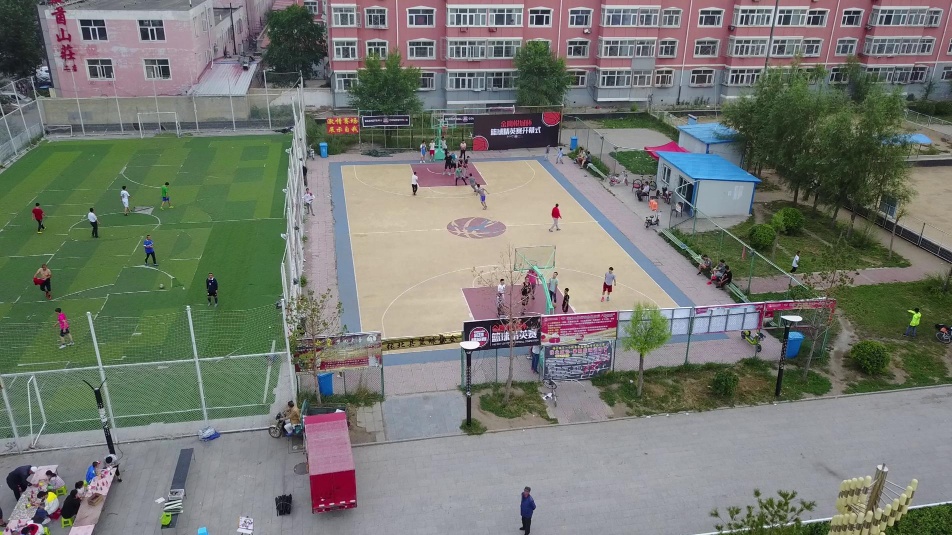} \\[0.5ex]
  \vspace{-2mm}
  \caption{Scene}
\end{subfigure}
\begin{subfigure}{0.24\linewidth} 
  \centering
\includegraphics[width=1\linewidth]{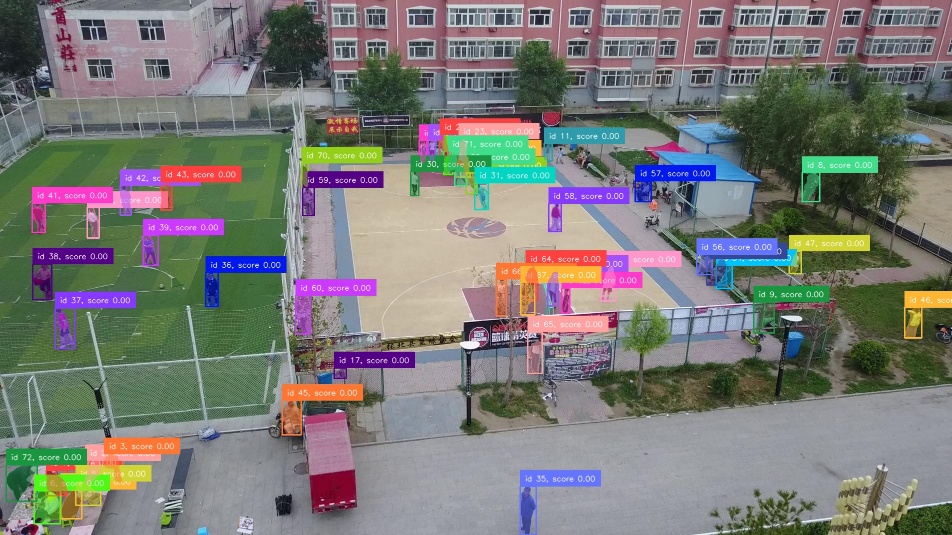} \\[0.5ex]
  \vspace{-2mm}
  \caption{SAM2 \cite{ravi2024sam}}
\end{subfigure}
\begin{subfigure}{0.24\linewidth} 
  \centering
\includegraphics[width=1\linewidth]{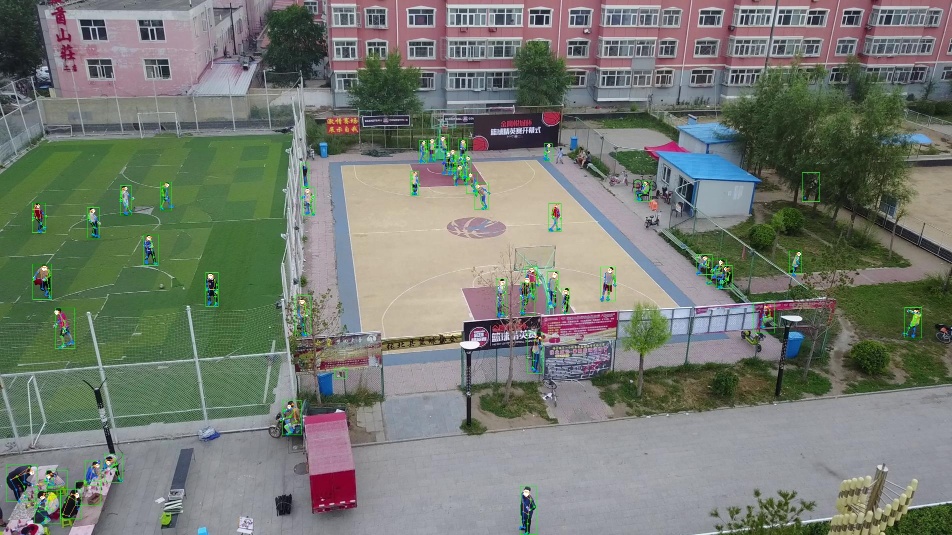} \\[0.5ex]
  \vspace{-2mm}
  \caption{ViTPose \cite{xu2022vitpose}}
\end{subfigure}
\begin{subfigure}{0.24\linewidth} 
  \centering
\includegraphics[width=1\linewidth]{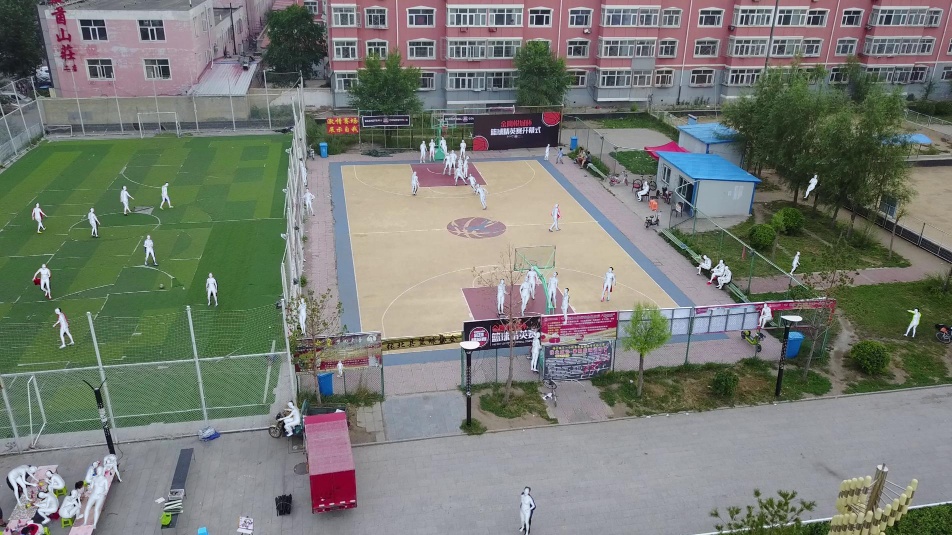} \\[0.5ex]
  \vspace{-2mm}
  \caption{HMR2.0 \cite{goel2023humans}}
\end{subfigure}
\vspace{-3mm}
\caption{
\textbf{Visual Examples of Initial Estimates.} Each initial estimate is used as input for the subsequent human-scene reconstruction process.
} 
\vspace{-2mm}
\label{fig:preprocess_data}
\end{figure}

\subsection{Dataset Details}
We select four scenes from each dataset, with most scenes comprising approximately 200 images in total, including both training and test sets. 
\begin{enumerate}[left=0.3cm, itemsep=-0.2ex] 
\item VisDrone: uav0013\_00000, uav0079\_00480, uav0084\_00000, uav0099\_02109
\item Manipal-UAV: 40\_GND\_P1, 40\_VGT\_P1, 50\_GND\_P1, 50\_RD\_P1
\item Okutama-Action: 1\_2\_2, 1\_2\_6, 1\_2\_8, 1\_2\_9
\end{enumerate}

In the VisDrone dataset, humans have an average size of approximately 29×78 pixels, while in the Manipal-UAV dataset, the average size is around 7×17 pixels. In the Okutama dataset, humans typically appear at an average size of 17×44 pixels.

\subsection{Training Details}

\noindent\textbf{Initialization} 
Figure \ref{fig:preprocess_data} presents visual examples of human masks generated by SAM2 \cite{ravi2024sam}, 2D keypoint poses estimated by VitPose \cite{xu2022vitpose}, and human meshes reconstructed by HMR2.0 \cite{goel2023humans}. 
We observe that human bounding boxes are generally more reliable than the reconstructed human mesh parameters. 
Therefore, we set $\eta_{box}$ to 1.2 to filter out erroneous projected human meshes in image space that appear excessively large compared to their corresponding bounding boxes.

\noindent\textbf{Joint Human-Scene Reconstruction}
To reconstruct the background mesh, we first filter point maps based on confidence scores. 
The confidence threshold $\eta_{conf}$ is determined according to the UAV's altitude. Specifically, we set $\eta_{conf}$ to 5 for the Manipal-UAV dataset, 20 for Okutama-Action, and 40 for the VisDrone dataset. 
For high-altitude captures (e.g., 40–50 meters), a lower confidence threshold is used since most point maps correspond to static background. 
In contrast, for low-altitude scenarios, a higher confidence threshold is applied to exclude point maps from dynamic regions.
For scale optimization, we first initialize the scale $\sigma$ to 40. 
We then select approximately 5–10 human instances and sample 5 frames at intervals of 5 frames. Keypoints corresponding to the main body are used for optimization
We set the learning rate $\alpha$ to 0.1 and perform $T_{opt}$ to 30 optimization steps. 
We note that the root translation is typically associated with the pelvis joint. 
To convert the ground contact point to the root translation, we apply a vertical offset of 1 unit.

\noindent\textbf{3D Gaussian Sceen Modeling} 
To optimize all Gaussians, we use various loss functions and regularizations in Eq. (5) in the main paper. 
The opacity regularization $L_{o}$ is defined as
\begin{equation} \label{eq:opacity}
L_{o} = \sum -o_{all}log(o_{all}) - \sum m_{sky} log(1-o_{all})
\end{equation}

where $o_{all}$ is the rendered opacity map from all Gaussian attributes and $m_{sky}$ is the sky mask. 
Most of the UAV scenes selected do not include sky regions. Only two scenes—uav0013\_00000 and uav0099\_02109 from the VisDrone dataset—contain visible sky areas. The remaining ten scenes exclude sky regions from consideration.

Following GART \cite{lei2024gart} and OmniRe \cite{chen2025omnire}, $L_{smpl}$ includes KNN regularization $L_{knn}$ and temporal smoothness regularization $L_{smooth}$. We apply KNN regularization for all Gaussian attributes to constrain the transformation and appearance of neighbors in order to reduce the artifacts. 
\begin{equation} \label{eq:opacity}
L_{knn} = \sum_{(\mu, r, s, c, o) \in g_{h}} \lambda_{knn} STD_{i \in KNN(\mu_{i})}(\mu_{i}, r_{i}, s_{i}, c_{i}, o_{i}), 
\end{equation}
where $STD$ denotes the standard deviation. 
Furthermore, we apply temporal smoothness regularization to ensure that human body poses $\theta_t$ vary smoothly over time by encouraging consistency between the current pose and its adjacent frames,
\begin{equation} \label{eq:opacity}
L_{smooth} = 0.5 \lambda_{smooth} ||(\theta_{t} - \theta_{t-1}) + (\theta_{t} - \theta_{t+1} )||.
\end{equation}

We set $\lambda_{pho}$ to 0.2, $\lambda_{o}$ to 0.05, $\lambda_{knn}$ to 0.01, $\lambda_{smooth}$ to 0.001. 
We optimize all Gaussian attributes and all SMPL parameters using Adam optimizer for 30,000 iterations. 
For optimizing background Gaussian attributes, we set the learning rates for rotations, scales, zeroth-order spherical harmonics, the remaining spherical harmonics to 0.001, 0.005, 0.0025, and 0.000125 respectively. 
To optimize the human Gaussian attributes, we set the learning rates for rotations, scales, zeroth-order spherical harmonics, and higher-order spherical harmonics to 0.0016, 0.005, 0.0025, and 0.000125, respectively. For human pose parameters, the learning rates for rotation and translation are set to 0.00001 and 0.0005, respectively.
For densification for Gaussian attributes, we utilize the strategy of AbsGS \cite{ye2024absgs}. The densification threshold for the position gradient is set to 0.0003.

\section{Experimental Results}
\subsection{Zoom-in Human Regions}
\begin{figure}[h]
    \centering
    \includegraphics[width=1\linewidth]{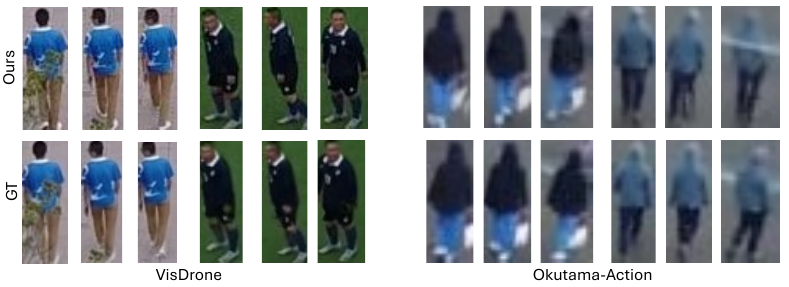}
     \caption{\textbf{Visual Examples of Zoom-in Human Regions} on VisDrone and Okutama-Action Datasets}
     \label{fig:human_only_regions} 
\end{figure}

Fig. \ref{fig:human_only_regions} shows example visualizations of human regions, comparing our method with the ground truth images. In the VisDrone dataset, human figures are approximately 30×79 pixels in size, while in the Okutama-Action dataset, they are around 17×44 pixels.

\subsection{Per-scene Breakdown}
To supplement Table 1 of the main paper, we present per-scene experimental results in terms of PSNR, SSIM, and LPIPS, as shown in Table \ref{tab:breakdown_visdrone}, \ref{tab:breakdown_okutama}, and \ref{tab:breakdown_manipal}. 
Table~\ref{tab:breakdown_visdrone_human}, \ref{tab:breakdown_okutama_human} and \ref{tab:breakdown_manipal_human} provide a breakdown of image quality metrics focused on human-only regions, complementing the results in Table 2 of the main paper.

\noindent\textbf{VisDrone Dataset} 
Additionally, we provide a detailed breakdown of the ablation study results to complement Table 3.

\begin{table*}[h]
    \centering
    \caption{\textbf{Breakdown results on the VisDrone dataset} \cite{cao2021visdrone}. Red, orange, and yellow indicate the first, second, and third best performing algorithms for each metric.}
    \begin{adjustbox}{width=0.99\linewidth,center}
    \begin{tabular}{l|ccc|ccc|ccc|ccc|ccc}
    \toprule
    \multirow{2}{*}{Method }&\multicolumn{3}{c|}{uav0013\_00000}&\multicolumn{3}{c|}{uav0079\_00480}&\multicolumn{3}{c|}{uav0084\_00000}&\multicolumn{3}{c|}{uav0099\_02109}&\multicolumn{3}{c}{Mean}\\
    & PSNR \(\uparrow\)  & SSIM \(\uparrow\)  & LPIPS \(\downarrow\) & PSNR \(\uparrow\)  & SSIM \(\uparrow\)  & LPIPS \(\downarrow\) & PSNR \(\uparrow\)  & SSIM \(\uparrow\)  & LPIPS \(\downarrow\) & PSNR \(\uparrow\)  & SSIM \(\uparrow\)  & LPIPS \(\downarrow\) & PSNR \(\uparrow\)  & SSIM \(\uparrow\)  & LPIPS \(\downarrow\) \\
    \midrule
TK-Planes \cite{maxey2024tk}   &  \cellcolor{tabthird}27.10 & 0.747 &  \cellcolor{tabthird}0.398 &  \cellcolor{tabthird}22.36 &  \cellcolor{tabthird}0.614 &  \cellcolor{tabthird}0.493 &  \cellcolor{tabthird}25.30 & 0.788 & 0.338 &  \cellcolor{tabthird}21.17 &  \cellcolor{tabthird}0.427 &  \cellcolor{tabthird}0.569 &  \cellcolor{tabthird}23.98 &  \cellcolor{tabthird}0.644 & 0.450 \\
3DGS \cite{kerbl20233d}        & \cellcolor{tabsecond}27.21 & \cellcolor{tabsecond}0.833 & \cellcolor{tabsecond}0.211 & \cellcolor{tabsecond}24.57 & \cellcolor{tabsecond}0.779 & \cellcolor{tabsecond}0.214 & 23.21 &  \cellcolor{tabthird}0.861 & \cellcolor{tabsecond}0.180 & \cellcolor{tabsecond}23.38 & \cellcolor{tabsecond}0.655 & \cellcolor{tabsecond}0.274 & \cellcolor{tabsecond}24.59 & \cellcolor{tabsecond}0.782 & \cellcolor{tabsecond}0.220 \\
4DGS \cite{yang2024deformable} & 26.74 &  \cellcolor{tabthird}0.761 & 0.401 & 18.27 & 0.507 & 0.546 & \cellcolor{tabsecond}26.40 & \cellcolor{tabsecond}0.885 &  \cellcolor{tabthird}0.208 & 18.90 & 0.358 & 0.603 & 22.58 & 0.628 &  \cellcolor{tabthird}0.440 \\
DeformableGS \cite{wu20244d}   & 12.66 & 0.522 & 0.667 & 13.72 & 0.477 & 0.691 & 21.22 & 0.757 & 0.376 & 14.22 & 0.330 & 0.724 & 15.46 & 0.522 & 0.614 \\
UAV4D (Ours)                    &  \cellcolor{tabfirst}28.13 &  \cellcolor{tabfirst}0.847 &  \cellcolor{tabfirst}0.149 &  \cellcolor{tabfirst}25.52 &  \cellcolor{tabfirst}0.801 &  \cellcolor{tabfirst}0.137 &  \cellcolor{tabfirst}26.82 &  \cellcolor{tabfirst}0.886 &  \cellcolor{tabfirst}0.074 &  \cellcolor{tabfirst}23.66 &  \cellcolor{tabfirst}0.665 &  \cellcolor{tabfirst}0.227 &  \cellcolor{tabfirst}26.03 &  \cellcolor{tabfirst}0.800 &  \cellcolor{tabfirst}0.147 \\
\hline
Ours wo SMPL   & 25.85 & 0.775 & 0.161 & 24.70 & 0.765 & 0.232 & 23.59 & 0.870 & 0.199 & 23.21 & 0.642 & 0.258 & 24.33 & 0.763 & 0.212 \\
Ours wo Scale  & 26.43 & 0.785 & 0.151 & 17.49 & 0.493 & 0.549 & 26.49 & 0.787 & 0.140 & 21.70 & 0.539 & 0.174 & 23.02 & 0.651 & 0.254 \\
Ours wo Refine & 27.56 & 0.792 & 0.192 & 25.50 & 0.801 & 0.136 & 26.81 & 0.886 & 0.074 & 23.32 & 0.612 & 0.260 & 25.79 & 0.772 & 0.165 \\
    \bottomrule
    \end{tabular}
    \end{adjustbox}
    \label{tab:breakdown_visdrone}
\end{table*}

\begin{table*}[h]
    \centering
    \caption{\textbf{Breakdown results on the VisDrone dataset} \cite{cao2021visdrone}, focusing on \textbf{human-only} regions. Red, orange, and yellow indicate the first, second, and third best performing algorithms for each metric.}
    \begin{adjustbox}{width=0.85\linewidth,center}
    \begin{tabular}{l|cc|cc|cc|cc|cc}
    \toprule
    \multirow{2}{*}{Method }&\multicolumn{2}{c|}{uav0013\_00000}&\multicolumn{2}{c|}{uav0079\_00480}&\multicolumn{2}{c|}{uav0084\_00000}&\multicolumn{2}{c|}{uav0099\_02109}&\multicolumn{2}{c}{Mean}\\
    & PSNR \(\uparrow\)  & SSIM \(\uparrow\)  & PSNR \(\uparrow\)  & SSIM \(\uparrow\) & PSNR \(\uparrow\)  & SSIM \(\uparrow\)  & PSNR \(\uparrow\)  & SSIM \(\uparrow\)  & PSNR \(\uparrow\)  & SSIM \(\uparrow\)  \\
    \midrule
TK-Planes \cite{maxey2024tk}   &  \cellcolor{tabthird}18.54 &  \cellcolor{tabthird}0.586 &  \cellcolor{tabthird}17.28 &  \cellcolor{tabthird}0.453 &  \cellcolor{tabthird}18.66 & 0.560 & \cellcolor{tabsecond}16.10 & \cellcolor{tabsecond}0.579 & \cellcolor{tabsecond}17.65 &  \cellcolor{tabthird}0.544 \\
3DGS \cite{kerbl20233d}        & \cellcolor{tabsecond}18.72 & \cellcolor{tabsecond}0.640 & \cellcolor{tabsecond}17.50 & \cellcolor{tabsecond}0.512 & 16.81 &  \cellcolor{tabthird}0.565 &  \cellcolor{tabthird}13.01 &  \cellcolor{tabthird}0.474 &  \cellcolor{tabthird}16.51 & \cellcolor{tabsecond}0.548 \\
4DGS \cite{yang2024deformable} & 17.32 & 0.538 & 14.61 & 0.372 &  \cellcolor{tabfirst}20.43 &  \cellcolor{tabfirst}0.675 & 12.65 & 0.455 & 16.25 & 0.510 \\
DeformableGS \cite{wu20244d}   & 12.06 & 0.295 & 11.23 & 0.248 & 15.47 & 0.476 & 11.59 & 0.395 & 12.59 & 0.354 \\
UAV4D (Ours)                    &  \cellcolor{tabfirst}19.07 &  \cellcolor{tabfirst}0.672 &  \cellcolor{tabfirst}17.82 &  \cellcolor{tabfirst}0.525 & \cellcolor{tabsecond}18.79 & \cellcolor{tabsecond}0.629 &  \cellcolor{tabfirst}16.24 &  \cellcolor{tabfirst}0.590 &  \cellcolor{tabfirst}17.98 &  \cellcolor{tabfirst}0.604 \\
    \bottomrule
    \end{tabular}
    \end{adjustbox}
    \label{tab:breakdown_visdrone_human}
\end{table*}
\begin{table*}[h]
    \centering
    \caption{\textbf{Breakdown results on the Okutama-Action dataset} \cite{barekatain2017okutama}. Red, orange, and yellow indicate the first, second, and third best performing algorithms for each metric}
    \begin{adjustbox}{width=0.99\linewidth,center}
    \begin{tabular}{l|ccc|ccc|ccc|ccc|ccc}
    \toprule
    \multirow{2}{*}{Method}&\multicolumn{3}{c|}{1\_2\_2}&\multicolumn{3}{c|}{1\_2\_6}&\multicolumn{3}{c|}{1\_2\_8}&\multicolumn{3}{c|}{1\_2\_9}&\multicolumn{3}{c}{Mean}\\
    & PSNR \(\uparrow\)  & SSIM \(\uparrow\)  & LPIPS \(\downarrow\) & PSNR \(\uparrow\)  & SSIM \(\uparrow\)  & LPIPS \(\downarrow\) & PSNR \(\uparrow\)  & SSIM \(\uparrow\)  & LPIPS \(\downarrow\) & PSNR \(\uparrow\)  & SSIM \(\uparrow\)  & LPIPS \(\downarrow\) & PSNR \(\uparrow\)  & SSIM \(\uparrow\)  & LPIPS \(\downarrow\) \\
    \midrule
TK-Planes \cite{maxey2024tk}   & 30.63 & 0.808 & 0.369 & 24.43 & 0.634 & 0.393 & 31.89 & 0.829 & 0.384 & 24.97 & 0.601 & 0.438 & 27.98 & 0.718 & 0.396 \\
3DGS \cite{kerbl20233d}        & 30.90 &  \cellcolor{tabthird}0.869 &  \cellcolor{tabthird}0.168 & 25.79 &  \cellcolor{tabthird}0.796 &  \cellcolor{tabthird}0.139 & 33.34 &  \cellcolor{tabthird}0.913 &  \cellcolor{tabthird}0.182 &  \cellcolor{tabthird}27.83 & \cellcolor{tabsecond}0.847 & \cellcolor{tabsecond}0.109 & 29.46 &  \cellcolor{tabthird}0.856 &  \cellcolor{tabthird}0.150 \\
4DGS \cite{yang2024deformable} & \cellcolor{tabsecond}33.16 &  \cellcolor{tabthird}0.869 & 0.291 & \cellcolor{tabsecond}28.33 & \cellcolor{tabsecond}0.844 & 0.181 &  \cellcolor{tabthird}33.60 & 0.879 & 0.294 & \cellcolor{tabsecond}28.38 & 0.780 & 0.245 &  \cellcolor{tabthird}30.87 & 0.843 & 0.253 \\
DeformableGS \cite{wu20244d}   &  \cellcolor{tabfirst}34.51 &  \cellcolor{tabfirst}0.945 &  \cellcolor{tabfirst}0.077 &  \cellcolor{tabfirst}28.49 &  \cellcolor{tabfirst}0.897 &  \cellcolor{tabfirst}0.096 & \cellcolor{tabsecond}35.74 &  \cellcolor{tabfirst}0.948 & \cellcolor{tabsecond}0.089 & 26.15 &  \cellcolor{tabthird}0.788 &  \cellcolor{tabthird}0.196 &  \cellcolor{tabfirst}31.22 & \cellcolor{tabsecond}0.894 & \cellcolor{tabsecond}0.114 \\
UAV4D (Ours)                    &  \cellcolor{tabthird}31.76 & \cellcolor{tabsecond}0.897 & \cellcolor{tabsecond}0.105 &  \cellcolor{tabthird}26.57 & \cellcolor{tabsecond}0.844 & \cellcolor{tabsecond}0.100 &  \cellcolor{tabfirst}35.96 & \cellcolor{tabsecond}0.947 &  \cellcolor{tabfirst}0.075 &  \cellcolor{tabfirst}29.49 &  \cellcolor{tabfirst}0.901 &  \cellcolor{tabfirst}0.056 & \cellcolor{tabsecond}30.94 &  \cellcolor{tabfirst}0.897 &  \cellcolor{tabfirst}0.084 \\
    \bottomrule
    \end{tabular}
    \end{adjustbox}
    \label{tab:breakdown_okutama}
\end{table*}
\begin{table*}[h]
    \centering
    \caption{\textbf{Breakdown results on the Okutama-Action dataset} \cite{barekatain2017okutama}, focusing on \textbf{human-only} regions. Red, orange, and yellow indicate the first, second, and third best performing algorithms for each metric.}
    \begin{adjustbox}{width=0.85\linewidth,center}
    \begin{tabular}{l|cc|cc|cc|cc|cc}
    \toprule
    \multirow{2}{*}{Method }&\multicolumn{2}{c|}{1\_2\_2}&\multicolumn{2}{c|}{1\_2\_6}&\multicolumn{2}{c|}{1\_2\_8}&\multicolumn{2}{c|}{1\_2\_9}&\multicolumn{2}{c}{Mean}\\
    & PSNR \(\uparrow\)  & SSIM \(\uparrow\)  & PSNR \(\uparrow\)  & SSIM \(\uparrow\) & PSNR \(\uparrow\)  & SSIM \(\uparrow\)  & PSNR \(\uparrow\)  & SSIM \(\uparrow\)  & PSNR \(\uparrow\)  & SSIM \(\uparrow\)  \\
    \midrule
TK-Planes \cite{maxey2024tk}   & 18.94 & 0.635 & 20.91 & 0.671 & 18.34 & 0.667 &  \cellcolor{tabfirst}18.39 &  \cellcolor{tabfirst}0.550 & \cellcolor{tabsecond}19.14 &  \cellcolor{tabthird}0.631 \\
3DGS \cite{kerbl20233d}        &  \cellcolor{tabfirst}19.97 &  \cellcolor{tabfirst}0.714 &  \cellcolor{tabthird}21.05 &  \cellcolor{tabthird}0.701 &  \cellcolor{tabthird}18.76 &  \cellcolor{tabthird}0.694 & 16.50 & 0.513 &  \cellcolor{tabthird}19.07 & \cellcolor{tabsecond}0.656 \\
4DGS \cite{yang2024deformable} & 15.80 & 0.504 &  \cellcolor{tabfirst}22.28 &  \cellcolor{tabfirst}0.735 & \cellcolor{tabsecond}19.65 & \cellcolor{tabsecond}0.725 & 14.99 & 0.446 & 18.18 & 0.602 \\
DeformableGS \cite{wu20244d}   & \cellcolor{tabsecond}19.52 & \cellcolor{tabsecond}0.702 & 13.43 & 0.505 & 15.51 & 0.565 &  \cellcolor{tabthird}16.86 &  \cellcolor{tabthird}0.514 & 16.33 & 0.571 \\
Our Method                     &  \cellcolor{tabthird}19.01 &  \cellcolor{tabthird}0.667 & \cellcolor{tabsecond}21.07 & \cellcolor{tabsecond}0.704 &  \cellcolor{tabfirst}19.70 &  \cellcolor{tabfirst}0.734 & \cellcolor{tabsecond}18.19 & \cellcolor{tabsecond}0.549 &  \cellcolor{tabfirst}19.49 &  \cellcolor{tabfirst}0.664 \\ 
    \bottomrule
    \end{tabular}
    \end{adjustbox}
    \label{tab:breakdown_okutama_human}
\end{table*}

\begin{table*}[h]
    \centering
    \caption{\textbf{Breakdown results on the Manipal-UAV dataset} \cite{akshatha2023manipal}. Red, orange, and yellow indicate the first, second, and third best performing algorithms for each metric}
    \begin{adjustbox}{width=0.99\linewidth,center}
    \begin{tabular}{l|ccc|ccc|ccc|ccc|ccc}
    \toprule
    \multirow{2}{*}{Method}&\multicolumn{3}{c|}{40\_GND\_P1}&\multicolumn{3}{c|}{40\_VGT\_P1}&\multicolumn{3}{c|}{50\_GND\_P1}&\multicolumn{3}{c|}{50\_RD\_P1}&\multicolumn{3}{c}{Mean}\\
    & PSNR \(\uparrow\)  & SSIM \(\uparrow\)  & LPIPS \(\downarrow\) & PSNR \(\uparrow\)  & SSIM \(\uparrow\)  & LPIPS \(\downarrow\) & PSNR \(\uparrow\)  & SSIM \(\uparrow\)  & LPIPS \(\downarrow\) & PSNR \(\uparrow\)  & SSIM \(\uparrow\)  & LPIPS \(\downarrow\) & PSNR \(\uparrow\)  & SSIM \(\uparrow\)  & LPIPS \(\downarrow\) \\
    \midrule
TK-Planes \cite{maxey2024tk}   & 28.09 & 0.842 & 0.383 & 26.88 & 0.732 & 0.427 & 27.59 & 0.823 & 0.364 & 29.63 & 0.788 & 0.494 &  \cellcolor{tabthird}28.05 &  \cellcolor{tabthird}0.796 & 0.417 \\
3DGS \cite{kerbl20233d}        &  \cellcolor{tabthird}31.02 &  \cellcolor{tabthird}0.907 & \cellcolor{tabsecond}0.204 &  \cellcolor{tabthird}27.54 & \cellcolor{tabsecond}0.807 & \cellcolor{tabsecond}0.234 &  \cellcolor{tabthird}30.32 & \cellcolor{tabsecond}0.911 & \cellcolor{tabsecond}0.191 & \cellcolor{tabsecond}30.72 & \cellcolor{tabsecond}0.845 & \cellcolor{tabsecond}0.283 & \cellcolor{tabsecond}29.90 & \cellcolor{tabsecond}0.867 & \cellcolor{tabsecond}0.228 \\
4DGS \cite{yang2024deformable} & 21.27 & 0.710 & 0.503 &  \cellcolor{tabfirst}27.99 &  \cellcolor{tabthird}0.796 &  \cellcolor{tabthird}0.318 &  \cellcolor{tabfirst}31.00 &  \cellcolor{tabthird}0.910 &  \cellcolor{tabthird}0.248 & 23.34 & 0.705 & 0.569 & 25.90 & 0.780 &  \cellcolor{tabthird}0.410 \\
DeformableGS \cite{wu20244d}   & \cellcolor{tabsecond}31.28 & \cellcolor{tabsecond}0.915 &  \cellcolor{tabthird}0.251 & 13.10 & 0.564 & 0.624 & 16.04 & 0.687 & 0.536 &  \cellcolor{tabthird}30.13 &  \cellcolor{tabthird}0.827 &  \cellcolor{tabthird}0.361 & 22.64 & 0.748 & 0.443 \\
UAV4D (Ours)                     &  \cellcolor{tabfirst}31.34 &  \cellcolor{tabfirst}0.917 &  \cellcolor{tabfirst}0.181 & \cellcolor{tabsecond}27.88 &  \cellcolor{tabfirst}0.814 &  \cellcolor{tabfirst}0.191 & \cellcolor{tabsecond}30.86 &  \cellcolor{tabfirst}0.919 &  \cellcolor{tabfirst}0.143 &  \cellcolor{tabfirst}31.36 &  \cellcolor{tabfirst}0.849 &  \cellcolor{tabfirst}0.219 &  \cellcolor{tabfirst}30.36 &  \cellcolor{tabfirst}0.875 &  \cellcolor{tabfirst}0.184 \\
    \bottomrule
    \end{tabular}
    \end{adjustbox}
    \label{tab:breakdown_manipal}
\end{table*}

\begin{table*}[h]
    \centering
    \caption{\textbf{Breakdown results on the Manipal-UAV dataset} \cite{akshatha2023manipal}, focusing on \textbf{human-only} regions. Red, orange, and yellow indicate the first, second, and third best performing algorithms for each metric.}
    \begin{adjustbox}{width=0.99\linewidth,center}
    \begin{tabular}{l|cc|cc|cc|cc|cc}
    \toprule
    \multirow{2}{*}{Method }&\multicolumn{2}{c|}{40\_GND\_P1}&\multicolumn{2}{c|}{40\_VGT\_P1}&\multicolumn{2}{c|}{50\_GND\_P1}&\multicolumn{2}{c|}{50\_RD\_P1}&\multicolumn{2}{c}{Mean}\\
    & PSNR \(\uparrow\)  & SSIM \(\uparrow\)  & PSNR \(\uparrow\)  & SSIM \(\uparrow\) & PSNR \(\uparrow\)  & SSIM \(\uparrow\)  & PSNR \(\uparrow\)  & SSIM \(\uparrow\)  & PSNR \(\uparrow\)  & SSIM \(\uparrow\)  \\
    \midrule
TK-Planes \cite{maxey2024tk}   & 20.52 &  \cellcolor{tabthird}0.838 & \cellcolor{tabsecond}15.79 & 0.577 & 19.11 & 0.822 & 18.11 & 0.769 & 18.38 & 0.752 \\
3DGS \cite{kerbl20233d}        & 19.35 & 0.816 & 15.45 & 0.587 & 18.55 & 0.798 &  \cellcolor{tabthird}18.60 & \cellcolor{tabsecond}0.791 & 17.99 & 0.748 \\
4DGS \cite{yang2024deformable} & \cellcolor{tabsecond}21.41 & \cellcolor{tabsecond}0.845 &  \cellcolor{tabfirst}15.83 &  \cellcolor{tabfirst}0.603 & \cellcolor{tabsecond}20.98 & \cellcolor{tabsecond}0.850 & \cellcolor{tabsecond}18.68 &  \cellcolor{tabthird}0.783 & \cellcolor{tabsecond}19.23 & \cellcolor{tabsecond}0.770 \\
DeformableGS \cite{wu20244d}   &  \cellcolor{tabthird}20.66 & 0.837 &  \cellcolor{tabthird}15.64 & \cellcolor{tabsecond}0.597 &  \cellcolor{tabthird}20.27 &  \cellcolor{tabthird}0.843 & 18.05 & 0.771 &  \cellcolor{tabthird}18.65 &  \cellcolor{tabthird}0.762 \\
Our Method                     &  \cellcolor{tabfirst}21.92 &  \cellcolor{tabfirst}0.880 & 15.61 &  \cellcolor{tabthird}0.594 &  \cellcolor{tabfirst}21.24 &  \cellcolor{tabfirst}0.872 &  \cellcolor{tabfirst}18.76 &  \cellcolor{tabfirst}0.792 &  \cellcolor{tabfirst}19.38 &  \cellcolor{tabfirst}0.784 \\
    \bottomrule
    \end{tabular}
    \end{adjustbox}
    \label{tab:breakdown_manipal_human}
\end{table*}

\FloatBarrier
\subsection{Visual Comparison}
\begin{figure}[h]
\captionsetup[subfigure]{labelformat=empty}
\centering

\begin{subfigure}{0.192\linewidth}
  \centering
  \includegraphics[width=1\linewidth]{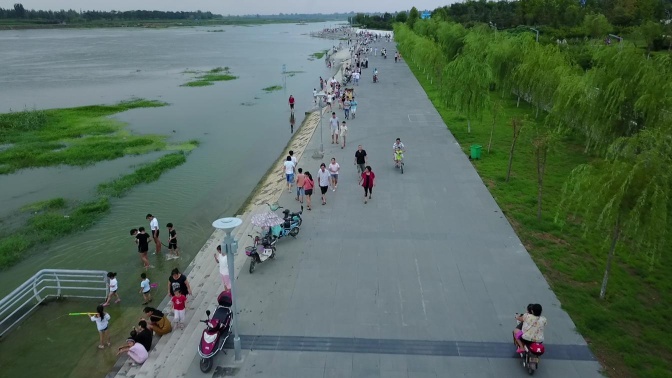} \\[0.5ex]
  \includegraphics[width=1\linewidth]{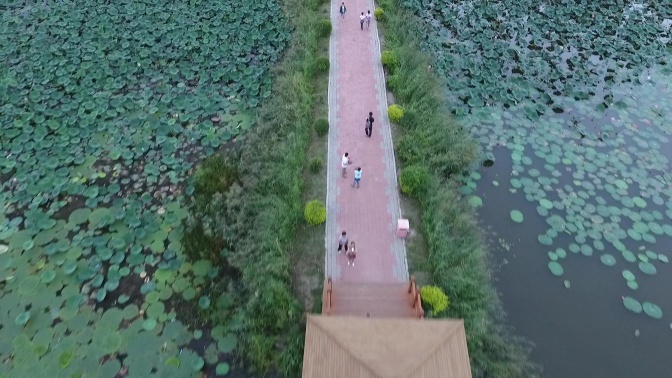}
  \vspace{-2mm}
  \caption{Scene}
\end{subfigure}
\begin{subfigure}{0.192\linewidth}
  \centering
  \includegraphics[width=1\linewidth]{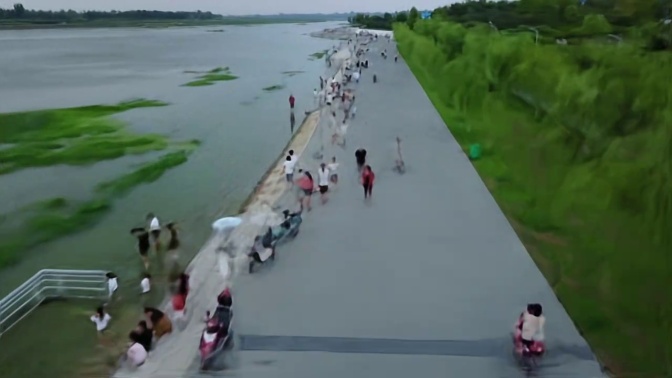} \\[0.5ex]
  \includegraphics[width=1\linewidth]{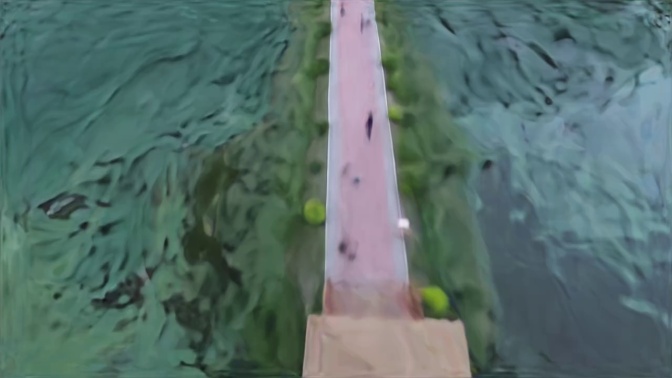}
  \vspace{-2mm}
  \caption{TK-Planes \cite{maxey2024tk}}
\end{subfigure}
\begin{subfigure}{0.192\linewidth}
  \centering
  \includegraphics[width=1\linewidth]{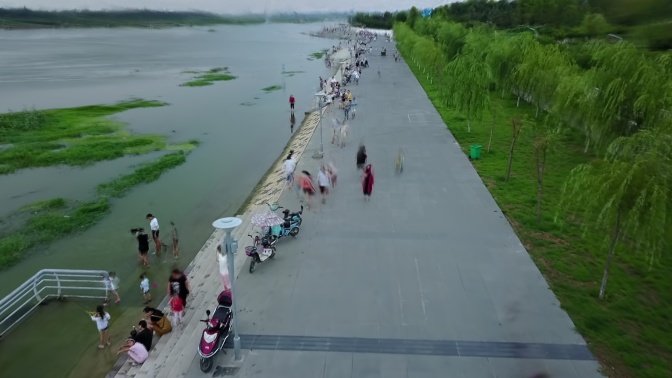} \\[0.5ex]
  \includegraphics[width=1\linewidth]{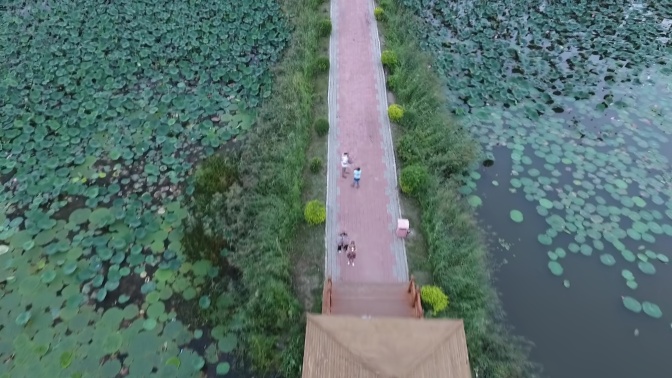}
  \vspace{-2mm}
  \caption{3DGS \cite{kerbl20233d}}
\end{subfigure}
\begin{subfigure}{0.192\linewidth}
  \centering
  \includegraphics[width=1\linewidth]{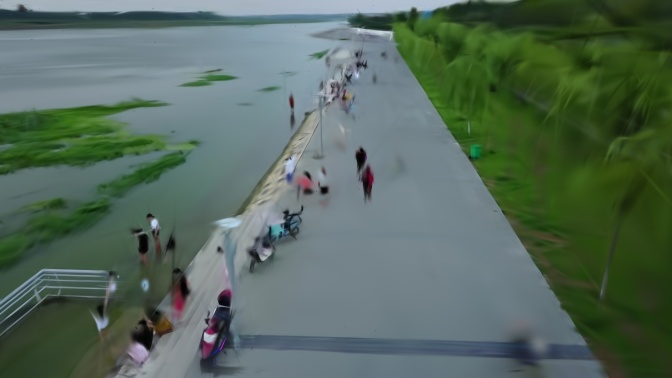} \\[0.5ex]
  \includegraphics[width=1\linewidth]{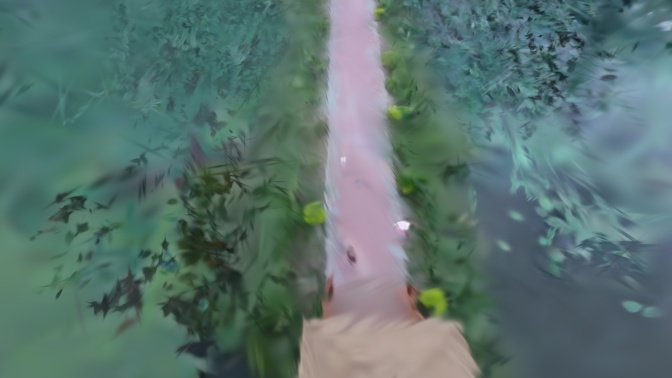}
  \vspace{-2mm}
  \caption{4DGS \cite{wu20244d}}
\end{subfigure}
\begin{subfigure}{0.192\linewidth}
  \centering
  \includegraphics[width=1\linewidth]{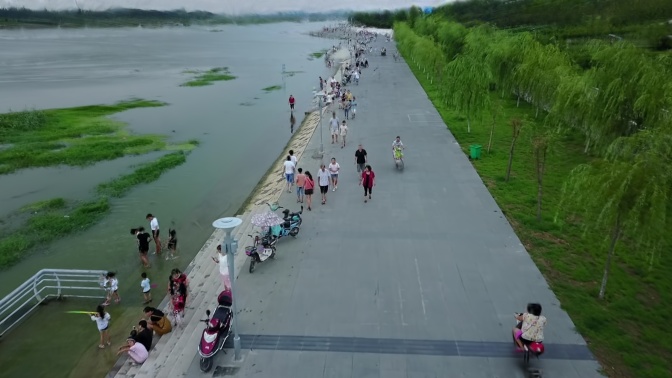} \\[0.5ex]
  \includegraphics[width=1\linewidth]{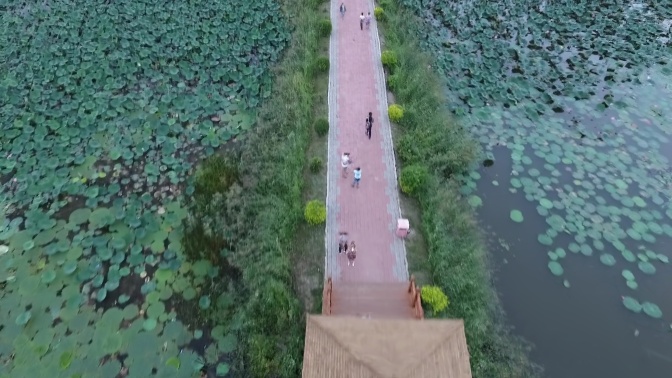}
  \vspace{-2mm}
  \caption{UAV4D (Ours)}
\end{subfigure}

\textbf{Visual Comparison: VisDrone Dataset \cite{cao2021visdrone}.} 

\begin{subfigure}{0.192\linewidth}
  \centering
  \includegraphics[width=1\linewidth]{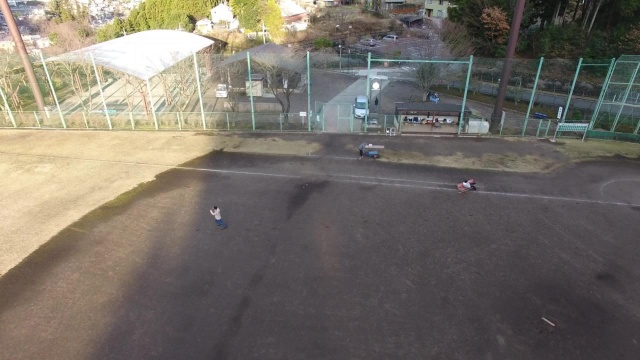} \\[0.5ex]
  \includegraphics[width=1\linewidth]{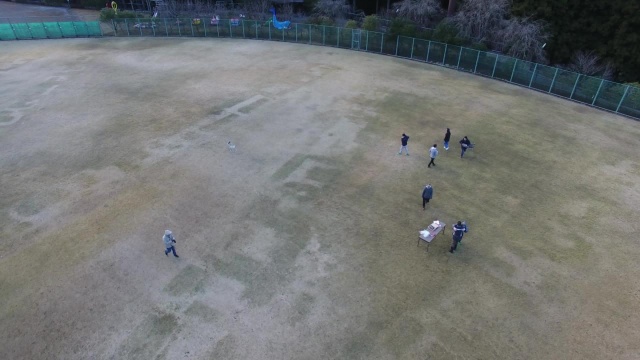}
  \vspace{-2mm}
  \caption{Scene}
\end{subfigure}
\begin{subfigure}{0.192\linewidth}
  \centering
  \includegraphics[width=1\linewidth]{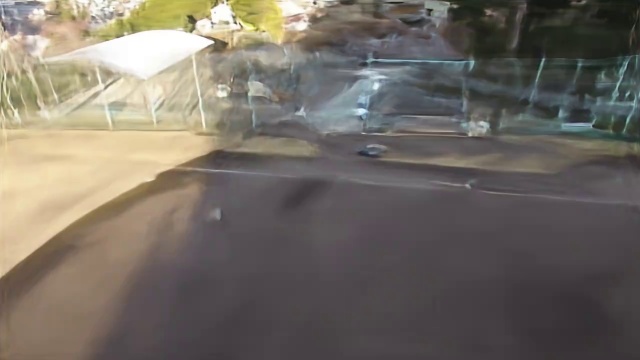} \\[0.5ex]
  \includegraphics[width=1\linewidth]{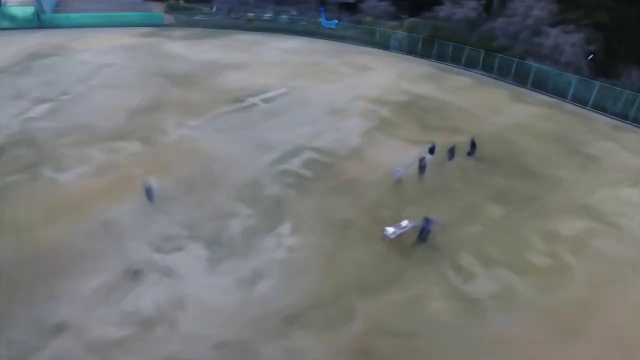}
  \vspace{-2mm}
  \caption{TK-Planes \cite{maxey2024tk}}
\end{subfigure}
\begin{subfigure}{0.192\linewidth}
  \centering
  \includegraphics[width=1\linewidth]{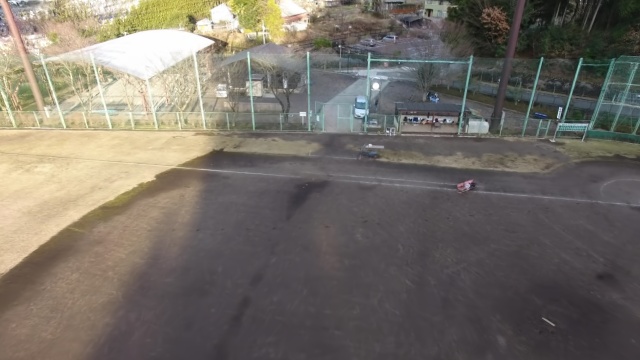} \\[0.5ex]
  \includegraphics[width=1\linewidth]{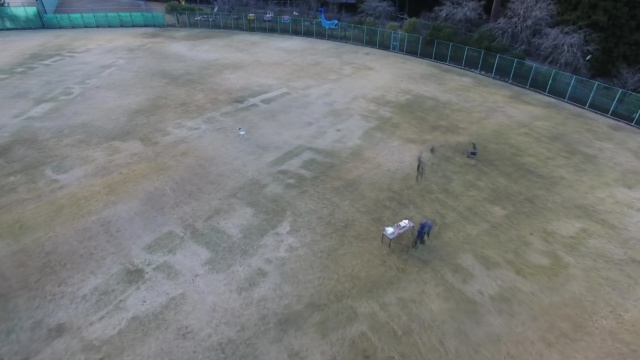}
  \vspace{-2mm}
  \caption{3DGS \cite{kerbl20233d}}
\end{subfigure}
\begin{subfigure}{0.192\linewidth}
  \centering
  \includegraphics[width=1\linewidth]{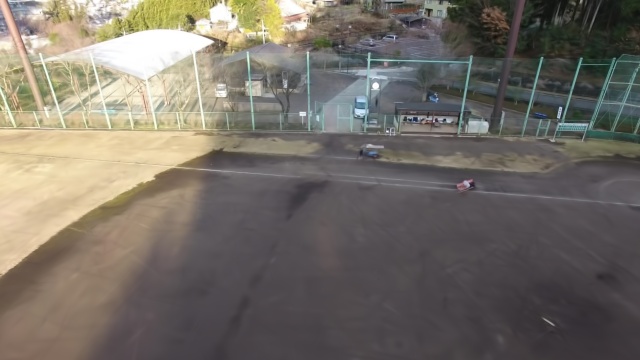} \\[0.5ex]
  \includegraphics[width=1\linewidth]{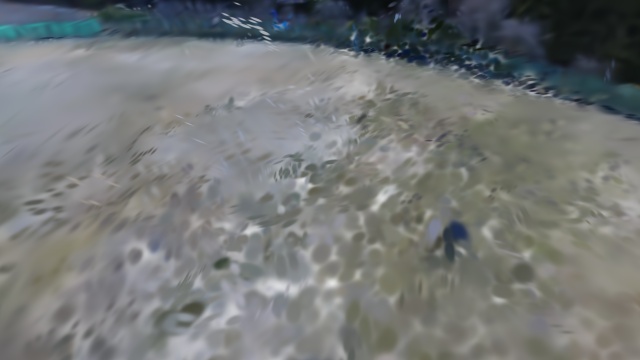}
  \vspace{-2mm}
  \caption{4DGS \cite{wu20244d}}
\end{subfigure}
\begin{subfigure}{0.192\linewidth}
  \centering
  \includegraphics[width=1\linewidth]{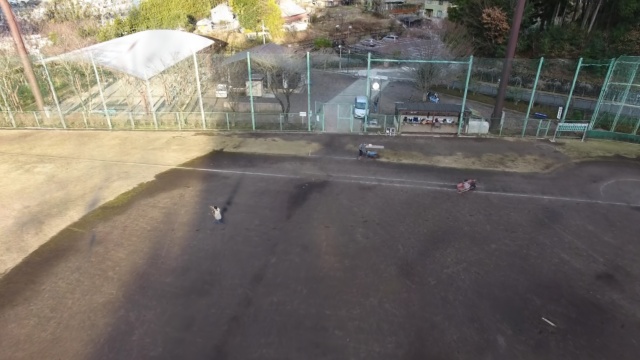} \\[0.5ex]
  \includegraphics[width=1\linewidth]{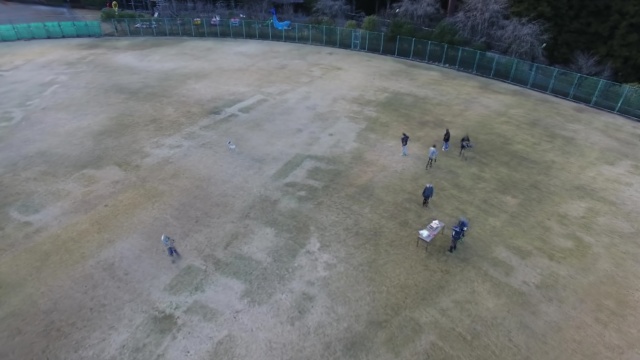}
  \vspace{-2mm}
  \caption{UAV4D (Ours)}
\end{subfigure}

\textbf{Visual Comparison: Okutama-Action Dataset \cite{barekatain2017okutama}.} 


\begin{subfigure}{0.192\linewidth}
  \centering
  \includegraphics[width=1\linewidth]{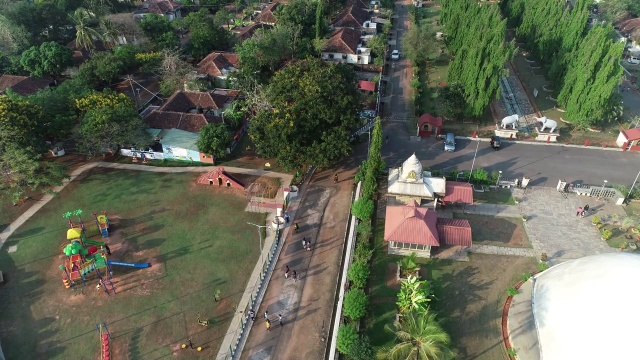} \\[0.5ex]
  \includegraphics[width=1\linewidth]{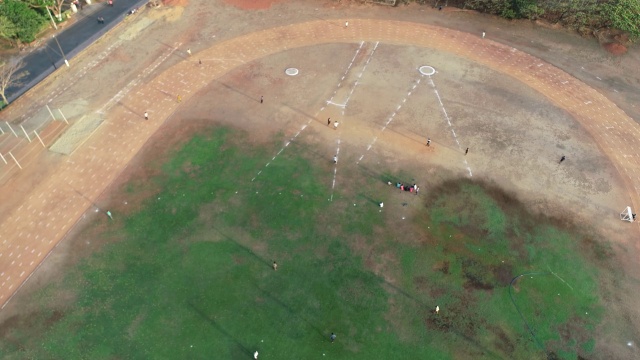}
  \vspace{-2mm}
  \caption{Scene}
\end{subfigure}
\begin{subfigure}{0.192\linewidth}
  \centering
  \includegraphics[width=1\linewidth]{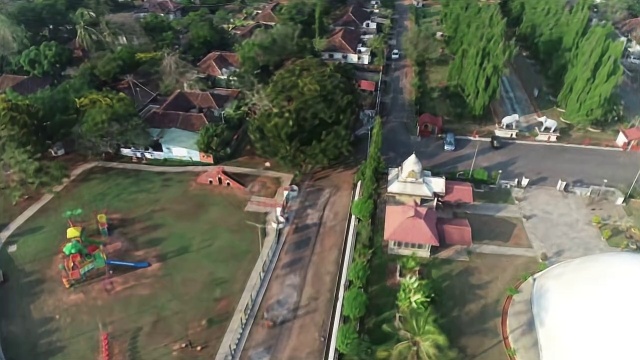} \\[0.5ex]
  \includegraphics[width=1\linewidth]{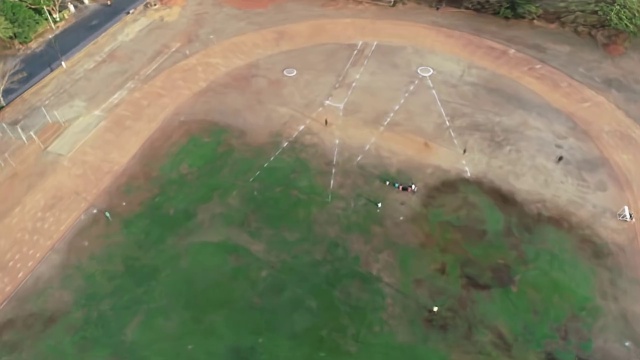}
  \vspace{-2mm}
  \caption{TK-Planes \cite{maxey2024tk}}
\end{subfigure}
\begin{subfigure}{0.192\linewidth}
  \centering
  \includegraphics[width=1\linewidth]{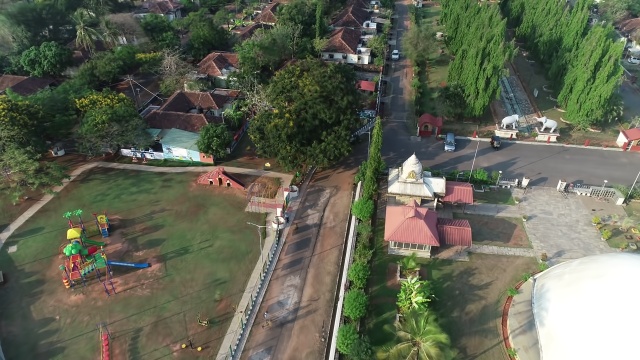} \\[0.5ex]
  \includegraphics[width=1\linewidth]{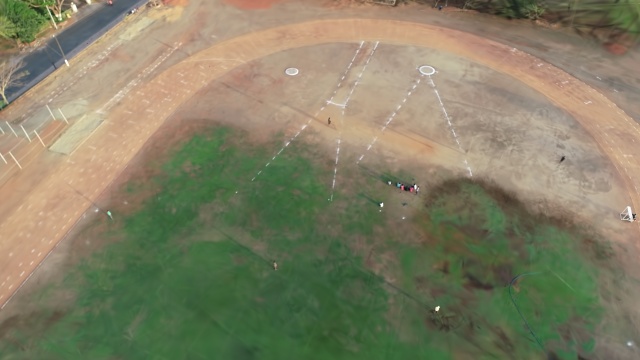}
  \vspace{-2mm}
  \caption{4DGS \cite{wu20244d}}
\end{subfigure}
\begin{subfigure}{0.192\linewidth}
  \centering
  \includegraphics[width=1\linewidth]{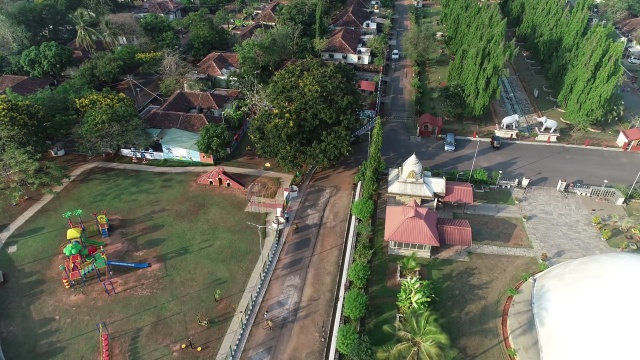} \\[0.5ex]
  \includegraphics[width=1\linewidth]{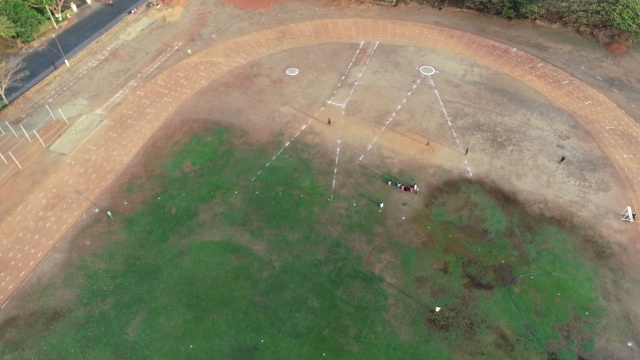}
  \vspace{-2mm}
  \caption{Deform-GS \cite{yang2024deformable}}
\end{subfigure}
\begin{subfigure}{0.192\linewidth}
  \centering
  \includegraphics[width=1\linewidth]{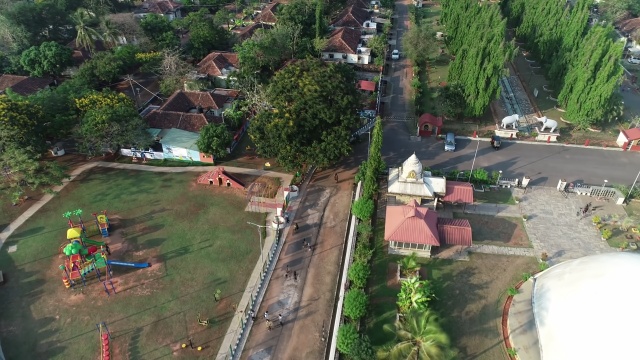} \\[0.5ex]
  \includegraphics[width=1\linewidth]{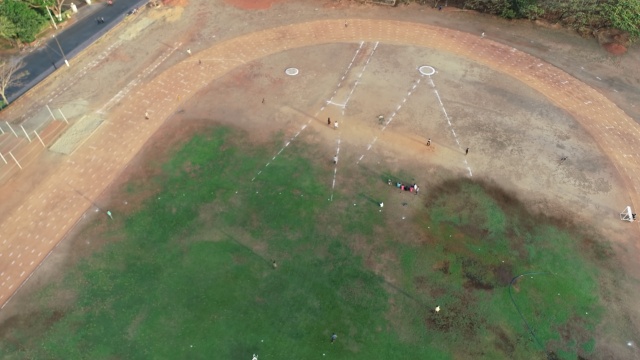}
  \vspace{-2mm}
  \caption{UAV4D (Ours)}
\end{subfigure}

\textbf{Visual Comparison: Manipal-UAV Dataset \cite{akshatha2023manipal}.} 

\caption{
\textbf{Visual Comparisons across UAV Datasets.} Rendering comparisons on VisDrone \cite{cao2021visdrone}, Okutama-Action \cite{barekatain2017okutama}, and Manipal-UAV \cite{akshatha2023manipal}. Our method consistently outperforms other baselines in handling dynamic scenes with small, moving humans.
}
\label{fig:combined_comparison}
\end{figure}

Figure \ref{fig:combined_comparison} presents the remaining scenes not included in the main paper due to space limitations. Additionally, we provide a project website at \href{https://uav4d.github.io/}{https://uav4d.github.io/}, which contains rendering videos for all scenes.

\FloatBarrier

\newpage

\end{document}